\documentclass{article}

\usepackage[preprint]{corl_2026} %
\usepackage[T1]{fontenc} %
\usepackage{graphicx}
\usepackage{etoc}
\usepackage{seqsplit}

\newcommand{\skill}[1]{\texttt{\seqsplit{#1}}}
\usepackage{tabularx}
\usepackage[utf8]{inputenc}
\usepackage{newunicodechar}
\newunicodechar{↔}{\leftrightarrow}

\title{Playful Agentic Robot Learning}
\usepackage{etoolbox}
\makeatletter
\patchcmd{\@maketitle}{\@title\par}{\@title\par\vskip 12pt}%
  {}{\typeout{!! WARNING: title-author spacing patch FAILED}}
\makeatother
\newcommand\blfootnote[1]{%
  \begingroup
  \renewcommand\thefootnote{}\footnote{#1}%
  \addtocounter{footnote}{-1}%
  \endgroup
}
\usepackage{booktabs}
\usepackage{algorithm}      %
\usepackage{algpseudocode}  %
\usepackage{amsmath}        %
\usepackage{amssymb}        %
\usepackage{multirow}
\usepackage{enumitem}
\author{
  \parbox{\textwidth}{\centering %
  \makebox[\textwidth][c]{Junyi Zhang$^{1,*,\dagger}$, Jiaxin Ge$^{1,*,\dagger}$, Hanjun Yoo$^{1,\dagger}$, Letian Fu$^{1,\ddagger}$, Zihan Yang$^{2,\ddagger}$, Yaowei Liu$^{2,\ddagger}$, Raj Saravanan$^{1,\ddagger}$} \\[4pt]
  \makebox[\textwidth][c]{Shaofeng Yin$^{2}$, Justin Yu$^{1}$, Dantong Niu$^{1}$, Zirui Wang$^{1}$, Roei Herzig$^{1}$, Ken Goldberg$^{1}$, Yutong Bai$^{1}$} \\[4pt]
  \makebox[\textwidth][c]{David M.~Chan$^{1}$, Ion Stoica$^{1}$, Angjoo Kanazawa$^{1}$, Jiahui Lei$^{1,\S}$, Haiwen Feng$^{1,2,\S}$, Trevor Darrell$^{1}$} \\
  \vspace{1.em}
  $^{1}$University of California, Berkeley \quad $^{2}$Impossible Research \\[7pt]
  {\url{https://Playful-RATs.github.io}}
  }
}
\usepackage{listings}

\definecolor{cvprblue}{rgb}{0.21,0.49,0.74} %
\hypersetup{
  colorlinks=true,
  citecolor=cvprblue,
  linkcolor=cvprblue,
  urlcolor=cvprblue,
}

\begin{document}
\maketitle
\blfootnote{ $^{*}$\,project leads, ordered by coin flip; \quad $^{\dagger,\ddagger,\S}$\,equal contribution}
\etocdepthtag.toc{mtchapter}

\begin{figure}[h]
\centering
\vspace{-2em}
    \includegraphics[width=1.01\linewidth]{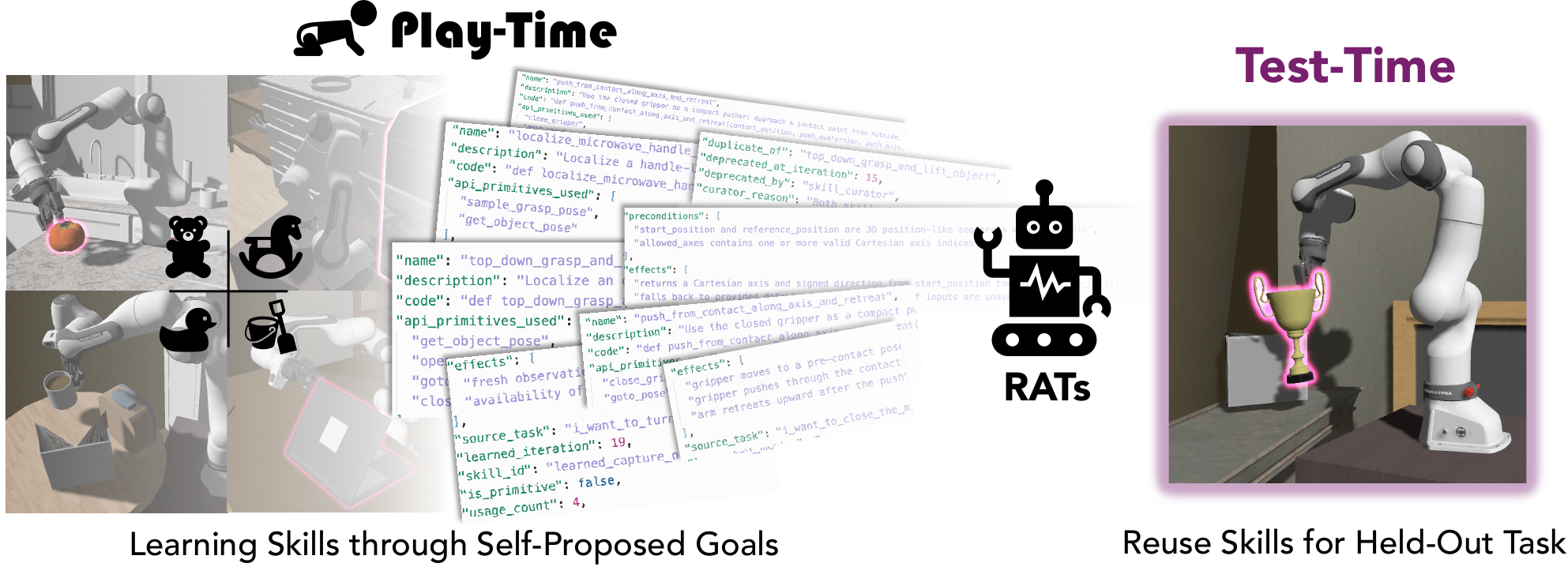}
    \vspace{-0.75em}
    \caption{
\textbf{\textsc{RATs} enables Playful Agentic Robot Learning.}
Prior to receiving extrinsic reward signals, play-based Robotics Agent Teams autonomously propose intrinsic goals, practice them through Code-as-Policy execution, and distill successful behaviors into a reusable code skill library. At test time, the learned skills are retrieved and reused to solve future tasks.
}
    \label{fig:teaser}
    \vspace{0.75em}
\end{figure}

\begin{abstract}
Current agentic robot systems can write executable Code-as-Policy programs, observe feedback, and revise behavior across multiple attempts, but they remain largely task-driven: reusable skills are acquired only after explicit instructions. We study \emph{Playful Agentic Robot Learning}, where an embodied coding agent uses self-directed play as a continual skill-learning stage before downstream tasks arrive. We introduce \textsc{RATs}, \underline{R}obotics \underline{A}gent \underline{T}eam\underline{s} designed for play-time skill acquisition. During play, \textsc{RATs} proposes novel yet learnable exploratory tasks, plans and executes robot-code policies, verifies intermediate progress, diagnoses failures, retries with dense, step-level feedback, and distills successful executions into a persistent code skill library. 
At test time, the agent reuses relevant skills from this frozen library to help solve new tasks.
Experiments in LIBERO-PRO and MolmoSpaces show that play-learned skills improve held-out downstream tasks over no play and random-play baselines, with a 20.6 and 17.0 percentage-point gains over CaP-Agent0 on LIBERO-PRO and MolmoSpaces, respectively. Moreover, the learned skills can be plugged into other inference-time Code-as-Policy agents by simply retrieving them into the context, improving RoboSuite and real-world transfer by 8.9 and 8.8 points, respectively, without finetuning the underlying model.
\end{abstract}


\section{Introduction}
\label{sec:introduction}

Recent foundation models have made robot learning increasingly \emph{agentic}: language and multimodal models can generate plans, call perception and control tools, write executable Code-as-Policy programs, observe feedback, and revise behavior over multiple attempts \citep{liang2023code, singh2023progprompt, vemprala2024chatgpt}. Such agents offer a modular and inspectable alternative to end-to-end vision-language-action policies \citep{zitkovich2023rt2, black2025pi0, physicalintelligence2025pi05}, and recent frameworks such as CaP-X~\citep{fu2026capx} show that embodied coding agents can benefit from multi-turn execution feedback, visual differencing, and automatic skill synthesis. 
However, most current agentic robot systems remain largely \emph{task-driven}: they learn only after receiving external instructions. Even when successful executions are stored as reusable skills, continual learning remains reactive rather than proactive.

Natural intelligence suggests a different model. Children acquire reusable skills through play before explicit goals are given, discovering controllable effects and practicing near the boundary of their competence \citep{piaget1952origins, gopnik2020childhood, smith2005developmental}. This idea has long motivated developmental robotics and intrinsic motivation, where agents favor experiences that are novel yet learnable \citep{schmidhuber1991curious, oudeyer2007intrinsic, baranes2013active, forestier2016modular}. We argue that these classical ideas are newly powerful in the era of Code-as-Policy agents: unlike earlier systems that explored fixed sensorimotor, goal, or feature spaces \citep{oudeyer2007intrinsic, baranes2013active, pathak2017curiosity, houthooft2016vime}, coding agents can express exploratory goals in language, execute them as programs, inspect outcomes, and save successful behaviors as callable code. This makes play a practical mechanism for acquiring reusable robot skills before downstream tasks are specified.

In this paper, we study \emph{Playful Agentic Robot Learning}: how an agentic robot system can use self-directed play as a continual skill-learning stage before deployment to downstream tasks. 
We introduce \textsc{RATs}, Robotics Agent Teams designed to realize this setting. \textsc{RATs} treats play not merely as open-ended exploration, but as an explicit skill-acquisition process. 
In each environment, the agent team proposes exploratory tasks, plans, and executes Code-as-Policy programs, verifies progress, diagnoses failures, retries with feedback, and distills successful executions into a persistent code skill library. 
The learned library is then reused at inference time to solve novel tasks. In this way, \textsc{RATs} asks not only how a robot should solve a task once it is given, but what skills the robot should practice and accumulate before being asked.

A central design goal of \textsc{RATs} is to make play informative enough for continual skill learning. A single task-level success or failure signal is often too sparse to explain what the robot has learned, which substep failed, or what reusable behavior should be saved. 
\textsc{RATs} therefore uses a structured agent team with planning, per-step verification, multi-attempt retry, failure diagnosis, and memory update. These components provide dense feedback over intermediate subgoals and execution attempts, allowing the system to preserve working parts of a policy, localize missing capabilities, and convert successful behaviors into reusable skills. Task proposal uses a simple novelty--learnability rule, encouraging the agent to practice interactions that expand the current skill library while remaining physically achievable \citep{oudeyer2007intrinsic, baranes2013active}.

We evaluate \textsc{RATs} in simulated play environments, including LIBERO-PRO and MolmoSpaces, and test whether skills acquired during play improve held-out downstream benchmark tasks. On LIBERO-PRO, \textsc{RATs} improves average success by 20.6 percentage points over CaP-Agent0; on MolmoSpaces, it improves average success by 17.0 points. The learned skills also transfer across environments: LIBERO-PRO skills plugged into CaP-Agent0 improve RoboSuite success by 8.9 points. These results suggest that play-learned code skills provide a practical, plug-and-play skill-library mechanism for improving agentic robot systems without finetuning the underlying model.

In summary, we formulate \emph{Playful Agentic Robot Learning}, where embodied coding agents acquire reusable skills through self-directed play before downstream tasks are given. We design \textsc{RATs}, Robotics Agent Teams for play-time continual skill learning, with planning, per-step verification, retry, diagnosis, and memory mechanisms that provide dense feedback for learning reusable skills from autonomous interaction. We show that the resulting skill library improves downstream performance both within the full \textsc{RATs} system and as a plug-and-play addition to other inference-time Code-as-Policy methods.

\section{Related Work}
\label{sec:related}

\textbf{Play, Curiosity, and Developmental Robotics.}
Play has long been viewed as a central mechanism for skill development in natural intelligence: before children are given explicit tasks, they explore objects, discover controllable effects, and practice emerging motor routines~\citep{piaget1952origins, vygotsky1978mind, pellegrini2009role, gopnik2020childhood}. Rather than random activity, play provides a self-directed curriculum that samples interactions that are novel, meaningful, and near the boundary of current competence. This view has inspired computational models of curiosity and intrinsic motivation, where agents acquire skills before external rewards by pursuing prediction error, information gain, novelty, learning progress, or intermediate difficulty~\citep{schmidhuber1991curious, schmidhuber2010formal, houthooft2016vime, oudeyer2007intrinsic, kaplan2007intrinsic, kidd2012goldilocks}. Developmental robotics instantiated these ideas through Intelligent Adaptive Curiosity, goal babbling, competence-progress-driven goal selection, and modular curiosity systems for discovering object interactions and tool-use precursors~\citep{rolf2010goal, baranes2013active, forestier2016modular, forestier2017intrinsically}. In robot learning, play has provided broad, task-agnostic interaction data for learning reusable manipulation behaviors, latent plans, and hierarchical policies without segmented task demonstrations~\citep{lynch2020learning, wang2023mimicplay}, while language-conditioned curiosity uses compositional goals to expand exploration~\citep{colas2020language}. \textsc{RATs} extends this play-driven skill-acquisition paradigm to Code-as-Policy agents, where self-proposed practice goals are executed as robot programs and distilled into reusable code skills.

\textbf{Continual Skill Learning and Curriculum in Robotics.}
Continual robot learning studies how agents acquire, retain, and reuse knowledge over extended experience~\citep{thrun1995lifelong, parisi2019continual, lesort2020continual}. In manipulation, this often takes the form of reusable skills, options, motor primitives, affordances, behavioral priors, or hierarchical policies that can be transferred and composed across tasks~\citep{sutton1999between, konidaris2012robot, kroemer2021review, pertsch2020long, lynch2021language, wan2024lotuscontinualimitationlearning, ma2024dreureka}. Curriculum learning provides a complementary mechanism for improving long-horizon learning by ordering tasks from simple to complex~\citep{bengio2009curriculum, narvekar2020curriculum}, selecting tasks, goals, or demonstrations near the agent's competence boundary, expanding from known states, or training goal-conditioned policies over diverse goals~\citep{florensa2018automatic, nair2018visual, pong2020skewfit, colas2020language, seita2019zpd}. These methods provide mechanisms for skill reuse and automatic task ordering, but typically assume a predefined task family, goal representation, reward function, or externally provided experience stream. \textsc{RATs} differs by treating the curriculum itself as an object of autonomous discovery in a Code-as-Policy agent.

\textbf{Code-as-Policy and Agentic Robot Learning.}
Code-as-Policy methods use large language or multimodal models to synthesize executable robot programs that compose perception, planning, and control APIs, making robot policies modular, inspectable, and reusable~\citep{liang2023code, singh2023progprompt, vemprala2024chatgpt, mu2024robocodexmultimodalcodegeneration, geminirobotics1d52025, shi2025maestro}. Related embodied LMM systems ground language plans in pretrained skills, affordance models, scene representations, visual feedback, or 3D value maps for long-horizon manipulation~\citep{ahn2022saycan, driess2023palme, huang2023voxposer, zeng2023socratic, mu2023embodied}. Agentic methods further use reasoning-action loops, execution feedback, self-refinement, and tool use to improve generated programs~\citep{yao2023react, shinn2023reflexion, madaan2023selfrefine, chen2024selfdebug}; in robotics, CaP-X shows that multi-turn feedback, visual differencing, parallel reasoning, skill synthesis, and verifiable rewards improve embodied coding agents~\citep{fu2026capx}. However, most robot Code-as-Policy systems remain task-driven: an agent writes and revises code for a given instruction, and reusable skills are accumulated only as byproducts of solving provided tasks. Open-ended LMM agents such as Voyager show that automatic curricula and executable skill libraries can support lifelong exploration~\citep{wang2023voyager}, while sleep-time compute suggests spending computation before queries arrive to improve future performance~\citep{lin2025sleeptime}. \textsc{RATs} brings this skill-acquisition view to physically grounded Code-as-Policy learning, asking what the robot should practice before downstream tasks are specified.

\section{Robotics Agent Teams for Playful Agentic Robot Learning}
\label{sec:method}

\begin{algorithm}[t]
\caption{\textsc{RATs} at Play}
\label{alg:rats}
\begin{algorithmic}[1]
\Require Training Env $\mathcal{E}_{train}$; Test Env $\mathcal{E}_{test}$; Iterations $N$
\Ensure Final Evaluation Score $S_{learned}$, Learned Skill Library $\mathcal{L}$, Initial primitive library $\mathcal{L}_0$
\State $\mathcal{L} \gets \mathcal{L}_0$, $\mathcal{M} \gets \emptyset$
    \Comment{Initialize skill library with primitives and empty memory}

\Statex \hspace{1.2em}\textit{// Phase 1: Play-Time Skill Acquisition}
\For{$t = 1 \dots N$} 
    \State $\tau_t \gets \textsc{ProposeTask}(\mathcal{E}_{train},\, \mathcal{L},\, \mathcal{M})$ 
        \Comment{\textbf{Step 1: Propose task}}
    \State $\text{success}_t,\, \pi_t \gets \textsc{RunTask}(\tau_t,\, \mathcal{L})$ 
        \Comment{\textbf{Step 2: Agent system runs task}}
    \State $\mathcal{L},\, \mathcal{M} \gets \textsc{UpdateMemory}(\text{success}_t,\, \pi_t,\, \mathcal{L},\, \mathcal{M})$ 
        \Comment{\textbf{Step 3: Update skill \& memory}}
\EndFor

\Statex \hspace{1.2em}\textit{// Phase 2: Evaluation with Learned Skill Library (via} \textsc{RATs} \textit{Exec. or plug-in CaP-Agent0)}
\State $S_{learned} \gets \textsc{Evaluate}(\mathcal{E}_{test},\, \textsc{Agent}(\mathcal{L}))$ 
    \Comment{\textbf{Step 4: Use learned skills on test set}}
\State \Return $S_{learned},\, \mathcal{L}$
\end{algorithmic}
\end{algorithm}

\subsection{Problem Setup and Overview}
\label{sec:setup}

\textsc{RATs} is a multi-agent Code-as-Policy system that enables \textit{Playful Agentic Robot Learning}, \textit{i.e.}, optimizing and accumulating reusable skills through interactions with the environment, driven by intrinsic reward signals.
\textbf{Code-as-Policy Formulation}: In a standard Code-as-Policy (CaP) framework, an agent is provided with an environment context $c$, some existing primitive functions $f$, and a language instruction $l$. The agent then synthesizes an executable program $\pi$ (\textit{e.g.}, Python code) conditioned on $(c, f, l)$ that can be executed to accomplish the task. 
\textbf{Play-Time Formulation}:
We study a play-time setting in which the above external task instruction $l$ is removed. Instead, the agent autonomously proposes and practices self-generated tasks $\tau_t$ in a play environment $\mathcal{E}_{\text{play}}$, with the goal of acquiring skills that improve later downstream task solving.

\textbf{Method Overview}: During play-time, \textsc{RATs} maintains a skill library $\mathcal{L}$ and a failure memory $\mathcal{M}$. The skill library is defined as
$\mathcal{L} = \mathcal{L}_0 \cup \mathcal{L}_{\text{learned}}$, where $\mathcal{L}_0 \equiv f$ contains the initial primitives and $\mathcal{L}_{\text{learned}}$ stores code skills extracted from successful play executions. The failure memory $\mathcal{M}$ stores compact lessons from failed attempts, such as missing preconditions or corrections useful for retry.
After $N$ play iterations, the learned library $\mathcal{L}$ is reused at test time. The objective is to obtain a library that improves performance on unseen test tasks over using the initial primitives $\mathcal{L}_0$ alone.
Algorithm~\ref{alg:rats} describes the full loop of \textsc{RATs} at Play.
Concretely, \textsc{RATs} is organized as three collaborating teams: a \textbf{Task Proposer} team that selects which play tasks to practice (Sec.~\ref{sec:proposer}), an \textbf{Execution} team that writes, runs, verifies, and diagnoses robot-control code (Sec.~\ref{sec:execution_framework}), and a \textbf{Memory-Management} team that distills outcomes into the skill library and failure memory (Sec.~\ref{sec:updates}).

\subsection{Task Proposer Team}
\label{sec:proposer}
To accumulate skills without external rewards, the agent must be driven by intrinsic motivation.
Inspired by developmental psychology, where young children discover robust physical capabilities through unstructured, self-directed play~\cite{oudeyer2007intrinsic}, we formalize ``play" as a bottom-up task-generation process.
Rather than optimizing for externally provided goals, the agent proposes tasks based on direct observations of the scene structure and objects, while drawing on a record of its own prior attempts.
To enable such exploration, the agent's task selection is guided by an intrinsic curiosity reward that favors tasks that are novel yet not too difficult. We design a two-stage task-proposal method to operationalize this principle.

\textbf{Candidate Task Generation.} 
At iteration $t$, we condition the proposer LLM on the current scene context $c_t$, the current learned skill library $\mathcal{L}$, and failure memory $\mathcal{M}$. 
We explicitly prompt the LLM to be exploratory about the environment.
Guided by this playful prior, the LLM generates a candidate pool $\mathcal{T}_t$ of task descriptions.

\textbf{Goldilocks-Driven Task Selection.} 
To select the task $\tau_t$ from the candidate pool $\mathcal{T}_t$, we employ the ``Goldilocks'' principle to target the task that is neither too easy nor too difficult~\cite{baranes2013active, kidd2012goldilocks}. Formally, we evaluate each candidate $\tau$ by maximizing an analytical score: 
$\tau_t = \arg\max_{\tau \in \mathcal{T}_t} \left[ \mathcal{N}(\tau) \cdot \mathcal{F}(\tau) \right]
$
This objective balances exploration and learnability through two components:

(1) \textbf{Object-Skill Novelty $\mathcal{N}(\tau)$:} Defined as $\frac{1}{|O(\tau) \times S(\tau)|} \sum_{(o,s)} \frac{1}{\sqrt{N(o,s) + 1}}$, where $O(\tau)$ and $S(\tau)$ are the objects and required skills, and $N(o,s)$ is their historical attempt count. This term encourages exploration by favoring rarely-tried combinations.  \\(2) \textbf{Competence Frontier $\mathcal{F}(\tau)$:} Formulated as $4\,\bar{r}(\tau)\,\bigl(1-\bar{r}(\tau)\bigr)$, where $\bar{r}(\tau) = \frac{1}{|S(\tau)|}\sum_{s} \hat{r}(s)$ is the average Wilson-bounded empirical success rate $\hat{r}(s)$ of the required skills $s$. This filter peaks at the learnability sweet spot ($\bar{r} \approx 0.5$), encouraging semi-familiar tasks over trivial ($\bar{r} \to 1$) or impossible ($\bar{r} \to 0$) ones.
Once a task is selected, an \textit{Environment Creator} agent builds the corresponding play environment for LIBERO-PRO.
    
\begin{figure}[t]
    \centering
    \includegraphics[width=\linewidth]{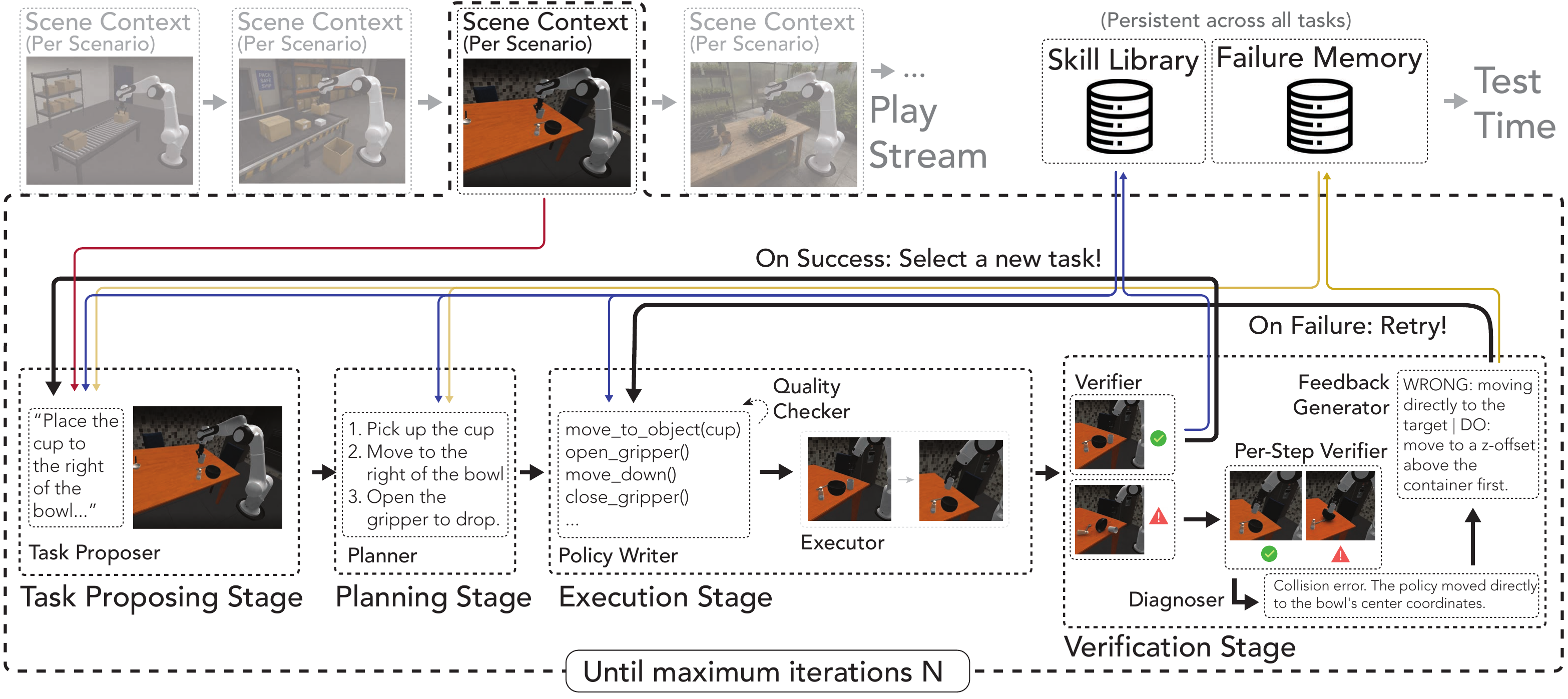}
    \vspace{-1.5em}
\caption{
\textbf{\textsc{RATs} at play.}
\textsc{RATs} proposes self-directed play tasks, solves them with a Code-as-Policy agent team, and uses verification and diagnosis feedback to retry failed attempts. Successful behaviors are distilled into a reusable skill library, which is later retrieved at test time for target tasks.
}
    \label{fig:f2}
    \vspace{-1em}
\end{figure}

\subsection{Execution Team}
\label{sec:execution_framework}

Once a task $\tau_t$ is proposed, \textsc{RATs} executes it through a Code-as-Policy agent team, as shown in Fig.~\ref{fig:f2}. The team is organized into three roles. 
\textbf{Planning Agent.} Given the environment, task description, failure memory, and available skills, the planner produces an ordered plan and annotates each step with relevant retrieved skills. 
\textbf{Execution Agents.} The \textit{Policy Writer} converts the plan into executable robot-control code, which is then run in the environment. When a failure diagnosis identifies a persistent local bottleneck, such as a difficult grasp or articulation, a \textit{SubAgent} can practice the sub-action in isolation and return a reusable helper routine. 
\textbf{Verification Agents.} The verification agents provide feedback at multiple levels: the \textit{Planner Verifier} checks whether the plan is physically grounded in the scene, the \textit{Quality Checker} screens generated code for invalid or unsafe patterns, the \textit{Goal Verifier} evaluates final task success, the \textit{Per-Step Verifier} produces step-level verdicts, and the \textit{Failure Diagnoser} summarizes the failure mode and suggests corrections for retry.

\textbf{Agent Team Orchestration.}
\textsc{RATs} couples these roles into a Write-Execute-Verify-Diagnose loop. The planner first generates a plan, which is checked and refined if it is inconsistent with the initial scene. The team then enters an inner loop with a retry budget. At each attempt, the Policy Writer generates code, the approved code is executed, and the Goal and Per-Step Verifiers evaluate the outcome. If the task succeeds, the loop stops. Otherwise, the Failure Diagnoser returns feedback that guides the next attempt. This feedback allows the system to preserve working substeps, revise faulty code or plans, and spawn a SubAgent when a missing capability should be practiced separately.

\subsection{Memory-Management Team}
\label{sec:updates}

\textsc{RATs} maintains two persistent stores: the skill library $\mathcal{L}$ and the failure memory $\mathcal{M}$. They are updated after each play attempt and periodically curated to keep retrieval useful.

\textbf{Immediate Post-Episode Updates.}
After each task attempt, \textsc{RATs} updates both stores according to the outcome.
On success, the system extracts self-contained, semantically meaningful functions from the successful code, writes docstrings, and inserts them into $\mathcal{L}$ as \emph{experimental} skills.
On failure, it records the failed episode in $\mathcal{M}$ and distills it into a compact lesson, such as a missing precondition or a correction useful for future retries.
Regardless of success, \textsc{RATs} tracks how reliably each invoked skill works over time. Newly added skills start as \emph{experimental}; skills that repeatedly succeed are promoted to \emph{verified} and prioritized in retrieval, while skills that repeatedly fail are marked as \emph{deprecated} and hidden from future planning.

\textbf{Periodic Memory Curation and Skill Proposal.}
Because both stores grow over time, \textsc{RATs} runs maintenance every $K$ iterations (default $K{=}5$). The \textit{Memory Curator} merges or rewrites near-duplicate skills and removes redundant lessons. When recent failures reveal repeated missing capabilities, the \textit{Skill Proposer} drafts candidate helper functions from existing primitives. These proposed helpers enter $\mathcal{L}$ as \emph{experimental} skills and earn reliability through later use.

\subsection{Evaluation With the Learned Skill Library}
\label{sec:test_time}
After the play-time skill development, the system switches to solving externally provided tasks. During the test-time phase, all intrinsic exploration and memory-update mechanisms (\textit{i.e.}, the Task Proposer and Memory Curator) are disabled.
Instead, the system uses its existing knowledge base to solve the tasks. The learned skill library $\mathcal{L}$ can be used in two ways:
(1) \textbf{Plug-and-Play Execution:} To demonstrate the standalone value of the skills acquired during play-time, the frozen library $\mathcal{L}$ can be plugged directly into a standard single-agent Code-as-Policy baseline (\textit{e.g.}, CaP-Agent0).
(2) \textbf{\textsc{RATs} Execution:} The Execution Team is deployed to solve the task with the skill library from play-time. The Planner now draws on the learned skills, allowing the team to bypass low-level physical bottlenecks and compose robust plans for complex tasks.

\providecommand{\liberoTableWidth}{0.6\linewidth}
\providecommand{\molmoTableWidth}{0.65\linewidth}
\providecommand{\robosuiteTableWidth}{0.9\linewidth}
\providecommand{\ablationTableWidth}{0.75\linewidth}
\section{Experimental Results}
\label{sec:result}
We evaluate whether skills learned through autonomous play improve Code-as-Policy agents at test time. 
We report in-domain generalization on LIBERO-PRO and MolmoSpaces (Sec.~\ref{subsec:in-domain-molmo}), cross-environment transfer to RoboSuite~\cite{robosuite2020} (Sec.~\ref{subsec:cross-env-transfer}), ablations isolating the effect of curiosity-driven play (Sec.~\ref{subsec:ablation}), and preliminary real-world transfer (Sec.~\ref{subsec:real-world}).
\begin{table}[t]
    \centering
    \small
    \setlength{\tabcolsep}{4pt}
    \caption{\textbf{LIBERO-PRO in-domain evaluation.} ``Pos.'' corresponds to the
    initial-position swap split, and ``Task'' corresponds to the task
    perturbation split. ``Avg.'' is averaged over all six splits.}
    \label{tab:libero-pro-main}
    \resizebox{\liberoTableWidth}{!}{%
    \begin{tabular}{lccccccc}
        \toprule
        Method
        & \multicolumn{2}{c}{Object}
        & \multicolumn{2}{c}{Goal}
        & \multicolumn{2}{c}{Spatial}
        & Avg. \\
        \cmidrule(lr){2-3}
        \cmidrule(lr){4-5}
        \cmidrule(lr){6-7}
        & Pos. & Task
        & Pos. & Task
        & Pos. & Task
        & \\
        \midrule
        OpenVLA
            & 0.0 & 0.0
            & 0.0 & 0.0
            & 0.0 & 0.0
            & 0.0 \\
        $\pi_0$
            & 0.0 & 0.0
            & 0.0 & 0.0
            & 0.0 & 0.0
            & 0.0 \\
        $\pi_{0.5}$
            & 17.0 & 1.0
            & 38.0 & 0.0
            & 20.0 & 1.0
            & 12.8 \\
        \midrule
        \textsc{CaP-Agent0}
            & 27.0 & 31.0
            & 29.0 & 16.0
            & 13.0 & 23.0
            & 23.2 \\
        \textsc{RATs (Ours)}
            & \textbf{61.0} & \textbf{63.0}
            & \textbf{43.0} & \textbf{36.0}
            & \textbf{29.0} & \textbf{31.0}
            & \textbf{43.8} \\
        \bottomrule
    \end{tabular}
    }
    \vspace{-0.5em}
\end{table}

\begin{table}[t]
    \centering
    \small
    \setlength{\tabcolsep}{7pt}
    \caption{\textbf{MolmoSpaces in-domain evaluation.} Each category has 10
    tasks, with 10 trials per task.}
    \label{tab:molmospace-main}
    \resizebox{\molmoTableWidth}{!}{%
    \begin{tabular}{lccccc}
        \toprule
        Method
        & Open
        & Close
        & Pick
        & Pick-and-Place
        & Avg. \\
        \midrule
        \textsc{CaP-Agent0}
            & 14.0 & 36.0 & 23.0 & 11.0 & 21.0  \\
        \textsc{RATs (Ours)}
            & \textbf{20.0} & \textbf{73.0} & \textbf{37.0} & \textbf{22.0} & \textbf{38.0} \\
        \bottomrule
    \end{tabular}
    
    }
    \vspace{-1.em}
\end{table}

\begin{figure}[t]
    \vspace{-2.5em}
    \includegraphics[width=0.95\linewidth]{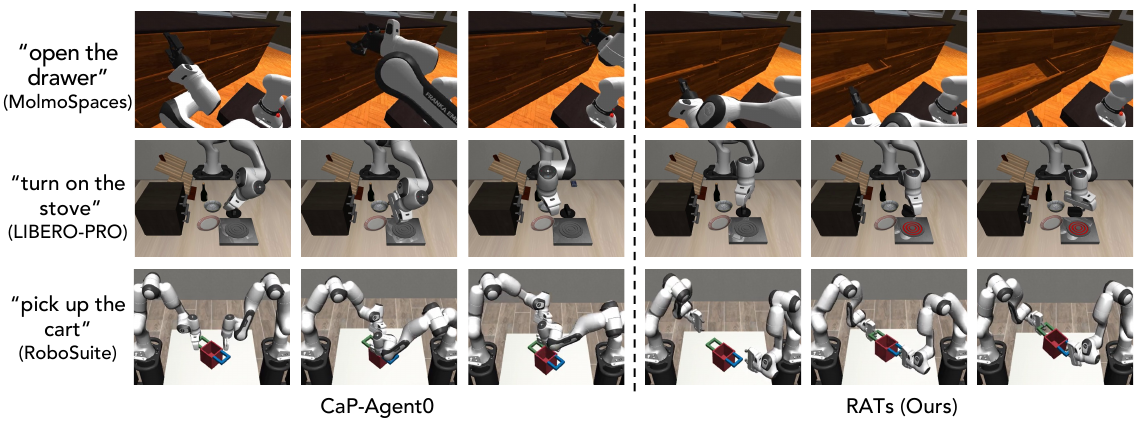}
    \vspace{-0.75em}
\caption{
\textbf{Qualitative comparisons in simulation.}
}
    \label{fig:f3}
    \vspace{-0.75em}
\end{figure}
\begin{figure}[t]
    \includegraphics[width=0.98\linewidth]{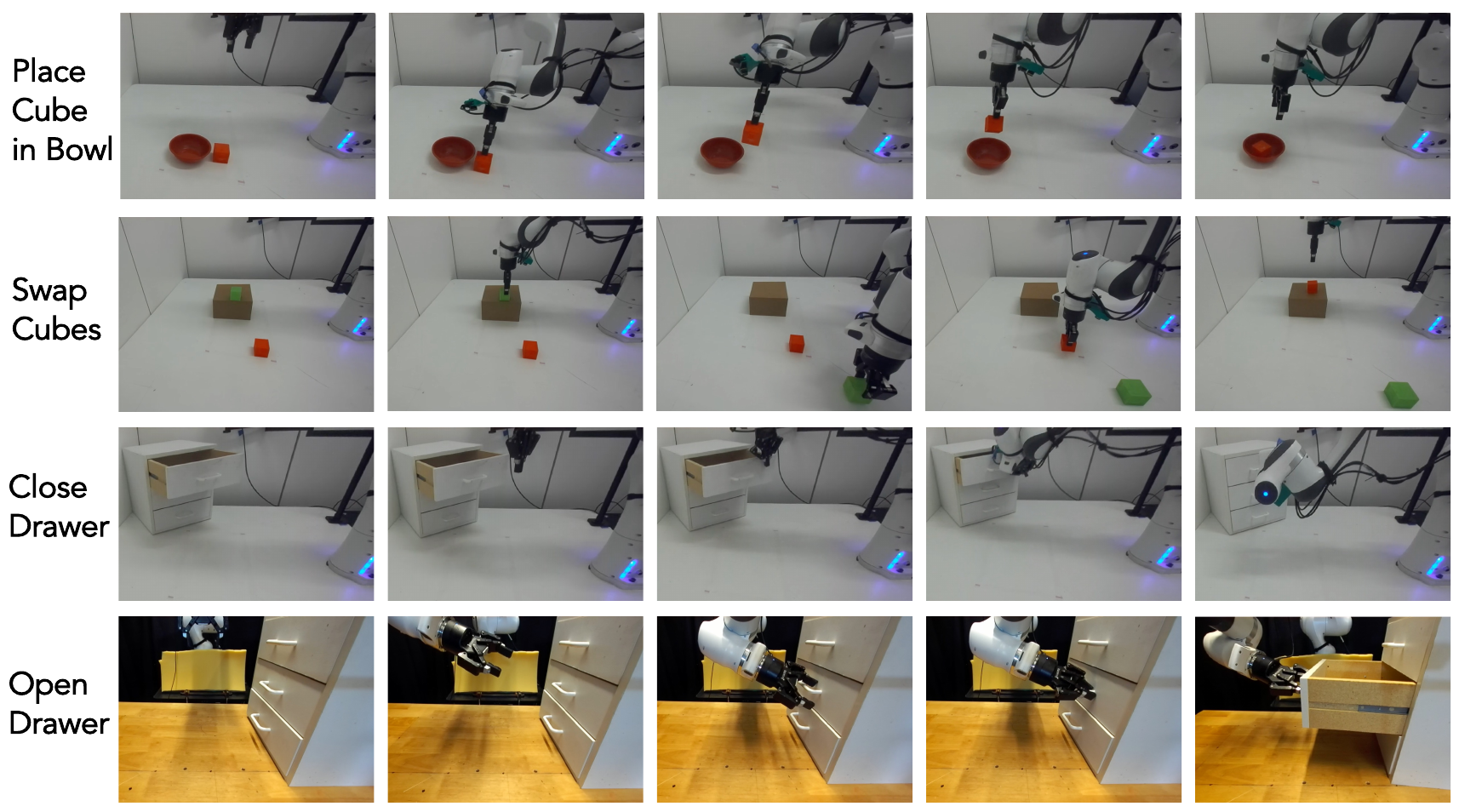}
    \vspace{-0.5em}
\caption{
\textbf{Qualitative results of sim-to-real transfer.}
}
    \label{fig:real}
    \vspace{-1em}
\end{figure}
\subsection{Experiment Details}
\label{subsec:exp-details}
\textbf{Benchmarks.} We use three manipulation suites. \emph{LIBERO-PRO}~\cite{liu2023libero,liberopro}
tests Object, Goal, and Spatial generalization tasks. 
It perturbs each task along three axes (object, goal, spatial), each
with an initial-position swap (``Pos.'') and a task perturbation (``Task'').
\emph{MolmoSpaces}~\cite{kim2026molmospaces} tests Open, Close, Pick, and Pick \& Place tasks.
It scores success from grounded scene state and natural-language criteria, complementing the predicate-based tasks of LIBERO-PRO. 
Both LIBERO-PRO and MolmoSpaces serve as play environments and in-domain evaluation benchmarks; we describe both benchmarks in detail in Appendix~\ref{sec:appendix_evaluation_benchmarks}.
\emph{RoboSuite} is a separate simulator, held out from play, used only for
cross-environment transfer.

\textbf{Methods.} We compare against VLA policies, OpenVLA~\cite{openvla},
$\pi_0$~\citep{black2025pi0}, and
$\pi_{0.5}$~\citep{physicalintelligence2025pi05}, and against
\emph{CaP-Agent0}~\cite{fu2026capx}, the Code-as-Policy agent run with only the primitive library $\mathcal{L}_0$ (the ``No Play'' condition). For \textsc{RATs}, we play on LIBERO-PRO and MolmoSpaces environment for 50 iterations each with \texttt{gemini-3.1pro-preview}. See Appendix~\ref{sec:appendix_agent_team} for more details.

\textbf{Metric and Evaluation Modes.} We report task success rate and evaluate the
learned library in two modes: a \emph{plug-in} mode that retrieves skills into CaP-Agent0's context, and a \emph{full-system} mode (\textsc{RATs} Exec.) that runs the complete task execution system.

\subsection{In-Domain Evaluation}
\label{subsec:in-domain-molmo}

On LIBERO-PRO, we evaluate on 60 held-out tasks under 10 initializations each, for a total of 600 trials.
Table~\ref{tab:libero-pro-main} shows \textsc{RATs} raises average success from 23.2\% (CaP-Agent0) to 43.8\% (+20.6 pp) and outperforms all VLA baselines, the best of which ($\pi_{0.5}$) reaches 12.8\%. Gains are largest on the object splits
(61.0\% and 63.0\%) but also appear on the goal and spatial splits, indicating
that the play-learned skills generalize beyond the tasks practiced during play.
For MolmoSpaces, we sample 10 tasks from each of four task categories (opening, closing, picking, pick and place) and run 10 trials each, for a total of 400 trials.
Table~\ref{tab:molmospace-main} shows \textsc{RATs} improves average success from 21.0\% to 38.0\% and improves significantly on all categories. Refer to Fig.~\ref{fig:f3} for the qualitative results.

\subsection{Cross-Environment Transfer}
\label{subsec:cross-env-transfer}
We test whether skills learned in one simulation environment transfer to another. Specifically, we plug the skill library learned from LIBERO-PRO play into CaP-Agent0 and evaluate it on RoboSuite, which is never seen during play.
We use five randomizations $\times$ 10 trials
(50 trials) per task.
Table~\ref{tab:robosuite-transfer} shows that transferred skills raise average success from 40.3\% to 49.1\% (+8.9 pp), with broad gains on cube lifting (+16.0 pp) and two-arm lifting (+24.0 pp). Notably, this largest gain is \textit{cross-embodiment}: although the skills are practiced on the single-arm LIBERO-PRO robot, they still transfer to this two-arm collaborative task.

\begin{table}[t]
    \centering
    \small
    \vspace{-2em}
    \setlength{\tabcolsep}{7pt}
    \caption{\textbf{Cross-environment transfer on RoboSuite and preliminary real-world evaluation.}
    Skills are learned by \textsc{RATs} during LIBERO-PRO play and reused with the CaP-Agent0 system. Each RoboSuite task has 50 trials,
    and each real-world task has 40 trials.}
    \label{tab:robosuite-transfer}
    \resizebox{\robosuiteTableWidth}{!}{%
    {\renewcommand{\arraystretch}{1.05}
    \begin{tabular}{lccc}
        \toprule
        Task
        & \textsc{CaP-Agent0}
        & \textsc{CaP-Agent0 + \textsc{RATs} Skills (LIBERO-PRO)}
        & $\Delta$ \\
        \midrule
        Cube lifting
            & 34/50 (68.0\%)
            & \textbf{42/50 (84.0\%)}
            & +16.0 pp \\
        Cube restacking
            & 17/50 (34.0\%)
            & \textbf{23/50 (46.0\%)}
            & +12.0 pp \\
        Cube stacking
            & 23/50 (46.0\%)
            & \textbf{30/50 (60.0\%)}
            & +14.0 pp \\
        Nut assembly
            & 0/50 (0.0\%)
            & 0/50 (0.0\%)
            & 0.0 pp \\
        Spill wiping
            & 50/50 (100.0\%)
            & 50/50 (100.0\%)
            & 0.0 pp \\
        Two-arm handover
            & \textbf{12/50 (24.0\%)}
            & 10/50 (20.0\%)
            & -4.0 pp \\
        Two-arm lifting
            & 5/50 (10.0\%)
            & \textbf{17/50 (34.0\%)}
            & +24.0 pp \\
        \midrule
        \textbf{Average (RoboSuite)}
            & 141/350 (40.3\%)
            & \textbf{172/350 (49.1\%)}
            & +8.9 pp \\
        \midrule
        Pick up red cube
            & 14/40 (35.0\%)
            & \textbf{17/40 (42.5\%)}
            & +7.5 pp \\
        Place cube in bowl
            & 10/40 (25.0\%)
            & \textbf{14/40 (35.0\%)}
            & +10.0 pp \\
        \midrule
        \textbf{Average (Real World)}
            & 24/80 (30.0\%)
            & \textbf{31/80 (38.8\%)}
            & +8.8 pp \\
        \bottomrule
    \end{tabular}
    }
    \vspace{-2em}
    }
\end{table}

\subsection{Ablating Play Strategy and Test-Time Execution}
\label{subsec:ablation}
To understand the performance gains, we ablate our system on LIBERO-PRO along two dimensions:
(1) Play Strategy: We compare \emph{No Play} (primitives only), \emph{Random Play} (50 iterations of randomly sampled tasks), and \emph{Curious Play} (50 iterations of play using our task proposer).
(2) Test-Time System: We evaluate the learned library using either a standard \emph{CaP-Agent0} baseline or our full \textsc{RATs} system.
Table~\ref{tab:libero-ablation} shows two findings:
First, curiosity is essential for effective play: under CaP-Agent0, \emph{Random Play} provides little gain over \emph{No Play} (23.2\% $\to$ 24.7\%), while \emph{Curious Play} improves performance to 32.3\%.
Second, play and test-time execution are complementary: improving only execution raises performance from 23.2\% to 36.3\%, while improving only play raises it to 32.3\%; combining both yields the best, 44.3\%. See Appendix~\ref{sec:appendix_play_process} for more play details. We report additional ablations on MolmoSpaces and a token-cost analysis in Appendix~\ref{sec:appendix_additional_studies}.

\begin{table}[t]
    \centering
    \scriptsize
    \setlength{\tabcolsep}{3.5pt}
    {\renewcommand{\arraystretch}{0.95}
    \caption{\textbf{LIBERO-PRO ablation over play strategy and test-time system.} All
    play-based variants use 50 play iterations. All play-time skills are learned by our proposed \textsc{RATs} system. For the \textsc{RATs} Exec. test-time system, we use 5 trials per task rather than 10.}
    \label{tab:libero-ablation}
    \resizebox{\ablationTableWidth}{!}{%
    \begin{tabular}{llccccccc}
        \toprule
        \multirow{2}{*}{Test-Time System}
        & \multirow{2}{*}{Play-Time Skills}
        & \multicolumn{2}{c}{Object}
        & \multicolumn{2}{c}{Goal}
        & \multicolumn{2}{c}{Spatial}
        & Avg. \\
        \cmidrule(lr){3-4}
        \cmidrule(lr){5-6}
        \cmidrule(lr){7-8}
        & & Pos. & Task & Pos. & Task & Pos. & Task & \\
        \midrule
        \multirow{3}{*}{\textsc{CaP-Agent0}}
            & No Play
            & 27.0 & 31.0 & 29.0 & 16.0 & 13.0 & 23.0 & 23.2 \\
        & Random Play
            & 20.0 & 28.0 & 32.0 & 16.0 & 20.0 & 32.0 & 24.7 \\
        & Curious Play
            & 51.0 & 47.0 & 34.0 & 20.0 & 19.0 & 23.0 & 32.3 \\
        \midrule
        \multirow{3}{*}{\textsc{RATs Exec.}}
            & No Play
            & 54.0 & 58.0 & 32.0 & 24.0 & 20.0 & \textbf{30.0} & 36.3 \\
        & Random Play
            & 54.0 & 46.0 & 34.0 & 44.0 & 24.0 & 28.0 & 38.3 \\
        & Curious Play
            & \textbf{60.0} & \textbf{60.0}
            & \textbf{48.0} & \textbf{38.0}
            & \textbf{30.0} & \textbf{30.0}
            & \textbf{44.3} \\
        \bottomrule
    \end{tabular}
    }
    }
    \vspace{-1.5em}
\end{table}

\subsection{Sim-to-Real Evaluation}
\label{subsec:real-world}

Finally, we evaluate whether the skill library acquired during simulated play can be reused on a physical robot. We export the skill library learned after 50 iterations of play in LIBERO-PRO and deploy it directly, without real-world play or finetuning. Given real-world manipulation instructions such as cube picking and placing, the system retrieves relevant skills from the library and executes policies through the robot's control APIs. As in Table~\ref{tab:robosuite-transfer} (bottom), adding \textsc{RATs}-learned skills improves over the \textsc{CaP-Agent0} baseline by 8.8 percentage points on this real-world task set. These results suggest that play-learned code skills can be reused on real hardware for simple manipulation tasks. We show demonstrations in Figure~\ref{fig:real}, with additional details in Appendix~\ref{sec:appendix_additional_realworld}.

\section{Conclusion}
\label{sec:conclusion}

We introduced \textsc{RATs}, a playful agentic robot learning framework that
acquires reusable Code-as-Policy skills before downstream tasks are given.
Across LIBERO-PRO and MolmoSpaces, \textsc{RATs} substantially improves over CaP-Agent0 and VLA
baselines, and ablations show that random play under the same budget produces
much smaller gains. Cross-environment results on RoboSuite and real-world evaluations further suggest that
skills learned through self-proposed play can transfer beyond the original play
environment. Together, these results support playful skill acquisition as a
practical way to improve agentic robot systems without finetuning the underlying
model.

\section*{Limitations}
While \textsc{RATs} is an initial step toward playful agentic robot learning, its evaluation remains primarily simulation-based, requiring larger-scale physical deployment to validate robust sim-to-real transfer. Moreover, play-time skill acquisition is constrained by the diversity of available simulation environments, limiting the range of objects, dynamics, and affordances the agent can practice. Additionally, improper skill reuse can hurt performance when retrieved skills do not fit the downstream task, showing the need for better retrieval and context-aware selection. Finally, \textsc{RATs} increases inference costs, relies heavily on VLM verification, and remains bounded by primitive-level control APIs, which limits dexterous manipulation and motivates lighter feedback mechanisms and richer low-level controllers for real-world scaling.

\section*{Acknowledgments}
We thank Baifeng Shi, XuDong Wang, and Jianyuan Wang for helpful feedback. We thank Simeon Adebola for the help during the real-world evaluation. The UC Berkeley authors acknowledge support from the BAIR Commons Humanoid Intelligence Center.

\bibliography{example}  %

\begin{thebibliography}{61}
\providecommand{\natexlab}[1]{#1}
\providecommand{\url}[1]{\texttt{#1}}
\expandafter\ifx\csname urlstyle\endcsname\relax
  \providecommand{\doi}[1]{doi: #1}\else
  \providecommand{\doi}{doi: \begingroup \urlstyle{rm}\Url}\fi

\bibitem[Liang et~al.(2023)Liang, Huang, Xia, Xu, Hausman, Ichter, Florence, and Zeng]{liang2023code}
J.~Liang, W.~Huang, F.~Xia, P.~Xu, K.~Hausman, B.~Ichter, P.~Florence, and A.~Zeng.
\newblock Code as policies: Language model programs for embodied control.
\newblock In \emph{2023 IEEE International Conference on Robotics and Automation (ICRA)}, pages 9493--9500. IEEE, 2023.

\bibitem[Singh et~al.(2023)Singh, Blukis, Mousavian, Goyal, Xu, Tremblay, Fox, Thomason, and Garg]{singh2023progprompt}
I.~Singh, V.~Blukis, A.~Mousavian, A.~Goyal, D.~Xu, J.~Tremblay, D.~Fox, J.~Thomason, and A.~Garg.
\newblock {ProgPrompt}: Generating situated robot task plans using large language models.
\newblock In \emph{2023 IEEE International Conference on Robotics and Automation (ICRA)}, pages 11523--11530. IEEE, 2023.

\bibitem[Vemprala et~al.(2024)Vemprala, Bonatti, Bucker, and Kapoor]{vemprala2024chatgpt}
S.~Vemprala, R.~Bonatti, A.~Bucker, and A.~Kapoor.
\newblock {ChatGPT} for robotics: Design principles and model abilities.
\newblock \emph{IEEE Access}, 12:\penalty0 55682--55696, 2024.

\bibitem[Zitkovich et~al.(2023)Zitkovich, Yu, Xu, Xu, Xiao, Xia, Wu, Wohlhart, Welker, Wahid, Vuong, Vanhoucke, Tran, Soricut, Singh, Singh, Sermanet, Sanketi, Salazar, Ryoo, Reymann, Rao, Pertsch, Mordatch, Michalewski, Lu, Levine, Lee, Lee, Leal, Kuang, Kalashnikov, Julian, Joshi, Irpan, Ichter, Hsu, Herzog, Hausman, Gopalakrishnan, Fu, Florence, Finn, Dubey, Driess, Ding, Choromanski, Chen, Chebotar, Carbajal, Brown, Brohan, Arenas, and Han]{zitkovich2023rt2}
B.~Zitkovich, T.~Yu, S.~Xu, P.~Xu, T.~Xiao, F.~Xia, J.~Wu, P.~Wohlhart, S.~Welker, A.~Wahid, Q.~Vuong, V.~Vanhoucke, H.~Tran, R.~Soricut, A.~Singh, J.~Singh, P.~Sermanet, P.~R. Sanketi, G.~Salazar, M.~S. Ryoo, K.~Reymann, K.~Rao, K.~Pertsch, I.~Mordatch, H.~Michalewski, Y.~Lu, S.~Levine, L.~Lee, T.-W.~E. Lee, I.~Leal, Y.~Kuang, D.~Kalashnikov, R.~Julian, N.~J. Joshi, A.~Irpan, B.~Ichter, J.~Hsu, A.~Herzog, K.~Hausman, K.~Gopalakrishnan, C.~Fu, P.~Florence, C.~Finn, K.~A. Dubey, D.~Driess, T.~Ding, K.~M. Choromanski, X.~Chen, Y.~Chebotar, J.~Carbajal, N.~Brown, A.~Brohan, M.~G. Arenas, and K.~Han.
\newblock {RT-2}: Vision-language-action models transfer web knowledge to robotic control.
\newblock In \emph{Proceedings of The 7th Conference on Robot Learning}, volume 229 of \emph{Proceedings of Machine Learning Research}, pages 2165--2183. PMLR, 2023.
\newblock URL \url{https://proceedings.mlr.press/v229/zitkovich23a.html}.

\bibitem[Black et~al.(2025)Black, Brown, Driess, Esmail, Equi, Finn, Fusai, Groom, Hausman, Ichter, Jakubczak, Jones, Ke, Levine, Li-Bell, Mothukuri, Nair, Pertsch, Shi, Smith, Tanner, Vuong, Walling, Wang, and Zhilinsky]{black2025pi0}
K.~Black, N.~Brown, D.~Driess, A.~Esmail, M.~R. Equi, C.~Finn, N.~Fusai, L.~Groom, K.~Hausman, B.~Ichter, S.~Jakubczak, T.~Jones, L.~Ke, S.~Levine, A.~Li-Bell, M.~Mothukuri, S.~Nair, K.~Pertsch, L.~X. Shi, L.~Smith, J.~Tanner, Q.~Vuong, A.~Walling, H.~Wang, and U.~Zhilinsky.
\newblock {$\pi_0$}: A vision-language-action flow model for general robot control.
\newblock In \emph{Proceedings of Robotics: Science and Systems (RSS)}, Los Angeles, CA, USA, 2025.

\bibitem[{Physical Intelligence} et~al.(2025){Physical Intelligence}, Black, Brown, Darpinian, Dhabalia, Driess, Esmail, Equi, Finn, Fusai, Galliker, Ghosh, Groom, Hausman, Ichter, Jakubczak, Jones, Ke, LeBlanc, Levine, Li-Bell, Mothukuri, Nair, Pertsch, Ren, Shi, Smith, Springenberg, Stachowicz, Tanner, Vuong, Walke, Walling, Wang, Yu, and Zhilinsky]{physicalintelligence2025pi05}
{Physical Intelligence}, K.~Black, N.~Brown, J.~Darpinian, K.~Dhabalia, D.~Driess, A.~Esmail, M.~R. Equi, C.~Finn, N.~Fusai, M.~Y. Galliker, D.~Ghosh, L.~Groom, K.~Hausman, B.~Ichter, S.~Jakubczak, T.~Jones, L.~Ke, D.~LeBlanc, S.~Levine, A.~Li-Bell, M.~Mothukuri, S.~Nair, K.~Pertsch, A.~Z. Ren, L.~X. Shi, L.~Smith, J.~T. Springenberg, K.~Stachowicz, J.~Tanner, Q.~Vuong, H.~Walke, A.~Walling, H.~Wang, L.~Yu, and U.~Zhilinsky.
\newblock {$\pi_{0.5}$}: A vision-language-action model with open-world generalization.
\newblock In \emph{Proceedings of the 9th Conference on Robot Learning (CoRL)}, Proceedings of Machine Learning Research. PMLR, 2025.

\bibitem[Fu et~al.(2026)Fu, Yu, El-Refai, Kou, Xue, Huang, Xiao, Wang, Li, Shi, Wu, Sastry, Zhu, Goldberg, and Fan]{fu2026capx}
M.~Fu, J.~Yu, K.~El-Refai, E.~Kou, H.~Xue, H.~Huang, W.~Xiao, G.~Wang, F.-F. Li, G.~Shi, J.~Wu, S.~Sastry, Y.~Zhu, K.~Goldberg, and L.~Fan.
\newblock {CaP-X}: A framework for benchmarking and improving coding agents for robot manipulation.
\newblock \emph{arXiv preprint arXiv:2603.22435}, 2026.

\bibitem[Piaget(1952)]{piaget1952origins}
J.~Piaget.
\newblock \emph{The Origins of Intelligence in Children}.
\newblock International Universities Press, New York, 1952.

\bibitem[Gopnik(2020)]{gopnik2020childhood}
A.~Gopnik.
\newblock Childhood as a solution to explore-exploit tensions.
\newblock \emph{Philosophical Transactions of the Royal Society B: Biological Sciences}, 375\penalty0 (1803):\penalty0 20190502, 2020.
\newblock \doi{10.1098/rstb.2019.0502}.

\bibitem[Smith and Gasser(2005)]{smith2005developmental}
L.~B. Smith and M.~Gasser.
\newblock The development of embodied cognition: Six lessons from babies.
\newblock \emph{Artificial Life}, 11\penalty0 (1-2):\penalty0 13--29, 2005.
\newblock \doi{10.1162/1064546053278973}.

\bibitem[Schmidhuber(1991)]{schmidhuber1991curious}
J.~Schmidhuber.
\newblock Curious model-building control systems.
\newblock In \emph{Proceedings of the International Joint Conference on Neural Networks (IJCNN)}, volume~2, pages 1458--1463, Singapore, 1991. IEEE.

\bibitem[Oudeyer et~al.(2007)Oudeyer, Kaplan, and Hafner]{oudeyer2007intrinsic}
P.-Y. Oudeyer, F.~Kaplan, and V.~V. Hafner.
\newblock Intrinsic motivation systems for autonomous mental development.
\newblock \emph{IEEE Transactions on Evolutionary Computation}, 11\penalty0 (2):\penalty0 265--286, 2007.
\newblock \doi{10.1109/TEVC.2006.890271}.

\bibitem[Baranes and Oudeyer(2013)]{baranes2013active}
A.~Baranes and P.-Y. Oudeyer.
\newblock Active learning of inverse models with intrinsically motivated goal exploration in robots.
\newblock \emph{Robotics and Autonomous Systems}, 61\penalty0 (1):\penalty0 49--73, 2013.

\bibitem[Forestier and Oudeyer(2016)]{forestier2016modular}
S.~Forestier and P.-Y. Oudeyer.
\newblock Modular active curiosity-driven discovery of tool use.
\newblock In \emph{2016 IEEE/RSJ International Conference on Intelligent Robots and Systems (IROS)}, pages 3965--3972, Daejeon, Korea, 2016. IEEE.

\bibitem[Pathak et~al.(2017)Pathak, Agrawal, Efros, and Darrell]{pathak2017curiosity}
D.~Pathak, P.~Agrawal, A.~A. Efros, and T.~Darrell.
\newblock Curiosity-driven exploration by self-supervised prediction.
\newblock In \emph{Proceedings of the 34th International Conference on Machine Learning (ICML)}, volume~70 of \emph{Proceedings of Machine Learning Research}, pages 2778--2787, 2017.

\bibitem[Houthooft et~al.(2016)Houthooft, Chen, Duan, Schulman, De~Turck, and Abbeel]{houthooft2016vime}
R.~Houthooft, X.~Chen, Y.~Duan, J.~Schulman, F.~De~Turck, and P.~Abbeel.
\newblock {VIME}: Variational information maximizing exploration.
\newblock In \emph{Advances in Neural Information Processing Systems 29 (NeurIPS)}, 2016.

\bibitem[Vygotsky(1978)]{vygotsky1978mind}
L.~S. Vygotsky.
\newblock \emph{Mind in Society: The Development of Higher Psychological Processes}.
\newblock Harvard University Press, Cambridge, MA, 1978.
\newblock Edited by Michael Cole, Vera John-Steiner, Sylvia Scribner, and Ellen Souberman.

\bibitem[Pellegrini(2009)]{pellegrini2009role}
A.~D. Pellegrini.
\newblock \emph{The Role of Play in Human Development}.
\newblock Oxford University Press, Oxford, UK, 2009.
\newblock ISBN 9780195367324.

\bibitem[Schmidhuber(2010)]{schmidhuber2010formal}
J.~Schmidhuber.
\newblock Formal theory of creativity, fun, and intrinsic motivation (1990--2010).
\newblock \emph{IEEE Transactions on Autonomous Mental Development}, 2\penalty0 (3):\penalty0 230--247, 2010.

\bibitem[Kaplan and Oudeyer(2007)]{kaplan2007intrinsic}
F.~Kaplan and P.-Y. Oudeyer.
\newblock In search of the neural circuits of intrinsic motivation.
\newblock \emph{Frontiers in Neuroscience}, 1\penalty0 (1):\penalty0 225--236, 2007.
\newblock \doi{10.3389/neuro.01.1.1.017.2007}.

\bibitem[Kidd et~al.(2012)Kidd, Piantadosi, and Aslin]{kidd2012goldilocks}
C.~Kidd, S.~T. Piantadosi, and R.~N. Aslin.
\newblock The goldilocks effect: Human infants allocate attention to visual sequences that are neither too simple nor too complex.
\newblock \emph{PLOS ONE}, 7\penalty0 (5):\penalty0 e36399, 2012.
\newblock \doi{10.1371/journal.pone.0036399}.

\bibitem[Rolf et~al.(2010)Rolf, Steil, and Gienger]{rolf2010goal}
M.~Rolf, J.~J. Steil, and M.~Gienger.
\newblock Goal babbling permits direct learning of inverse kinematics.
\newblock \emph{IEEE Transactions on Autonomous Mental Development}, 2\penalty0 (3):\penalty0 216--229, 2010.
\newblock \doi{10.1109/TAMD.2010.2062511}.

\bibitem[Forestier et~al.(2022)Forestier, Portelas, Mollard, and Oudeyer]{forestier2017intrinsically}
S.~Forestier, R.~Portelas, Y.~Mollard, and P.-Y. Oudeyer.
\newblock Intrinsically motivated goal exploration processes with automatic curriculum learning.
\newblock \emph{Journal of Machine Learning Research}, 23\penalty0 (152):\penalty0 1--41, 2022.

\bibitem[Lynch et~al.(2020)Lynch, Khansari, Xiao, Kumar, Tompson, Levine, and Sermanet]{lynch2020learning}
C.~Lynch, M.~Khansari, T.~Xiao, V.~Kumar, J.~Tompson, S.~Levine, and P.~Sermanet.
\newblock Learning latent plans from play.
\newblock In L.~P. Kaelbling, D.~Kragic, and K.~Sugiura, editors, \emph{Proceedings of the Conference on Robot Learning}, volume 100 of \emph{Proceedings of Machine Learning Research}, pages 1113--1132. PMLR, 30 Oct--01 Nov 2020.
\newblock URL \url{https://proceedings.mlr.press/v100/lynch20a.html}.

\bibitem[Wang et~al.(2023)Wang, Fan, Sun, Zhang, Fei-Fei, Xu, Zhu, and Anandkumar]{wang2023mimicplay}
C.~Wang, L.~Fan, J.~Sun, R.~Zhang, L.~Fei-Fei, D.~Xu, Y.~Zhu, and A.~Anandkumar.
\newblock Mimicplay: Long-horizon imitation learning by watching human play.
\newblock In J.~Tan, M.~Toussaint, and K.~Darvish, editors, \emph{Proceedings of The 7th Conference on Robot Learning}, volume 229 of \emph{Proceedings of Machine Learning Research}, pages 201--221. PMLR, 06--09 Nov 2023.
\newblock URL \url{https://proceedings.mlr.press/v229/wang23a.html}.

\bibitem[Colas et~al.(2020)Colas, Karch, Lair, Dussoux, Moulin-Frier, Dominey, and Oudeyer]{colas2020language}
C.~Colas, T.~Karch, N.~Lair, J.-M. Dussoux, C.~Moulin-Frier, P.~F. Dominey, and P.-Y. Oudeyer.
\newblock Language as a cognitive tool to imagine goals in curiosity-driven exploration.
\newblock In \emph{Advances in Neural Information Processing Systems 33 (NeurIPS)}, 2020.

\bibitem[Thrun and Mitchell(1995)]{thrun1995lifelong}
S.~Thrun and T.~M. Mitchell.
\newblock Lifelong robot learning.
\newblock \emph{Robotics and Autonomous Systems}, 15\penalty0 (1--2):\penalty0 25--46, 1995.

\bibitem[Parisi et~al.(2019)Parisi, Kemker, Part, Kanan, and Wermter]{parisi2019continual}
G.~I. Parisi, R.~Kemker, J.~L. Part, C.~Kanan, and S.~Wermter.
\newblock Continual lifelong learning with neural networks: A review.
\newblock \emph{Neural Networks}, 113:\penalty0 54--71, 2019.
\newblock \doi{10.1016/j.neunet.2019.01.012}.

\bibitem[Lesort et~al.(2020)Lesort, Lomonaco, Stoian, Maltoni, Filliat, and D{\'i}az-Rodr{\'i}guez]{lesort2020continual}
T.~Lesort, V.~Lomonaco, A.~Stoian, D.~Maltoni, D.~Filliat, and N.~D{\'i}az-Rodr{\'i}guez.
\newblock Continual learning for robotics: Definition, framework, learning strategies, opportunities and challenges.
\newblock \emph{Information Fusion}, 58:\penalty0 52--68, 2020.

\bibitem[Sutton et~al.(1999)Sutton, Precup, and Singh]{sutton1999between}
R.~S. Sutton, D.~Precup, and S.~Singh.
\newblock Between {MDPs} and semi-{MDPs}: A framework for temporal abstraction in reinforcement learning.
\newblock \emph{Artificial Intelligence}, 112\penalty0 (1--2):\penalty0 181--211, 1999.

\bibitem[Konidaris et~al.(2012)Konidaris, Kuindersma, Grupen, and Barto]{konidaris2012robot}
G.~Konidaris, S.~Kuindersma, R.~Grupen, and A.~Barto.
\newblock Robot learning from demonstration by constructing skill trees.
\newblock \emph{The International Journal of Robotics Research}, 31\penalty0 (3):\penalty0 360--375, 2012.

\bibitem[Kroemer et~al.(2021)Kroemer, Niekum, and Konidaris]{kroemer2021review}
O.~Kroemer, S.~Niekum, and G.~Konidaris.
\newblock A review of robot learning for manipulation: Challenges, representations, and algorithms.
\newblock \emph{Journal of Machine Learning Research}, 22\penalty0 (30):\penalty0 1--82, 2021.

\bibitem[Pertsch et~al.(2020)Pertsch, Lee, and Lim]{pertsch2020long}
K.~Pertsch, Y.~Lee, and J.~J. Lim.
\newblock Accelerating reinforcement learning with learned skill priors.
\newblock In \emph{Proceedings of the 4th Conference on Robot Learning (CoRL)}, volume 155 of \emph{Proceedings of Machine Learning Research}, pages 188--204, 2020.

\bibitem[Lynch and Sermanet(2021)]{lynch2021language}
C.~Lynch and P.~Sermanet.
\newblock Language conditioned imitation learning over unstructured data.
\newblock In \emph{Proceedings of Robotics: Science and Systems (RSS)}, 2021.
\newblock \doi{10.15607/RSS.2021.XVII.047}.

\bibitem[Wan et~al.(2024)Wan, Zhu, Shah, and Zhu]{wan2024lotuscontinualimitationlearning}
W.~Wan, Y.~Zhu, R.~Shah, and Y.~Zhu.
\newblock Lotus: Continual imitation learning for robot manipulation through unsupervised skill discovery, 2024.
\newblock URL \url{https://arxiv.org/abs/2311.02058}.

\bibitem[Ma et~al.(2024)Ma, Liang, Wang, Wang, Zhu, Fan, Bastani, and Jayaraman]{ma2024dreureka}
Y.~J. Ma, W.~Liang, H.-J. Wang, S.~Wang, Y.~Zhu, L.~Fan, O.~Bastani, and D.~Jayaraman.
\newblock {DrEureka}: Language model guided sim-to-real transfer.
\newblock In \emph{Robotics: Science and Systems (RSS)}, 2024.
\newblock URL \url{https://arxiv.org/abs/2406.01967}.

\bibitem[Bengio et~al.(2009)Bengio, Louradour, Collobert, and Weston]{bengio2009curriculum}
Y.~Bengio, J.~Louradour, R.~Collobert, and J.~Weston.
\newblock Curriculum learning.
\newblock In \emph{Proceedings of the 26th Annual International Conference on Machine Learning (ICML)}, pages 41--48. ACM, 2009.

\bibitem[Narvekar et~al.(2020)Narvekar, Peng, Leonetti, Sinapov, Taylor, and Stone]{narvekar2020curriculum}
S.~Narvekar, B.~Peng, M.~Leonetti, J.~Sinapov, M.~E. Taylor, and P.~Stone.
\newblock Curriculum learning for reinforcement learning domains: A framework and survey.
\newblock \emph{Journal of Machine Learning Research}, 21\penalty0 (181):\penalty0 1--50, 2020.

\bibitem[Florensa et~al.(2018)Florensa, Held, Geng, and Abbeel]{florensa2018automatic}
C.~Florensa, D.~Held, X.~Geng, and P.~Abbeel.
\newblock Automatic goal generation for reinforcement learning agents.
\newblock In \emph{Proceedings of the 35th International Conference on Machine Learning (ICML)}, volume~80 of \emph{Proceedings of Machine Learning Research}, pages 1515--1528, 2018.

\bibitem[Nair et~al.(2018)Nair, Pong, Dalal, Bahl, Lin, and Levine]{nair2018visual}
A.~Nair, V.~Pong, M.~Dalal, S.~Bahl, S.~Lin, and S.~Levine.
\newblock Visual reinforcement learning with imagined goals.
\newblock In \emph{Advances in Neural Information Processing Systems 31 (NeurIPS)}, 2018.

\bibitem[Pong et~al.(2020)Pong, Dalal, Lin, Nair, Bahl, and Levine]{pong2020skewfit}
V.~H. Pong, M.~Dalal, S.~Lin, A.~Nair, S.~Bahl, and S.~Levine.
\newblock {Skew-Fit}: State-covering self-supervised reinforcement learning.
\newblock In \emph{Proceedings of the 37th International Conference on Machine Learning (ICML)}, volume 119 of \emph{Proceedings of Machine Learning Research}, pages 7783--7792, 2020.

\bibitem[Seita et~al.(2019)Seita, Chan, Rao, Tang, Zhao, and Canny]{seita2019zpd}
D.~Seita, D.~Chan, R.~Rao, C.~Tang, M.~Zhao, and J.~Canny.
\newblock {ZPD} teaching strategies for deep reinforcement learning from demonstrations.
\newblock In \emph{Deep Reinforcement Learning Workshop at NeurIPS}, 2019.

\bibitem[Mu et~al.(2024)Mu, Chen, Zhang, Chen, Yu, Ge, Chen, Liang, Hu, Tao, Sun, Yu, Yang, Shao, Wang, Dai, Qiao, Ding, and Luo]{mu2024robocodexmultimodalcodegeneration}
Y.~Mu, J.~Chen, Q.~Zhang, S.~Chen, Q.~Yu, C.~Ge, R.~Chen, Z.~Liang, M.~Hu, C.~Tao, P.~Sun, H.~Yu, C.~Yang, W.~Shao, W.~Wang, J.~Dai, Y.~Qiao, M.~Ding, and P.~Luo.
\newblock Robocodex: Multimodal code generation for robotic behavior synthesis, 2024.
\newblock URL \url{https://arxiv.org/abs/2402.16117}.

\bibitem[Team et~al.(2025)Team, Abdolmaleki, Abeyruwan, Ainslie, Alayrac, Arenas, Balakrishna, Batchelor, Bewley, Bingham, et~al.]{geminirobotics1d52025}
G.~R. Team, A.~Abdolmaleki, S.~Abeyruwan, J.~Ainslie, J.-B. Alayrac, M.~G. Arenas, A.~Balakrishna, N.~Batchelor, A.~Bewley, J.~Bingham, et~al.
\newblock Gemini robotics 1.5: Pushing the frontier of generalist robots with advanced embodied reasoning, thinking, and motion transfer.
\newblock \emph{arXiv preprint arXiv:2510.03342}, 2025.

\bibitem[Shi et~al.(2025)Shi, Yang, Chao, Wan, Shao, Lei, Qian, Le, Chaudhari, Daniilidis, et~al.]{shi2025maestro}
J.~Shi, R.~Yang, K.~Chao, B.~S. Wan, Y.~S. Shao, J.~Lei, J.~Qian, L.~Le, P.~Chaudhari, K.~Daniilidis, et~al.
\newblock Maestro: Orchestrating robotics modules with vision-language models for zero-shot generalist robots.
\newblock In \emph{NeurIPS 2025 Workshop on Space in Vision, Language, and Embodied AI}, 2025.

\bibitem[Ahn et~al.(2022)Ahn, Brohan, Brown, Chebotar, Cortes, David, Finn, Fu, Gopalakrishnan, Hausman, Herzog, Ho, Hsu, Ibarz, Ichter, Irpan, Jang, Ruano, Jeffrey, Jesmonth, Joshi, Julian, Kalashnikov, Kuang, Lee, Levine, Lu, Luu, Parada, Pastor, Quiambao, Rao, Rettinghouse, Reyes, Sermanet, Sievers, Tan, Toshev, Vanhoucke, Xia, Xiao, Xu, Xu, Yan, and Zeng]{ahn2022saycan}
M.~Ahn, A.~Brohan, N.~Brown, Y.~Chebotar, O.~Cortes, B.~David, C.~Finn, C.~Fu, K.~Gopalakrishnan, K.~Hausman, A.~Herzog, D.~Ho, J.~Hsu, J.~Ibarz, B.~Ichter, A.~Irpan, E.~Jang, R.~J. Ruano, K.~Jeffrey, S.~Jesmonth, N.~Joshi, R.~Julian, D.~Kalashnikov, Y.~Kuang, K.-H. Lee, S.~Levine, Y.~Lu, L.~Luu, C.~Parada, P.~Pastor, J.~Quiambao, K.~Rao, J.~Rettinghouse, D.~Reyes, P.~Sermanet, N.~Sievers, C.~Tan, A.~Toshev, V.~Vanhoucke, F.~Xia, T.~Xiao, P.~Xu, S.~Xu, M.~Yan, and A.~Zeng.
\newblock Do as {I} can, not as {I} say: Grounding language in robotic affordances.
\newblock In \emph{Proceedings of the 6th Conference on Robot Learning (CoRL)}, volume 205 of \emph{Proceedings of Machine Learning Research}, pages 287--318, 2022.

\bibitem[Driess et~al.(2023)Driess, Xia, Sajjadi, Lynch, Chowdhery, Ichter, Wahid, Tompson, Vuong, Yu, Huang, Chebotar, Sermanet, Duckworth, Levine, Vanhoucke, Hausman, Toussaint, Greff, Zeng, Mordatch, and Florence]{driess2023palme}
D.~Driess, F.~Xia, M.~S.~M. Sajjadi, C.~Lynch, A.~Chowdhery, B.~Ichter, A.~Wahid, J.~Tompson, Q.~Vuong, T.~Yu, W.~Huang, Y.~Chebotar, P.~Sermanet, D.~Duckworth, S.~Levine, V.~Vanhoucke, K.~Hausman, M.~Toussaint, K.~Greff, A.~Zeng, I.~Mordatch, and P.~Florence.
\newblock {PaLM-E}: An embodied multimodal language model.
\newblock In \emph{Proceedings of the 40th International Conference on Machine Learning (ICML)}, volume 202 of \emph{Proceedings of Machine Learning Research}, pages 8469--8488, 2023.

\bibitem[Huang et~al.(2023)Huang, Wang, Zhang, Li, Wu, and Fei-Fei]{huang2023voxposer}
W.~Huang, C.~Wang, R.~Zhang, Y.~Li, J.~Wu, and L.~Fei-Fei.
\newblock {VoxPoser}: Composable {3D} value maps for robotic manipulation with language models.
\newblock In \emph{Proceedings of the 7th Conference on Robot Learning (CoRL)}, volume 229 of \emph{Proceedings of Machine Learning Research}, 2023.

\bibitem[Zeng et~al.(2023)Zeng, Attarian, Ichter, Choromanski, Wong, Welker, Tombari, Purohit, Ryoo, Sindhwani, Lee, Vanhoucke, and Florence]{zeng2023socratic}
A.~Zeng, M.~Attarian, B.~Ichter, K.~Choromanski, A.~Wong, S.~Welker, F.~Tombari, A.~Purohit, M.~Ryoo, V.~Sindhwani, J.~Lee, V.~Vanhoucke, and P.~Florence.
\newblock Socratic models: Composing zero-shot multimodal reasoning with language.
\newblock In \emph{The Eleventh International Conference on Learning Representations (ICLR)}, 2023.

\bibitem[Mu et~al.(2023)Mu, Zhang, Hu, Wang, Ding, Jin, Wang, Dai, Qiao, and Luo]{mu2023embodied}
Y.~Mu, Q.~Zhang, M.~Hu, W.~Wang, M.~Ding, J.~Jin, B.~Wang, J.~Dai, Y.~Qiao, and P.~Luo.
\newblock {EmbodiedGPT}: Vision-language pre-training via embodied chain of thought.
\newblock In \emph{Advances in Neural Information Processing Systems 36 (NeurIPS)}, pages 25081--25094, 2023.

\bibitem[Yao et~al.(2023)Yao, Zhao, Yu, Du, Shafran, Narasimhan, and Cao]{yao2023react}
S.~Yao, J.~Zhao, D.~Yu, N.~Du, I.~Shafran, K.~Narasimhan, and Y.~Cao.
\newblock {ReAct}: Synergizing reasoning and acting in language models.
\newblock In \emph{International Conference on Learning Representations}, 2023.
\newblock URL \url{https://openreview.net/forum?id=WE_vluYUL-X}.

\bibitem[Shinn et~al.(2023)Shinn, Cassano, Berman, Gopinath, Narasimhan, and Yao]{shinn2023reflexion}
N.~Shinn, F.~Cassano, E.~Berman, A.~Gopinath, K.~Narasimhan, and S.~Yao.
\newblock Reflexion: Language agents with verbal reinforcement learning.
\newblock In \emph{Advances in Neural Information Processing Systems 36 (NeurIPS)}, 2023.

\bibitem[Madaan et~al.(2023)Madaan, Tandon, Gupta, Hallinan, Gao, Wiegreffe, Alon, Dziri, Prabhumoye, Yang, Gupta, Majumder, Hermann, Welleck, Yazdanbakhsh, and Clark]{madaan2023selfrefine}
A.~Madaan, N.~Tandon, P.~Gupta, S.~Hallinan, L.~Gao, S.~Wiegreffe, U.~Alon, N.~Dziri, S.~Prabhumoye, Y.~Yang, S.~Gupta, B.~P. Majumder, K.~Hermann, S.~Welleck, A.~Yazdanbakhsh, and P.~Clark.
\newblock Self-refine: Iterative refinement with self-feedback.
\newblock In \emph{Advances in Neural Information Processing Systems 36 (NeurIPS)}, 2023.

\bibitem[Chen et~al.(2024)Chen, Lin, Sch{\"a}rli, and Zhou]{chen2024selfdebug}
X.~Chen, M.~Lin, N.~Sch{\"a}rli, and D.~Zhou.
\newblock Teaching large language models to self-debug.
\newblock In \emph{The Twelfth International Conference on Learning Representations (ICLR)}, 2024.

\bibitem[Wang et~al.(2024)Wang, Xie, Jiang, Mandlekar, Xiao, Zhu, Fan, and Anandkumar]{wang2023voyager}
G.~Wang, Y.~Xie, Y.~Jiang, A.~Mandlekar, C.~Xiao, Y.~Zhu, L.~Fan, and A.~Anandkumar.
\newblock Voyager: An open-ended embodied agent with large language models.
\newblock \emph{Transactions on Machine Learning Research (TMLR)}, 2024.

\bibitem[Lin et~al.(2025)Lin, Snell, Wang, Packer, Wooders, Stoica, and Gonzalez]{lin2025sleeptime}
K.~Lin, C.~Snell, Y.~Wang, C.~Packer, S.~Wooders, I.~Stoica, and J.~E. Gonzalez.
\newblock Sleep-time compute: Beyond inference scaling at test-time.
\newblock \emph{arXiv preprint arXiv:2504.13171}, 2025.

\bibitem[Zhu et~al.(2020)Zhu, Wong, Mandlekar, Mart\'{i}n-Mart\'{i}n, Joshi, Lin, Nasiriany, and Zhu]{robosuite2020}
Y.~Zhu, J.~Wong, A.~Mandlekar, R.~Mart\'{i}n-Mart\'{i}n, A.~Joshi, K.~Lin, S.~Nasiriany, and Y.~Zhu.
\newblock robosuite: A modular simulation framework and benchmark for robot learning.
\newblock In \emph{arXiv preprint arXiv:2009.12293}, 2020.

\bibitem[Liu et~al.(2023)Liu, Zhu, Gao, Feng, Liu, Zhu, and Stone]{liu2023libero}
B.~Liu, Y.~Zhu, C.~Gao, Y.~Feng, Q.~Liu, Y.~Zhu, and P.~Stone.
\newblock Libero: Benchmarking knowledge transfer for lifelong robot learning.
\newblock \emph{Advances in Neural Information Processing Systems}, 36:\penalty0 44776--44791, 2023.

\bibitem[Zhou et~al.(2025)Zhou, Xu, Tie, Chen, Zhang, Chu, Zhou, and Sun]{liberopro}
X.~Zhou, Y.~Xu, G.~Tie, Y.~Chen, G.~Zhang, D.~Chu, P.~Zhou, and L.~Sun.
\newblock Libero-pro: Towards robust and fair evaluation of vision-language-action models beyond memorization.
\newblock \emph{arXiv preprint arXiv:2510.03827}, 2025.

\bibitem[Kim et~al.(2026)Kim, Pumacay, Rayyan, Argus, Han, VanderBilt, Salvador, Deshpande, Hendrix, Jauhri, et~al.]{kim2026molmospaces}
Y.~Kim, W.~Pumacay, O.~Rayyan, M.~Argus, W.~Han, E.~VanderBilt, J.~Salvador, A.~Deshpande, R.~Hendrix, S.~Jauhri, et~al.
\newblock Molmospaces: A large-scale open ecosystem for robot navigation and manipulation.
\newblock \emph{arXiv preprint arXiv:2602.11337}, 2026.

\bibitem[Kim et~al.(2024)Kim, Pertsch, Karamcheti, Xiao, Balakrishna, Nair, Rafailov, Foster, Sanketi, Vuong, et~al.]{openvla}
M.~J. Kim, K.~Pertsch, S.~Karamcheti, T.~Xiao, A.~Balakrishna, S.~Nair, R.~Rafailov, E.~P. Foster, P.~R. Sanketi, Q.~Vuong, et~al.
\newblock Openvla: An open-source vision-language-action model.
\newblock In \emph{Conference on Robot Learning}, 2024.

\end{thebibliography}
\clearpage
\appendix

\etocdepthtag.toc{mtappendix}
\etocsettagdepth{mtchapter}{none}    
\etocsettagdepth{mtappendix}{subsection} 

\clearpage

\section*{Appendix}

\begingroup
  \parskip0pt
  \etocsettocstyle{\section*{Table of Contents}}{} 
  \tableofcontents
\endgroup

\clearpage

\section{Implementation Details about \textsc{RATs}}
\label{sec:appendix_agent_team}
This section provides implementation details for \textsc{RATs}
introduced in Sec.~\ref{sec:method}. 
We introduce the details of the three teams that make up \textsc{RATs}: the \emph{Task Proposer} team (play-time task proposal), the \emph{Execution} team (task execution), and the \emph{Memory-Management} team (skill library and failure memory updates).
We then describe how the learned skill
library is used during evaluation and summarize the prompt interfaces of each
agent.

\subsection{Details of Play-Time Task Proposal}
\label{sec:appendix_task_proposal}

\textbf{Candidate task generation.}
At each play iteration, the Task Proposer receives the current scene context, a
compact summary of the skill library $\mathcal{L}$, and a short history of
recent attempts. The skill summary includes each learned skill's name,
description, reliability tier, and empirical success statistics, but not its
full source code. This lets the proposer reason about available capabilities
and explored object-skill pairs without filling the context with code. The
proposer then generates a candidate pool $\mathcal{T}_t$ of play tasks, along
with the objects and skills required by each candidate.

The scene context is instantiated differently across environments. In LIBERO,
the proposer uses the object inventory and available manipulation primitives.
In MolmoSpaces, object identifiers are first grounded into readable phrases
from the current visual observation, and only visible, reachable objects are
shown as proposal targets. This prevents the proposer from selecting objects
that exist in the scene metadata but are not accessible from the robot's
current state.

\textbf{Goldilocks-driven task selection.}
After candidate generation, we score each candidate using the Goldilocks
objective in Sec.~\ref{sec:proposer}. The novelty term is computed from
historical counts over object-skill pairs. The competence estimate uses the
Wilson lower bound of each required skill's empirical success rate, rather than
the raw success rate, so that a rarely used skill is not treated as reliable
after one lucky success. We rank candidates by the product of object-skill
novelty and competence-frontier score. We also downweight tasks that closely
match recent failures. Diagnosable failures with initially plausible plans are
stored in a bounded retry bank, which can later propose simplified variants
with a decaying retry bonus.

\textbf{Environment creation.}
After a task is selected, the Environment Creator converts the proposal into an
executable task instance. In LIBERO, it generates a BDDL task specification,
validates its syntax and semantics, and instantiates the corresponding MuJoCo
environment. Static checks require all referenced objects, fixtures, regions,
predicates, and assets to match the proposal. If validation fails, the creator
performs one bounded repair and validates the repaired specification again. A
compact example is shown below.

\begin{verbatim}
(:language pick up the butter and put it on the plate)
(:objects butter_1 - butter plate_1 - plate)
(:init
  (On butter_1 kitchen_table_butter_init_region)
  (On plate_1 kitchen_table_plate_init_region))
(:goal (And (On butter_1 plate_1)))
\end{verbatim}

\textbf{Environment verification.}
Before a LIBERO task enters the execution stage, the Environment Verifier runs
deterministic checks over two reset seeds. The environment must instantiate and
render successfully, expose the simulator, realize each declared object body,
evaluate every goal predicate without error, and avoid severe initial
penetrations. In MolmoSpaces, task creation is constrained by the bridge
catalog. After rebinding the environment, the verifier checks that the proposed
target remains visually grounded. Rejected tasks return to the task proposing stage
rather than consuming an execution attempt.

\subsubsection{Example Trace of Play-Time Task Proposal}
\begin{figure}[t]
\centering
\includegraphics[width=\linewidth]{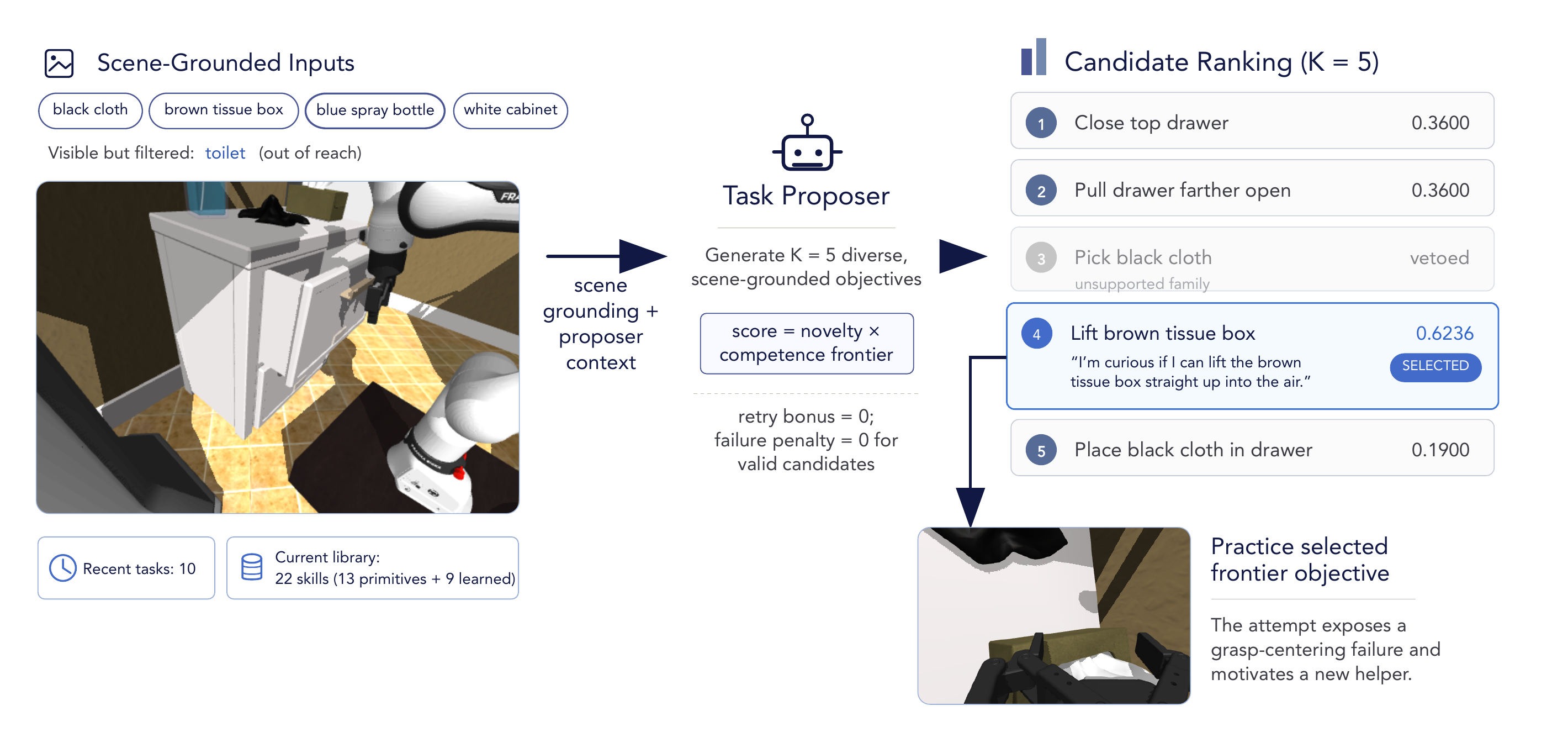}
\vspace{-1.5em}
\caption{
\textbf{Example trace of play-time task proposal.}
An example of the MolmoSpaces play-time task proposal process at iteration 15.
}
\label{fig:play_task_proposal}
\vspace{-1em}
\end{figure}

Table~\ref{tab:proposal_selection_example} and Figure~\ref{fig:play_task_proposal} show the saved candidate trace from
MolmoSpaces play iteration 15. Scene grounding first marks five objects as
visible: a blue spray bottle, black cloth, brown tissue box, toilet, and white
cabinet. Reachability filtering removes the toilet, leaving four usable
main targets for proposal. The Task Proposer receives these targets together
with the exact current library of 22 skills (13 primitives and 9 learned
skills) and the previous 10 task records, then generates $K=5$ candidates. The
selected tissue-box lift receives the highest composite score ($0.6236$): it is
novel but remains close enough to the robot's current grasping capabilities to
be worth practicing. The black-cloth pick is vetoed before ranking because its
interaction is unsupported in the active bridge configuration.

\textbf{Inputs to Task Proposer.}
The Task Proposer sees both low-level primitives and the learned skills
accumulated by earlier play. Its 13 primitives are grouped into
perception (\texttt{get\_observation}, \texttt{segment\_sam3\_text\_prompt},
\texttt{segment\_sam3\_point\_prompt}, \texttt{point\_prompt\_molmo}, and
\texttt{vlm\_verify}), grasp planning and control
(\texttt{plan\_grasp\_from\_point\_clouds}, \texttt{solve\_ik},
\texttt{move\_to\_joints}, \texttt{goto\_pose}, and
\texttt{goto\_home\_joint\_position}), and gripper control
(\texttt{open\_gripper}, \texttt{close\_gripper}, and
\texttt{select\_top\_down\_grasp}). Table~\ref{tab:proposal_skill_snapshot}
lists the nine learned skills by function name. The displayed success
rates are the raw metadata shown to the proposer; the analytical ranker uses
Wilson-bounded reliability instead.

\begin{table*}[t]
\centering
\scriptsize
\setlength{\tabcolsep}{4pt}
\caption{
\textbf{Learned-skill snapshot provided to the Task Proposer at
MolmoSpaces play iteration 15.}
The prompt additionally contains the 13 primitive function names listed in the
preceding paragraph. ``Raw rate'' is the success-rate metadata in the prompt, not the
Wilson-bounded value used for analytical selection.
}
\label{tab:proposal_skill_snapshot}
\vspace{1.5em}
\begin{tabular}{p{0.31\textwidth} p{0.42\textwidth} c c c}
\toprule
\textbf{Learned skill in prompt} & \textbf{Reusable capability} &
\textbf{Raw rate} & \textbf{Uses} & \textbf{Tier} \\
\midrule
\skill{localize\_object\_with\_molmo\_sam3} & Localize an object from agent-view Molmo prompting and SAM3 segmentation. & 0.3125 & 32 & Exp. \\
\skill{plan\_top\_down\_grasp\_at\_wrist} & Move the wrist camera over an object and plan a top-down grasp from segmented point clouds. & 0.2500 & 16 & Ver. \\
\skill{get\_axis\_aligned\_pull\_direction} & Compute an axis-aligned pull direction from the target toward the robot base. & 0.4286 & 7 & Ver. \\
\skill{select\_grasp\_for\_pulling} & Select a handle grasp whose approach opposes a requested pull direction. & 0.0000 & 0 & Exp. \\
\skill{execute\_top\_down\_grasp\_and\_lift} & Approach from above, close the gripper, and lift from a target 3D position. & 0.4167 & 12 & Ver. \\
\skill{push\_surface\_inward} & Push a surface inward to close a drawer or door. & 1.0000 & 2 & Exp. \\
\skill{execute\_grasp\_from\_transform} & Execute a grasp pose with a collision-avoidance height offset, then lift. & 0.2000 & 5 & Exp. \\
\skill{translate\_grasped\_object\_and\_release} & Translate a grasped object, release it, and retreat vertically. & 0.2000 & 5 & Exp. \\
\skill{push\_object\_closed} & Localize an open articulation, infer its outward direction, and push inward. & 1.0000 & 1 & Exp. \\
\bottomrule
\end{tabular}
\end{table*}

The proposer is also instructed to vary from the last ten attempts, avoid exact
verb-target duplicates, build one-variable-changed experiments from successes,
and simplify or change direction after failures. Table~\ref{tab:proposal_history}
transcribes the exact task-language field, success flag, and retry count inserted
into this prompt. Its diagnostic column condenses the accompanying
\texttt{failure\_reason} field while retaining the concrete failure mode and
suggested argument change.

\begin{table*}[t]
\centering
\scriptsize
\setlength{\tabcolsep}{4pt}

\caption{
\textbf{Recent objectives and outcomes supplied to the Task Proposer.}
The previous ten tasks and their successes are supplied to the Task Proposer. Relevant learned skills or failure diagnoses are listed as well.
}
\label{tab:proposal_history}
\vspace{1.5em}
{\renewcommand{\arraystretch}{1.1}
\begin{tabular}{c p{0.37\textwidth} c c p{0.42\textwidth}}
\toprule
& \textbf{Previous task language in prompt} & \textbf{Success} &
\textbf{Retries} & \textbf{Learned skill or failure diagnosis} \\
\midrule
1 & I wonder what happens if I lift the lever straight up into the air. & Yes & 0 & -- \\
2 & I want to see if I can push the top drawer all the way closed. & Yes & 0 & Learned \texttt{push\_surface\_inward}. \\
3 & Let's lift the brown sandal straight up. & Yes & 3 & -- \\
4 & I'm curious what happens when I lift the yellow button straight up. & Yes & 0 & -- \\
5 & I want to see what happens when I lift the toothpaste tube. & No & 4 & \texttt{grasp\_failure: argument\_level}. The $0.025$\,m grasp \texttt{z\_offset} left the fingers above the thin tube; the diagnosis suggests $0.0$ or $-0.01$\,m. \\
6 & Let's try to lift the blue clothes iron straight up and see what happens. & Yes & 0 & -- \\
7 & I wonder what happens if I lift the black videocassette straight up. & No & 4 & \texttt{collision: argument\_level}. Subtracting $0.035$\,m placed the grasp below the table; the diagnosis suggests subtracting $0.015$ or $0.02$\,m. \\
8 & I wonder if I can pull the walkie-talkie closer to me on the bed. & Yes & 0 & Learned two skills; \texttt{execute\_grasp\_from\_transform} and \texttt{translate\_grasped\_object\_and\_release}. \\
9 & I wonder if I can pull the metal ring closer to me across the table. & No & 4 & \texttt{grasp\_failure: rewrite\_needed}. The fingers missed or slipped from the knot; the diagnosis suggests targeting the center hole or subtracting $0.015$\,m from the grasp-pose height. \\
10 & I want to see if I can push the partially open drawer closed. & Yes & 0 & Learned \texttt{push\_object\_closed}. \\
\bottomrule
\end{tabular}
}
\end{table*}

\textbf{Why these candidates are proposed.}
Closing the top drawer is a nearby one-variable variation
of the two successful drawer-closing tasks and can reuse
\texttt{push\_object\_closed}. Pulling the drawer farther open probes the
opposite articulation direction while reusing the available pull-direction and
handle-grasp helpers. Picking up the cloth and placing it inside the open drawer
exercise two under-explored interactions on a visible object. Finally, lifting the
tissue box applies the verified top-down lift routine to a novel visible
object. This causal reading is an interpretation of the recorded prompt and
candidate rationales: the trace does not log token-level attribution inside the
proposer LLM.

\textbf{Why the tissue-box lift task is selected.}
The ranker uses
$s(\tau)=\mathcal{N}(\tau)\mathcal{F}(\tau)+w_BB_{\mathrm{retry}}(\tau)
-w_PP_{\mathrm{fail}}(\tau)$ for scoring tasks. Here, $\mathcal{F}$ is the competence-frontier learnability term. The analytical ranker computes
object-skill novelty and obtains $\mathcal{N}=1.0$ for every
candidate because each projected target-action pair is new. Surprise is used
only inside the retry-bank bonus path; this proposal turn has no surprise value, so $B_{\mathrm{retry}}=0$. None of the candidates
is sufficiently similar to a recent failed task to receive a failure penalty.

The remaining difference is therefore the competence frontier. For drawer closing and
opening, known primitive baselines
(\texttt{close\_gripper} and \texttt{open\_gripper}) and learned skills (\texttt{push\_object\_closed}) are used in competence calculation, yielding
$\bar{r}=0.9$ and $\mathcal{F}=4(0.9)(0.1)=0.36$. The cloth placement projects to \texttt{place\_in} family, for which the
library has no matching helper; the missing-skill default $\bar{r}=0.05$ gives
$\mathcal{F}=0.19$. The tissue-box lift projects to \texttt{lift} family, which
matches \texttt{execute\_top\_down\_grasp\_and\_lift}. That helper has five
successes in 12 uses: its raw prompt rate is $0.4167$, while its conservative
Wilson lower bound is $\bar{r}=0.1933$. Consequently,
$\mathcal{F}=4(0.1933)(1-0.1933)=0.6236$. The lift is neither already mastered
nor entirely unsupported, so it receives the highest Goldilocks score. After
selection, the tissue-box practice task remains unsuccessful after four
attempts and exposes a grasp-centering bottleneck, motivating the proposed
helper \texttt{adjust\_grasp\_to\_centroid}.

\begin{table*}[t]
\centering
\scriptsize
\setlength{\tabcolsep}{4pt}
\caption{
\textbf{Candidate-selection trace from MolmoSpaces play iteration 15.}
The Task Proposer generates five scene-grounded alternatives and the analytical
ranker selects a learnable frontier objective after bridge-compatibility
checks. A dash in the surprise column indicates that surprise
value is zero for this play-time proposer turn.
}
\label{tab:proposal_selection_example}
\vspace{1.5em}
\begin{tabular}{p{0.3\textwidth} l c c c c c c c p{0.05\textwidth}}
\toprule
\textbf{Candidate objective} & \textbf{Family} & $\mathcal{N}$ & $\bar{r}$ &
$\mathcal{F}$ & \textbf{Surp.} & $w_BB_{\mathrm{retry}}$ &
$w_PP_{\mathrm{fail}}$ & \textbf{Score} & \textbf{Result} \\
\midrule
Push the top drawer of the white cabinet all the way closed. & Close & 1.0000 & 0.9000 & 0.3600 & -- & 0.0000 & 0.0000 & 0.3600 & Valid \\
Pull the partially open drawer out even more. & Open & 1.0000 & 0.9000 & 0.3600 & -- & 0.0000 & 0.0000 & 0.3600 & Valid \\
Pick up the black cloth from the cabinet. & Pick & -- & -- & -- & -- & -- & -- & 0.0000 & Vetoed \\
Lift the brown tissue box straight up into the air. & Lift & 1.0000 & 0.1933 & 0.6236 & -- & 0.0000 & 0.0000 & 0.6236 & Selected \\
Put the black cloth inside the open drawer. & Place in & 1.0000 & 0.0500 & 0.1900 & -- & 0.0000 & 0.0000 & 0.1900 & Valid \\
\bottomrule
\end{tabular}
\end{table*}

\subsection{Details of \textsc{RATs} for Execution}
\label{sec:appendix_execution}

\textsc{RATs} executes each accepted task through a bounded
Write-Execute-Verify-Diagnose loop. The loop is designed to localize failures
to planning, code generation, or physical execution before the next retry.

\textbf{Planner.}
The Planner receives the task description, the initial scene observation,
retrieved lessons from failure memory, primitive APIs, and active learned
skills. Skills are ordered by reliability: verified skills are shown before
experimental skills, while deprecated skills are hidden by default. The Planner
outputs an ordered plan, annotates each step with relevant skills, and predicts
likely failure points for downstream inspection.

\textbf{Planner Verifier.}
Before code generation, the Planner Verifier checks whether the plan is grounded
in the initial observation. It looks for scene misperception, missing
preconditions, incorrect action ordering, and object-scope mismatch. If the
issue is correctable, the Planner revises the plan using the verifier's
feedback. This refinement repeats until the plan passes or reaches the
refinement budget.

\textbf{Policy Writer.}
The Policy Writer converts the verified plan into executable Python code. Its
prompt includes selected skill references, primitive API documentation,
retrieved failure lessons, and feedback from previous attempts. On retry, it
also receives per-step verification results and a summary of code segments that
already worked, encouraging local edits instead of rewriting the full policy.

\textbf{Quality Checker.}
Before execution, the Quality Checker statically screens the generated code. It
rejects syntax errors, unavailable API calls, unbounded loops, forbidden code
patterns, and malformed result reporting. This avoids spending robot
interaction budget on errors that can be detected from source code alone.

\textbf{Goal Verifier.}
After execution, the Goal Verifier determines task success. When a structured
predicate is available, success is evaluated from environment state. Otherwise,
the verifier uses a structured custom check or visual judgment. A policy crash
always counts as failure, even if the final visual state appears plausible.

\textbf{Per-Step Verifier.}
The Per-Step Verifier provides localized evidence for diagnosis. For each plan
step, it receives the step objective, corresponding code slice, critical runtime
values, and before/after visual evidence, including a short motion clip when
available. It returns a step-level verdict and explanation, distinguishing, for
example, a failed grasp from a successful grasp followed by an incorrect
placement.

\textbf{Failure Diagnoser.}
When a task fails, the Failure Diagnoser reads the execution trace, terminal
state, goal-verifier result, and per-step verdicts. It returns a failure
category, the first failed step, a concrete repair suggestion, and optional
routing flags. Code-local failures are routed back to the Policy Writer.
Plan-level failures trigger plan refinement. Persistent local physical
bottlenecks can spawn a SubAgent.

\textbf{SubAgent.}
The SubAgent practices a single failed sub-action, such as a grasp or
articulation, in the same reset state. It evaluates candidate scripts and
returns the first visually verified reusable helper routine. The winning
routine is inserted into the Policy Writer context for the current retry, so it
can be reused at the failed step. It is not automatically promoted into the
persistent skill library; promotion happens only after the routine contributes
to a successful end-to-end execution or is separately proposed and validated.
This keeps useful SubAgent discoveries available without polluting the library
with overfitted local fixes.

\subsection{Details of Skill Library and Failure Memory Updates}
\label{sec:appendix_skill_memory}

\textsc{RATs} maintains two persistent stores: the skill library $\mathcal{L}$
and the failure memory $\mathcal{M}$. Both are updated after each play attempt
and periodically curated to keep retrieval useful.

\textbf{Skill library.}
Each skill in $\mathcal{L}$ stores executable Python source code, a
description, preconditions, expected effects, dependencies, provenance, usage
counts, success counts, and a reliability tier. New code is parsed and
validated before insertion. The validation checks that the helper defines a
callable function, uses only available primitives or known dependencies, and
does not duplicate an existing skill. The library exposes a metadata-only view
for task proposal and a code-bearing view for planning and execution.

\textbf{Skill reliability.}
Skills follow a three-tier reliability lifecycle. A new skill starts as
\emph{experimental}. It is promoted to \emph{verified} after at least three
uses with empirical success rate at least $0.5$, and marked
\emph{deprecated} after at least ten uses with success rate at most $0.2$.
Verified skills can be demoted after sustained poor performance. Retrieval
prioritizes verified skills using Wilson-bounded reliability and hides
deprecated skills by default. At execution time, the runtime injects selected
skill definitions and their dependencies into the policy namespace, while
keeping active helpers available as a dependency-safety fallback.

\textbf{Skill extraction on success.}
After a successful end-to-end trajectory, the system extracts self-contained,
parameterized helper functions from the executed policy. The extracted helpers
capture reusable behavioral units rather than entire task scripts. They are
statically validated and inserted into $\mathcal{L}$ as experimental skills.
The successful trajectory also updates usage statistics for any learned skills
invoked during execution.

\textbf{Failure memory on failure.}
On failure, the system records the episode in $\mathcal{M}$. Each raw episode
stores the task, involved objects, failure category, failed plan step,
diagnostic explanation, attempted approach, and relevant code excerpt. Related
failures are distilled into compact lessons: when a condition occurs, avoid the
failed approach and use the suggested correction. The Planner retrieves lessons
by task and object overlap, allowing later attempts to benefit from prior
failures without replaying long traces in the prompt.

\textbf{Memory curation and skill proposal.}
Every $K$ play iterations, the Memory Curator merges, rewrites, or removes
redundant lessons and near-duplicate skills. When repeated failures reveal a
missing capability, the Skill Proposer drafts a candidate helper from
primitives and learned skills. Unlike skill extraction, which stores code that
has already succeeded, the Skill Proposer is anticipatory: the proposed helper
enters $\mathcal{L}$ as experimental and earns reliability only through later
use.

\subsection{Details of Evaluation With the Learned Skill Library}
\label{sec:appendix_test_time}

At test time, the learned library $\mathcal{L}$ is frozen and evaluated on
externally specified benchmark tasks. Task proposal and memory curation are disabled. We evaluate the learned library in two settings:
plug-and-play execution with a standard Code-as-Policy baseline, and full
\textsc{RATs} execution.

\paragraph{Plug-and-Play \textsc{CaP-Agent0} evaluation.}
In the plug-and-play setting, learned skills are loaded from the frozen library
and exposed to a standard single-agent Code-as-Policy baseline, \textsc{CaP-Agent0}.
Each selected skill's signature, description, and source code are added to the
baseline API context, and the function definitions are inserted into the
execution namespace. For large libraries, a lightweight selector can retrieve a
task-relevant subset before prompt construction. This setting omits the
\textsc{RATs} Planner, verification-driven retry loop, and failure-memory
retrieval, isolating the transfer value of the learned code library itself.

\paragraph{\textsc{RATs} evaluation.}
In the full \textsc{RATs} setting, the same frozen library is provided to the
Planner. The Planner receives primitive APIs and active learned skills,
prioritizes verified skills over experimental ones, and selects relevant skills
for individual plan steps. The Policy Writer then generates code conditioned on
these selected skills, while the runtime injects their executable definitions
and dependencies. Goal verification, per-step verification, diagnosis, and
bounded retry remain active. This setting measures the combined value of
play-time skill acquisition and the \textsc{RATs} execution loop on unseen
tasks.

\section{Additional Real-World Experimental Results}
\label{sec:appendix_additional_realworld}

\subsection{Additional Quantitative Results for Real-World Experiments}

We further evaluate whether skills learned from MolmoSpaces play can transfer to additional
real-world manipulation tasks. We add two real-world tasks that require object rearrangement and
articulated-object interaction. In \textit{Swap Cubes}, the robot must pick up the cube on the
platform, move it off the platform, and exchange it with the cube initially placed below the platform.
In \textit{Close Drawer}, the robot must close an initially open drawer. For both tasks, we compare
the standard \textsc{CaP-Agent0} system against \textsc{CaP-Agent0} augmented with skills learned
by \textsc{RATs} during MolmoSpaces play.

Table~\ref{tab:additional-real-world} shows that adding MolmoSpaces play-learned skills improves
performance on both tasks. On \textit{Swap Cubes}, the standard \textsc{CaP-Agent0} baseline fails
to solve the task, while the skill-augmented agent succeeds in 7 out of 30 trials. On
\textit{Close Drawer}, the success rate improves from 2/30 to 8/30. Averaged over the two tasks,
MolmoSpaces skills improve real-world success from 3.3\% to 25.0\%, corresponding to a
+21.7 percentage-point gain. These results suggest that play-learned skills from MolmoSpaces can
provide useful reusable behaviors for real-world manipulation beyond the original benchmark tasks.

\begin{table}[ht]
    \centering
    \setlength{\tabcolsep}{13pt}
    \caption{\textbf{Additional real-world evaluation with MolmoSpaces skills.}
    Skills are learned by \textsc{RATs} during MolmoSpaces play and reused with the
    \textsc{CaP-Agent0} system. }
    \label{tab:additional-real-world}
    \resizebox{\robosuiteTableWidth}{!}{%
    \begin{tabular}{lccc}
        \toprule
        Task
        & \textsc{CaP-Agent0}
        & \textsc{CaP-Agent0} + \textsc{RATs} Skills (MolmoSpaces)
        & $\Delta$ \\
        \midrule
        Swap Cubes
            & 0/30 (0.0\%)
            & \textbf{7/30 (23.3\%)}
            & +23.3 pp \\
        Close Drawer
            & 2/30 (6.7\%)
            & \textbf{8/30 (26.7\%)}
            & +20.0 pp \\
        \midrule
        \textbf{Average (Real World)}
            & 2/60 (3.3\%)
            & \textbf{15/60 (25.0\%)}
            & +21.7 pp \\
        \bottomrule
    \end{tabular}
    \vspace{-2em}
    }
\end{table}

\subsection{Additional Qualitative Results}
\label{sec:appendix_additional_qualitative}

We also provide additional qualitative examples of successful real-world executions using
\textsc{CaP-Agent0} augmented with \textsc{RATs} skills learned from MolmoSpaces play. As shown
in Figure~\ref{fig:real}, the skill-augmented agent successfully solves
the newly added \textit{Swap Cubes} and \textit{Close Drawer} tasks, as well as an \textit{Open Drawer}
task. Specifically, in \textit{Swap Cubes}, the robot exchanges a cube on the platform with a cube 
below the platform; in \textit{Close Drawer}, the robot closes an initially open drawer; and in 
\textit{Open Drawer}, the robot opens a closed drawer. These examples illustrate that the MolmoSpaces 
skill library can support both object rearrangement and articulated-object manipulation in the real world. 
We include video results of these real-world rollouts, along with additional simulation rollouts, 
on the project webpage.

\section{Details about Play Process}
\label{sec:appendix_play_process}

\subsection{Play-Time Objectives and Knowledge Accumulation}
\label{sec:appendix_play_accumulation}

We further analyze the objectives proposed during play and the
knowledge accumulated from solving them. Below we analyze a 50-iteration play
run in MolmoSpaces.

\begin{figure}[t]
\centering
\includegraphics[width=\linewidth]{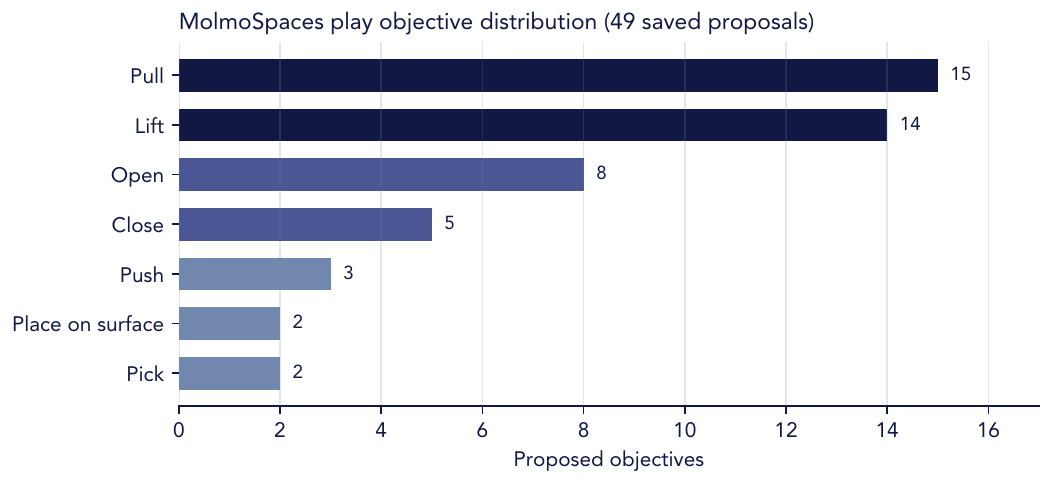}
\vspace{-1.5em}
\caption{
\textbf{Distribution of play objectives.}
Counts are computed from the 49 saved proposal records available for the
50-iteration run.
}
\label{fig:play_objective_distribution}
\vspace{-1em}
\end{figure}

\textbf{Distribution of proposed objectives.}
Figure~\ref{fig:play_objective_distribution} summarizes the run.
One iteration did not contain a saved proposal artifact due to an error, so the distribution
is computed from the remaining 49 records. The proposer covers seven
interaction families and more than 30 object categories, including articulated
furniture, doors, tools, shoes, kitchen objects, and small tabletop objects.
The dominant objectives are \emph{pull} and \emph{lift}, while the remaining
objectives include opening and closing articulations, pushing objects away,
placing objects on surfaces, and picking objects. This diversity is useful
because it exposes the robot to related but non-identical physical problems:
for example, pulling a drawer handle, dragging a flat tag, and sliding a shoe
all require horizontal motion but differ in perception, grasping, and contact
geometry.

\begin{figure}[t]
\centering
\includegraphics[width=\linewidth]{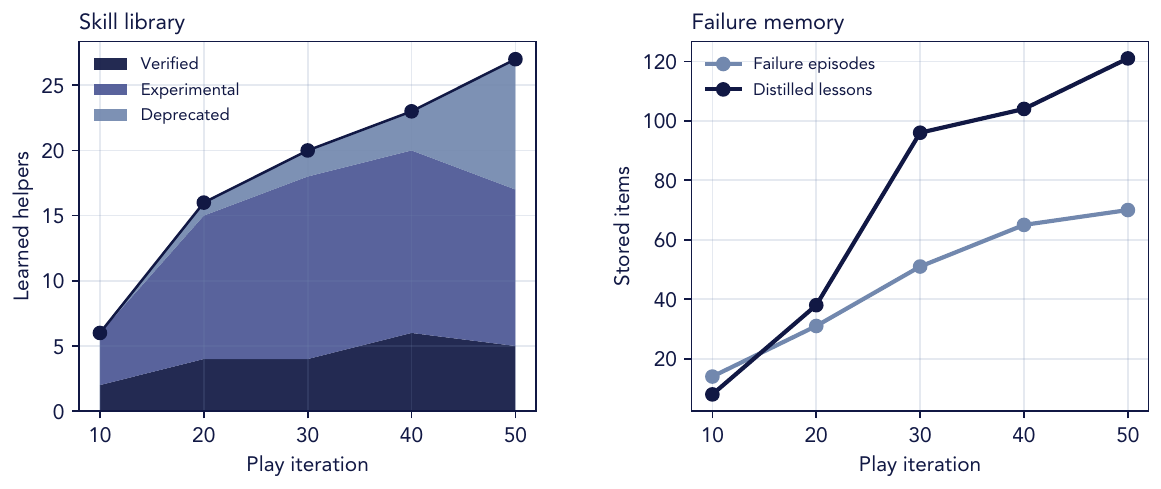}
\vspace{-1.5em}
\caption{
\textbf{Skill library and failure memory growth during
MolmoSpaces play.}
Reports learned skills, verified/experimental/deprecated skill
counts, raw failure episodes, and distilled lessons.
}
\label{fig:play_knowledge_growth}
\vspace{-1em}
\end{figure}

\textbf{Skill and memory growth.}
Play produces two complementary forms
of persistent knowledge. The 50-iteration play run saved its skill library and 
failure memory snapshots every 10 iterations. Figure~\ref{fig:play_knowledge_growth} reports the
saved 10-iteration snapshots. The learned library grows from 6 helpers at
iteration 10 to 27 helpers at iteration 50. Over the same period, failure
memory grows from 14 raw episodes and 8 distilled lessons to 70 episodes and
121 lessons. The three reliability tiers evolve as well: helpers can remain
\emph{experimental}, become \emph{verified} through successful reuse, or be
marked \emph{deprecated} after repeated failures. Thus, play does not merely
append code. It accumulates reusable capabilities while retaining negative
evidence and filtering helpers that do not generalize.

\textbf{Practicing novel yet learnable objectives.}
The proposed objectives are neither a fixed benchmark curriculum nor arbitrary
novelty maximization. At iteration 25, the robot
is asked to slide a black shoe toward itself. This is a new object-skill
combination with novelty score $1.0$ and frontier score $0.6756$. The task
reuses perception and pull helpers learned in earlier iterations, succeeds
after two attempts, and yields a new helper,
\texttt{slide\_grasped\_object\_and\_release}. Five iterations later, the
system reuses this shoe-derived helper to slide a Bluetooth speaker. This
example illustrates the intended behavior: the robot practices a
new objective that is challenging enough to add a capability but close enough
to its current library to remain solvable.

The same pattern appears elsewhere in the run. At iteration 28, a successful
black-metal-rail push yields a helper that is reused for a one-attempt push of
a handheld tool at iteration 35. A drawer-handle helper learned at iteration
43 is reused for a one-attempt drawer pull at iteration 46.

\textbf{Learning from unsuccessful play.}
Failed trajectories also contribute useful supervision. For example, failing to
open a sidetable drawer at iteration 2 produces experimental helpers for
axis-aligned pull-direction estimation and grasp selection. A failed tissue-box
lift at iteration 15 proposes a helper that adjusts a grasp toward the object
centroid. Later failures on a flat knife and an oven drawer propose helpers
for top-down object-aligned grasps and approach-axis filtering, respectively.
These examples illustrate the role of the failure memory and Skill Proposer
described in Sec.~\ref{sec:appendix_skill_memory}: the robot converts recurring
physical bottlenecks into explicit hypotheses for future practice rather than
discarding failed interaction traces.

\subsection{Learned-Skill Usage in MolmoSpaces Evaluation}
\label{sec:appendix_molmospaces_skill_usage}

We next analyze how skills learned from play are reused during evaluation. The test-time execution path
exposes learned non-primitive skills to the policy model and records the
skills that are actually invoked at runtime.

Across 400 evaluation trials, 391 trials invoke at least one learned skill.
The frozen library contains 27 learned skills, of which 14 are invoked during
evaluation, for a total of 5,169 learned-skill calls. Table~\ref{tab:eval_task_skill_usage}
reports performance and skill-call volume by task type. Learned skills are
used in nearly every evaluation trial, but their composition differs by task.
Opening tasks make the heaviest use of the library, averaging $19.0$ helper
calls per trial across 11 distinct learned skills. Pick tasks use a narrower
set of four learned skills, dominated by object localization and top-down grasping.
Pick-and-place tasks require a broader compositional chain and average $14.5$
helper calls per trial.

\begin{table*}[t]
\noindent
\caption{
\textbf{Learned-skill usage by MolmoSpaces evaluation task type.}
The library is learned in MolmoSpaces play and reused in MolmoSpaces
evaluation. Counts in parentheses report runtime invocations of the most-used
learned skills for each task type.
}
\label{tab:eval_task_skill_usage}
\vspace{1.5em}
\begin{minipage}{0.9\textwidth}

\small
\setlength{\tabcolsep}{2pt}
\renewcommand{\arraystretch}{1.15}

\begin{tabularx}{\textwidth}{lrrrrr X}
\toprule
\textbf{Task type} & \textbf{Trials} & \textbf{Success} & \textbf{Skill calls} & \textbf{Unique} & \textbf{Calls/trial} & \textbf{Most-used learned skills} \\
\midrule
Open & 100 & 20\% & 1,900 & 11 & 19.0 &
\texttt{get\_axis\_aligned\_pull\_direction} (614);
\texttt{localize\_and\_verify\_object\_point\_cloud} (429);
\texttt{localize\_object\_with\_molmo\_sam3} (254) \\

Close & 100 & 73\% & 866 & 6 & 8.7 &
\texttt{localize\_object\_with\_molmo\_sam3} (269);
\texttt{get\_axis\_aligned\_pull\_direction} (185);\newline
\texttt{push\_object\_closed} (178); \\

Pick & 100 & 37\% & 950 & 4 & 9.5 &
\texttt{localize\_and\_verify\_object\_point\_cloud} (705);
\texttt{execute\_top\_down\_grasp\_and\_lift} (139);
\texttt{get\_highest\_scoring\_grasp} (97) \\

Pick \& Place & 100 & 22\% & 1,453 & 8 & 14.5 &
\texttt{localize\_and\_verify\_object\_point\_cloud} (649);
\texttt{localize\_object\_with\_molmo\_sam3} (401);
\texttt{execute\_top\_down\_grasp\_and\_lift} (233) \\
\midrule
Total & 400 & 38\% & 5,169 & 14 & 12.9 & -- \\
\bottomrule
\end{tabularx}
\end{minipage}
\end{table*}

\begin{figure*}[t]
\includegraphics[width=0.9\textwidth]{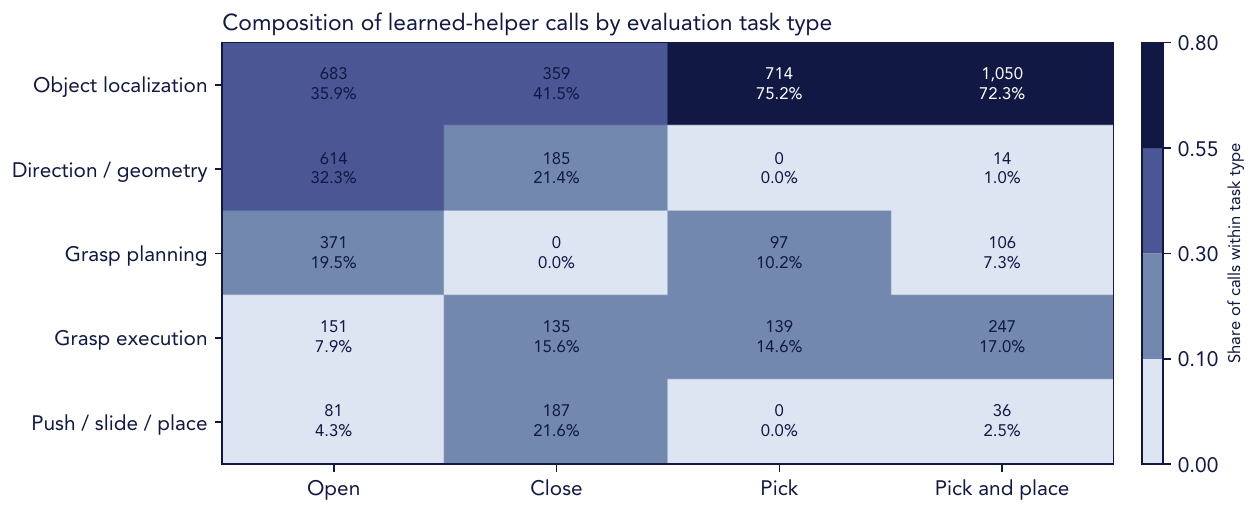}
\caption{
\textbf{Proportion of learned skill calls during MolmoSpaces evaluation.}
Each cell reports the runtime invocation count and the share of learned-skill
calls within that task type. Each column aggregates 100 evaluation trials.
}
\label{fig:eval_skill_categories}
\vspace{-1em}
\end{figure*}

\textbf{Proportions of learned skills.}
Figure~\ref{fig:eval_skill_categories} groups runtime calls by helper category.
Object-localization helpers are the most frequently reused component, with
2,806 invocations. The single most-used helper is
\texttt{localize\_and\_verify\_object\_point\_cloud}, with 1,873 calls.
However, the learned library is not limited to perception. Opening tasks
compose localization with direction estimation and grasp planning:
direction-and-geometry helpers account for $32.3\%$ of open-task calls, and
grasp-planning helpers account for $19.5\%$. Closing tasks use
localization together with push helpers, with push/slide/place helpers
accounting for $21.6\%$ of close-task calls. Pick and pick-and-place tasks are
more perception-heavy: localization accounts for $75.2\%$ and $72.3\%$ of
their calls, respectively. Pick-and-place tasks additionally chain grasp
execution, target localization, and the translation-and-release helper.
The category breakdown makes task-dependent compositions explicit.

\begin{figure*}[t]
\centering
\includegraphics[width=\textwidth]{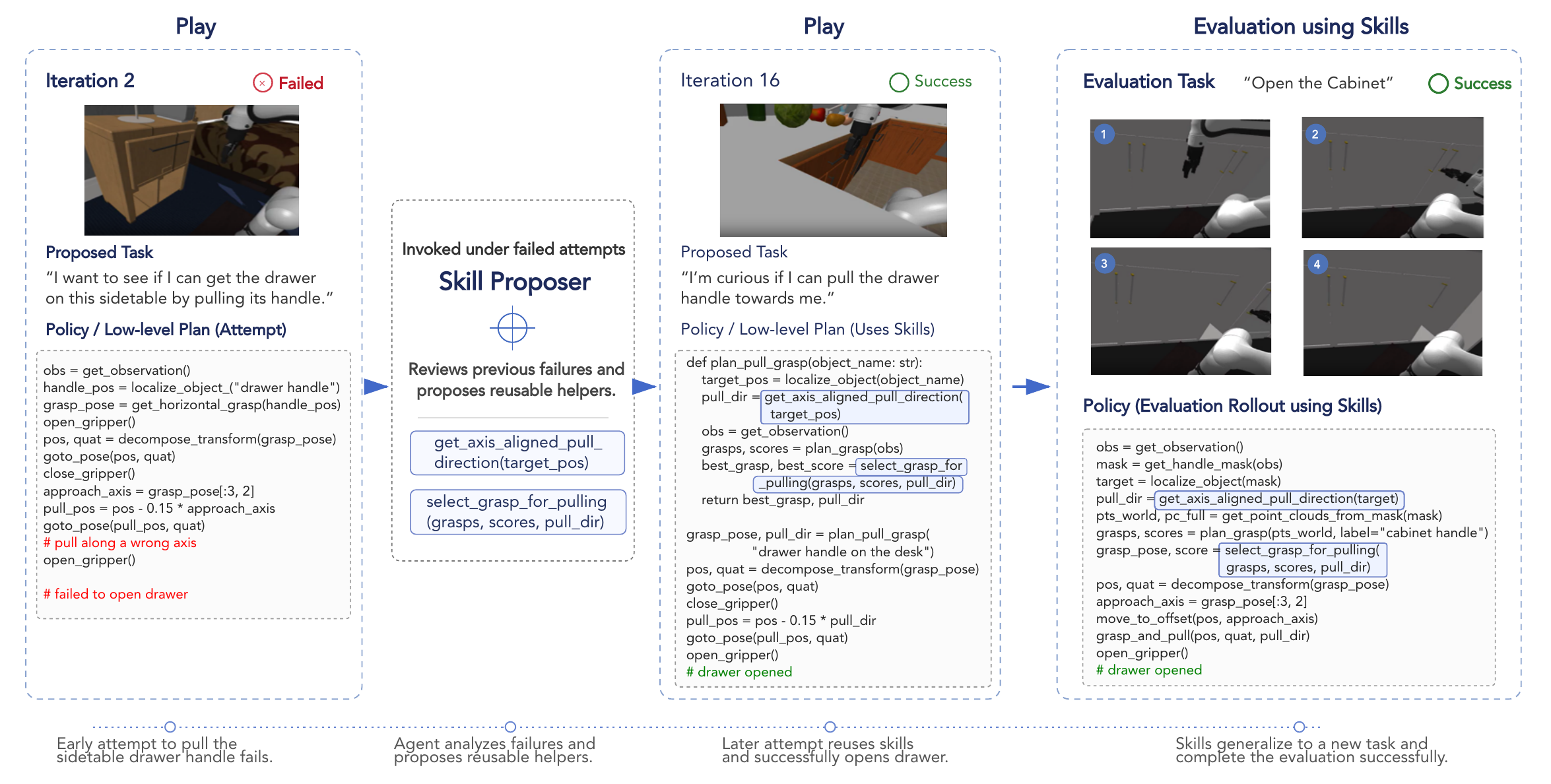}
\vspace{-1em}
\caption{
\textbf{Play-to-evaluation transfer lineage for a successful MolmoSpaces
evaluation trial.}
Play-time tasks lead to learned skills stored in the frozen skill library and then
to their compositional use in evaluation.
}
\label{fig:eval_skill_transfer_lineages}
\end{figure*}

\textbf{Examples of skill usage.}
Figure~\ref{fig:eval_skill_transfer_lineages} traces an evaluation success
back to representative play-time artifacts and the frozen skill library.
The robot opens a cabinet using two learned skills at test time.
The helper \texttt{get\_axis\_aligned\_pull\_direction} is called to
estimate an outward pull vector aligned with the cabinet geometry.
\texttt{select\_grasp\_for\_pulling} is called to select a grasp whose
approach direction is compatible with the intended pull. In the successful
final attempt, these learned components are composed into a single policy step
that localizes the cabinet handle, plans the grasp, grasps the handle, and
pulls the cabinet open.
These helpers were generated by Skill Proposer after a failed play iteration, while
trying to pull a handle on a sidetable. After failing, Skill Proposer reviews the failure
and proposes two reusable helpers, \texttt{get\_axis\_aligned\_pull\_direction} and
\texttt{select\_grasp\_for\_pulling}. These helpers are then reused during a later
play iteration, where the robot tries to pull a drawer handle towards it. This time,
it succeeds by using the learned skills, and therefore increasing the success rate of
those two learned skills. Finally, those two skills are selected and
used during evaluation, leading to a successful attempt.

\subsection{Qualitative Comparison with Direct Code Synthesis}
\label{sec:appendix_capx_comparison}

\begin{figure*}[t]
\centering
\includegraphics[width=\textwidth]{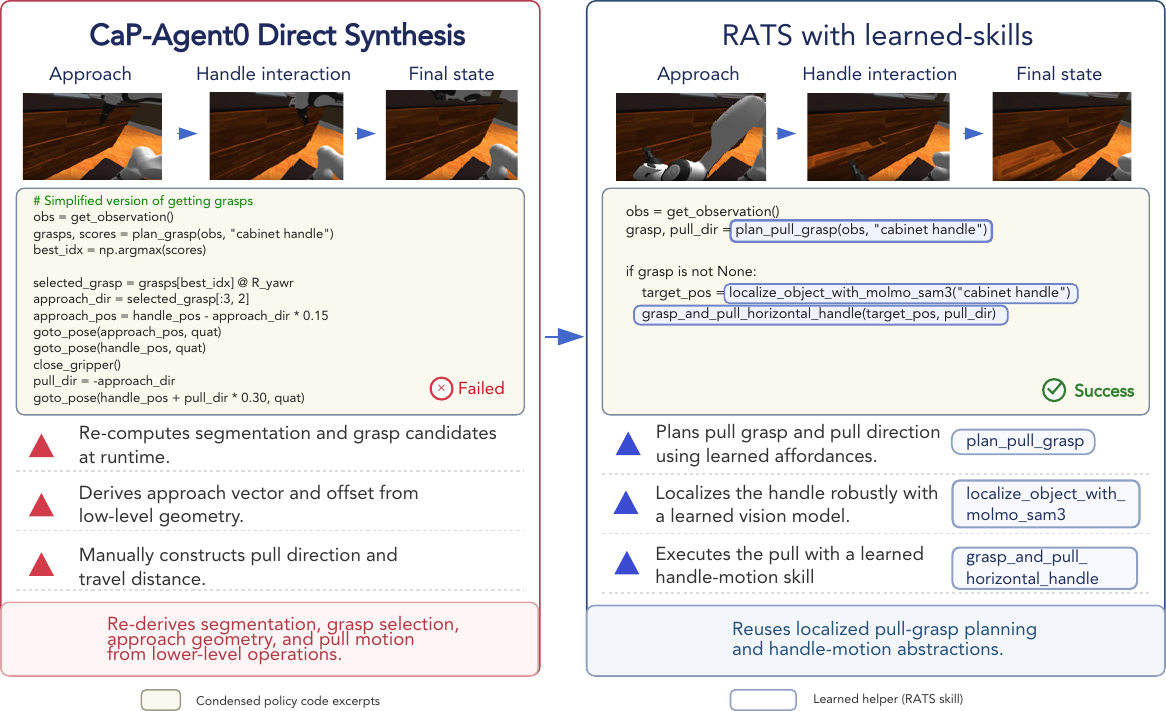}
\vspace{-1em}
\caption{
\textbf{Direct code synthesis versus synthesis with learned skills.}
The condensed source-backed excerpts explain why learned skills reduce
brittle low-level reasoning.
}
\label{fig:capx_code_comparison}
\vspace{-1em}
\end{figure*}

Figure~\ref{fig:capx_code_comparison} compares archived videos and code for the same
iTHOR drawer-opening objective. With the learned library, the robot approaches
the handle and leaves the drawer open. The direct-synthesis run without our learned
library does not complete the articulation.
Direct synthesis must repeatedly reconstruct grasp selection, approach geometry, and pull direction from low-level operations. The learned-library path instead composes reusable
localization, pull-grasp planning, and handle-motion helpers.

\begin{figure*}[t]
\centering
\includegraphics[width=\textwidth]{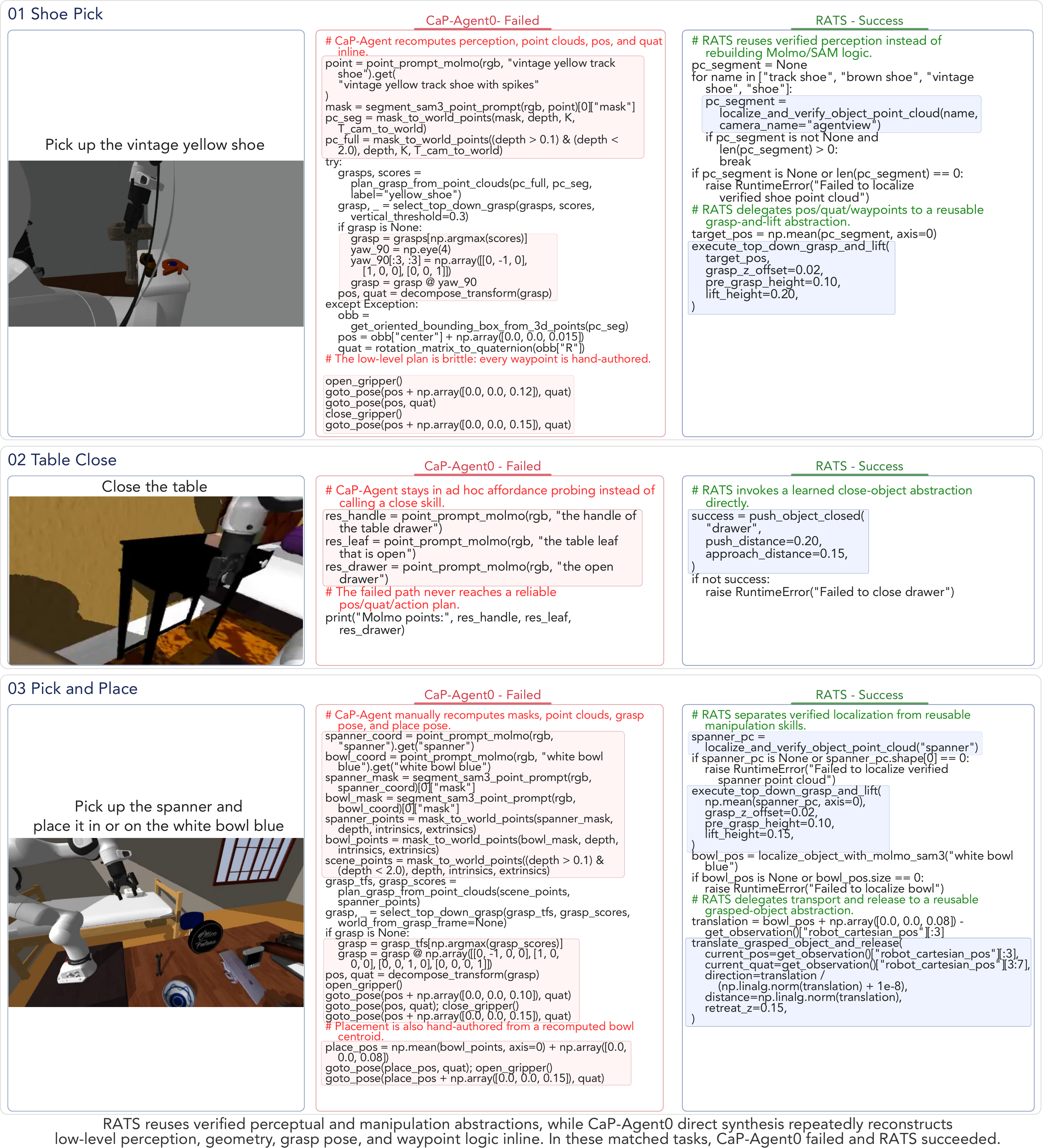}
\vspace{-1em}
\caption{
\textbf{More qualitative comparisons between direct code synthesis and \textsc{RATs} with
learned skills.}
Across shoe picking, table closing, and pick-and-place tasks, the \textsc{CaP-Agent0}
direct-synthesis policies fail while the \textsc{RATs} policies succeed. \textsc{CaP-Agent0} 
recomputes perception, geometry, \texttt{pos}/\texttt{quat}, and waypoint logic
inline; \textsc{RATs} calls learned skills for verified localization, grasping,
closing, transport, and release.
}
\label{fig:more_capx_code_comparison}
\end{figure*}

Below, we show more examples comparing direct code synthesis using \textsc{CaP-Agent0} against code synthesis using our learned skills. Figure~\ref{fig:more_capx_code_comparison} compares direct code synthesis against
synthesis with learned skills on three matched MolmoSpaces objectives:
\emph{Pick up the vintage yellow track shoe},
\emph{close the table}, and
\emph{Pick up the spanner and place it in or on the white bowl blue}.
For each objective, the direct-synthesis run fails while the \textsc{RATs} run succeeds.
The comparison shows the same qualitative pattern across task families. Direct
synthesis repeatedly reconstructs task state from low-level operations:
querying Molmo points, segmenting masks with SAM, converting masks to point
clouds, selecting or repairing grasp transforms, decomposing them into
\texttt{pos} and \texttt{quat}, and then hand-writing waypoint chains. In the
shoe-picking example, \textsc{CaP-Agent0} recomputes the shoe mask, full scene point cloud,
grasp transform, fallback yaw, and lift waypoint inline. \textsc{RATs} instead composes
\texttt{localize\_and\_verify\_object\_point\_cloud} with
\texttt{execute\_top\_down\_grasp\_and\_lift}.

The closing and pick-and-place examples highlight the same distinction at
different temporal scales. For closing the table, the failed direct-synthesis
code remains at the level of ad hoc affordance probes for handles, leaves, and
drawers, whereas \textsc{RATs} invokes the learned
\texttt{push\_object\_closed} helper. For pick-and-place, \textsc{CaP-Agent0} manually builds
the spanner and bowl masks, recomputes point clouds, plans a grasp, computes a
bowl centroid, and writes the release trajectory. \textsc{RATs} composes verified object
localization, \texttt{execute\_top\_down\_grasp\_and\_lift}, target
localization, and \texttt{translate\_grasped\_object\_and\_release}. These
examples illustrate why the learned library improves reliability: successful
interaction routines become named, reusable control abstractions instead of
being rediscovered from pixels and geometry in every episode.

\subsection{Usage of LIBERO-Derived Skills in RoboSuite Evaluation}
\label{sec:appendix_robosuite_skill_usage}

\begin{figure*}[t]
  \centering
  \includegraphics[width=\textwidth]{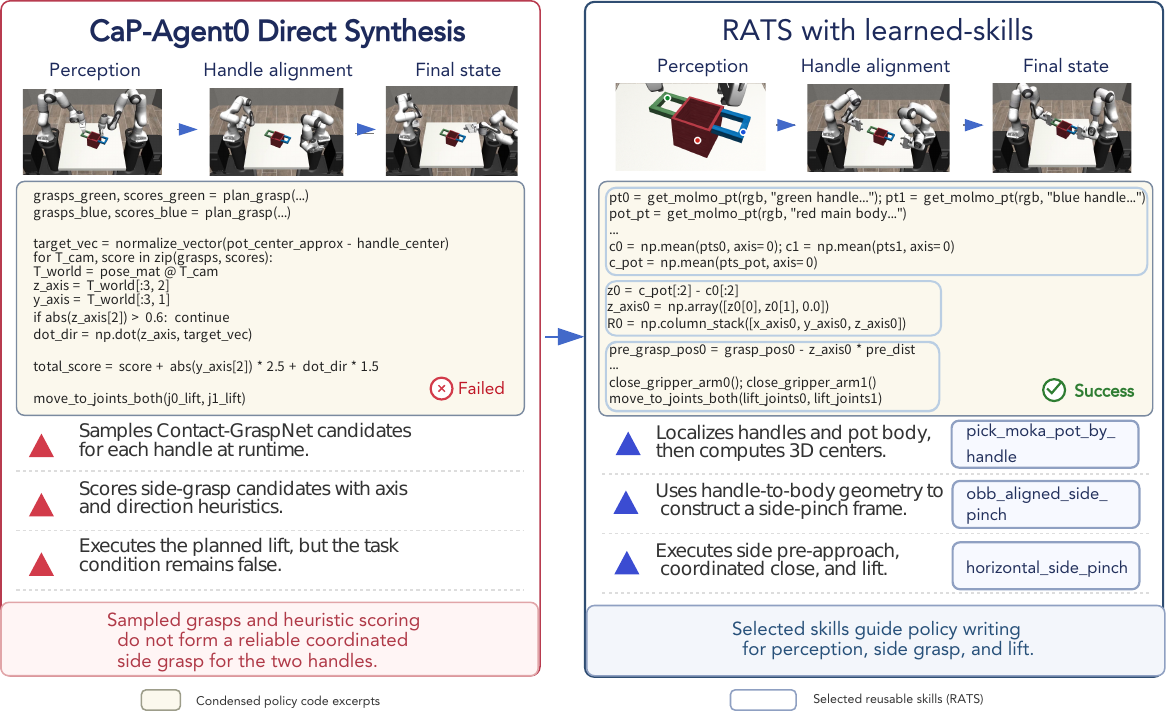}
  \caption{
  \textbf{LIBERO-to-RoboSuite skill transfer in two-arm lifting.}
  For the same RoboSuite \texttt{two\_arm\_lift} evaluation seed, direct code
  synthesis fails while RATS succeeds by reusing skills selected from a
  LIBERO-derived library.
  }
  \label{fig:libero_robosuite_code_comparison}
  \end{figure*}
  
Figure~\ref{fig:libero_robosuite_code_comparison} compares archived videos and generated code for the same RoboSuite \texttt{two\_arm\_lift} objective. The direct-synthesis run without the learned library samples Contact-GraspNet candidates for each handle and scores them using axis and direction heuristics. Although the generated policy reaches the lift stage, it does not complete the task.

With RATS, the policy writer selects relevant skills from a frozen library learned from LIBERO play-time tasks. These skills are not invoked as explicit API calls in the final RoboSuite policy. Instead, they are reused implicitly: the generated code follows the same structure of localizing task-relevant parts, computing 3D handle and object centers, constructing a side-pinch grasp frame from handle-to body geometry, and executing a coordinated close-and-lift motion. This illustrates cross-environment transfer from LIBERO skills to a RoboSuite evaluation task.

\section{Details about the Evaluation Benchmark}
\label{sec:appendix_evaluation_benchmarks}

During evaluation, the play-time skill library is frozen and the system solves
externally specified tasks. This section records the benchmark composition and
replay protocol for the two simulated evaluation domains.

\subsection{MolmoSpaces Evaluation Benchmark}
\label{sec:appendix_molmospaces_benchmark}

For MolmoSpaces, we evaluate on a compact held-out subset of the
benchmark. The subset contains
40 test-split episodes: ten each for opening, closing, picking, and
pick-and-place. It was constructed by randomly selecting ten
episodes per task type. It covers four components of the upstream
benchmark, namely Open-v1, Close-v1, Pick-v2-classic, and PnP-v2.

\begin{table}[t]
\centering
\caption{
\textbf{MolmoSpaces held-out core evaluation subset.}
The main benchmark sweep repeats each selected task ten times.
}
\label{tab:molmospaces_eval_benchmark}
\begin{tabular}{llrr}
\toprule
\textbf{Task type} & \textbf{Scene dataset} & \textbf{Tasks} & \textbf{Main rollouts} \\
\midrule
Open & iTHOR & 10 & 100 \\
Close & iTHOR & 10 & 100 \\
Pick & ProcTHOR-Objaverse & 10 & 100 \\
pick-and-place & ProcTHOR-Objaverse & 10 & 100 \\
\midrule
Total & -- & 40 & 400 \\
\bottomrule
\end{tabular}

\end{table}

The selected episodes span 37 houses. Each episode's JSON stores the scene
dataset and house index, the Franka initialization, object poses, task state,
natural-language referral expressions, and camera specification. Success is evaluated by the simulator task's \texttt{judge\_success()} predicate: articulated tasks check the requested
joint state, pick tasks, check the object target state, and pick-and-place tasks
check placement support on the receptacle.

The benchmark definition is independent of the number of repeated trials. The
main sweep uses ten trials per task, for $40 \times 10 = 400$ rollouts.

\subsection{LIBERO-PRO Evaluation Benchmark}
\label{sec:appendix_libero_pro_benchmark}

For LIBERO-PRO, we evaluate the object, goal, and spatial
categories under two perturbation types. The \emph{Pos} setting corresponds to
the \texttt{swap} suite, which perturbs object positions. The \emph{Task}
setting corresponds to the \texttt{task} suite, which perturbs the task
specification. Each suite contains ten language-conditioned manipulation
tasks.

\begin{table}[t]
\centering
\caption{
\textbf{LIBERO-PRO evaluation.}
Reported comparisons use ten initial-state trials per task.
}
\label{tab:libero_pro_eval_benchmark}
\begin{tabular}{llrr}
\toprule
\textbf{Setting} & \textbf{LIBERO-PRO suite} & \textbf{Tasks} & \textbf{Reported rollouts} \\
\midrule
Object Pos & \texttt{libero\_object\_swap} & 10 & 100 \\
Object Task & \texttt{libero\_object\_task} & 10 & 100 \\
Goal Pos & \texttt{libero\_goal\_swap} & 10 & 100 \\
Goal Task & \texttt{libero\_goal\_task} & 10 & 100 \\
Spatial Pos & \texttt{libero\_spatial\_swap} & 10 & 100 \\
Spatial Task & \texttt{libero\_spatial\_task} & 10 & 100 \\
\midrule
Total & -- & 60 & 600 \\
\bottomrule
\end{tabular}

\end{table}

LIBERO-PRO provides 50 initial states for each task. A complete sweep over all
available states therefore contains $6 \times 10 \times 50 = 3{,}000$
rollouts. For the reported comparisons, we cap evaluation at ten
initial-state trials per task, yielding $6 \times 10 \times 10 = 600$ rollouts
for each evaluated setup. The baseline and learned-library conditions use the
same task and trial budget.

\section{Additional Studies}
\label{sec:appendix_additional_studies}

\subsection{Ablation Results on MolmoSpaces}

We further ablate the effect of play-learned skills and the test-time execution system on
MolmoSpaces. This setting complements the LIBERO-PRO ablation in the main paper by evaluating
whether the same play-time skill acquisition mechanism remains useful in a scene-grounded
benchmark with open, close, pick, and pick-and-place task categories. As in the main ablation, all
play-based variants use 50 play iterations, and all play-time skills are learned by our proposed
\textsc{RATs} system.

Table~\ref{tab:molmospaces-ablation} shows that play-learned skills improve performance even when
they are only plugged into the standard \textsc{CaP-Agent0} test-time agent. Under \textsc{CaP-Agent0},
Curious Play increases the average success rate from 21.0\% to 25.8\%, with the largest gain on
closing tasks. When the same learned skills are used by the full \textsc{RATs} execution system,
performance improves more substantially, from 32.8\% without play to 38.0\% with Curious Play.
These results suggest that play-time skill learning and the structured \textsc{RATs} test-time
execution system provide complementary benefits: the learned skill library alone can improve a
standard Code-as-Policy agent, while the full execution system is better able to retrieve, verify, and
compose those skills for downstream tasks.

\begin{table}[t]
    \centering
    \setlength{\tabcolsep}{3.5pt}
    \caption{\textbf{MolmoSpaces ablation over play strategy and test-time system.} All
    play-based variants use 50 play iterations. All play-time skills are learned by our proposed \textsc{RATs} system.}
    \label{tab:molmospaces-ablation}
    \resizebox{\ablationTableWidth}{!}{%
    \begin{tabular}{llccccc}
        \toprule
        Test-Time System
        & Play-Time Skills
        & Open
        & Close
        & Pick
        & pick-and-place
        & Avg. \\
        \midrule
        \multirow{2}{*}{\textsc{CaP-Agent0}}
            & No Play
             & 14.0 & 36.0 & 23.0 & 11.0 & 21.0 \\
            & Curious Play
             & 17.0 & 62.0 & 14.0 & 10.0 & 25.8 \\
        \midrule
        \multirow{2}{*}{\textsc{RATs Exec.}}
            & No Play
             & 11.0 & 65.0 & \textbf{45.0} & 10.0 & 32.8 \\
            & Curious Play
             & \textbf{20.0} & \textbf{73.0} & 37.0 & \textbf{22.0} & \textbf{38.0} \\
        \bottomrule
    \end{tabular}
    }
\end{table}

\subsection{Token Cost Analysis}
\label{sec:token_cost_analysis}

Because \textsc{RATs} relies heavily on Large Language Models (LLMs) for task proposal, planning,
code generation, and failure diagnosis, we provide a detailed analysis of the token consumption
during both the autonomous play phase and the test-time evaluation phase. All LLM calls documented
in this section utilized \texttt{gpt-5.5}.

\textbf{Play-Time Token Consumption.} We analyzed a play-mode run in the
\texttt{libero\_spatial} environment. Over the play iterations, where each iteration is budgeted with
a maximum of five attempts, token consumption is heavily skewed toward failed iterations that
exhaust the retry budget. A component-wise breakdown in Table~\ref{tab:play_time_tokens} shows
that the Write-Execute-Verify-Diagnose loop is the primary cost driver. This indicates that while
intrinsic task proposal is relatively token-efficient, extracting diagnostics and revising policies from
repeated physical failures remains the dominant computational expense during play.

\begin{table}[htbp]
    \centering
    \caption{\textbf{Play-time token consumption by each agent component (10 iterations).}}
    \label{tab:play_time_tokens}
    \begin{tabular}{lrr}
        \toprule
        \textbf{Agent Role} & \textbf{Total Tokens} & \textbf{Share (\%)} \\
        \midrule
        Failure Diagnoser & 2,071,881 & 40.5\% \\
        Policy Writer & 1,472,949 & 28.8\% \\
        Failure Memory Distillation & 993,770 & 19.4\% \\
        Verifier & 194,318 & 3.8\% \\
        Memory Curator & 111,563 & 2.2\% \\
        Task Proposer & 98,505 & 1.9\% \\
        Planner & 78,359 & 1.5\% \\
        Environment Creator & 31,616 & 0.6\% \\
        Skill Library (Duplicate Check) & 31,377 & 0.6\% \\
        Skill Proposer & 30,969 & 0.6\% \\
        Feedback Generator & 5,875 & 0.1\% \\
        \midrule
        \textbf{Total} & \textbf{5,121,182} & \textbf{100.0\%} \\
        \bottomrule
    \end{tabular}
\end{table}

\textbf{Compute-Matched Test-Time Comparison.} To ensure a fair comparison, we account for the
upfront token cost incurred by \textsc{RATs} during autonomous play. A full 50-iteration play phase
consumes approximately 30M tokens. We amortize this play-time cost across the 60 held-out tasks in
LIBERO-PRO.

The baseline \textsc{CaP-Agent0} consumes approximately 1.6M tokens per task under a standard
10-turn retry budget. Adding the amortized play-time token cost of \textsc{RATs} to this baseline
grants \textsc{CaP-Agent0} a larger test-time compute budget, allowing it to run for roughly 15 turns
per task. We therefore evaluate a compute-matched baseline, \textsc{CaP-Agent0} with a 15-turn
budget, and compare it against the standard 10-turn \textsc{CaP-Agent0} agent augmented with the
skill library learned by \textsc{RATs} through Curious Play. This comparison isolates whether the
same additional token budget is more useful when spent reactively on extra test-time retries or
proactively on play-time skill acquisition.

\begin{table}[htbp]
    \centering
    \caption{\textbf{Compute-matched performance comparison on LIBERO-PRO.} The
    \textsc{CaP-Agent0} (15 turns) baseline is granted additional retry budget at test time to match
    the amortized token cost that \textsc{RATs} spent during play-time. The \textsc{CaP-Agent0} +
    \textsc{RATs} Skills row uses the standard 10-turn \textsc{CaP-Agent0} test-time system, with
    skills learned from 50 iterations of Curious Play.}
    \label{tab:compute_matched_comparison}
    \resizebox{\textwidth}{!}{%
    \begin{tabular}{lccccccr}
        \toprule
        \multirow{2}{*}{\textbf{Method}} & \multicolumn{2}{c}{\textbf{Object}} & \multicolumn{2}{c}{\textbf{Goal}} & \multicolumn{2}{c}{\textbf{Spatial}} & \multirow{2}{*}{\textbf{Avg.}} \\
        \cmidrule(lr){2-3} \cmidrule(lr){4-5} \cmidrule(lr){6-7}
        & Pos. & Task & Pos. & Task & Pos. & Task & \\
        \midrule
        \textsc{CaP-Agent0} (10 turns) & 27.0 & 31.0 & 29.0 & 16.0 & 13.0 & 23.0 & 23.2 \\
        \textsc{CaP-Agent0} (15 turns, compute-matched) & 28.0 & 32.0 & 30.0 & \textbf{24.0} & \textbf{22.0} & 20.0 & 26.0 \\
        \textsc{CaP-Agent0} (10 turns) + \textsc{RATs} Skills from Play
        & \textbf{51.0} & \textbf{47.0} & \textbf{34.0} & 20.0 & 19.0 & \textbf{23.0} & \textbf{32.3} \\
        \bottomrule
    \end{tabular}%
    }
\end{table}

As shown in Table~\ref{tab:compute_matched_comparison}, simply allocating more test-time compute
to the baseline yields only modest improvements, increasing average success from 23.2\% to 26.0\%.
In contrast, spending the comparable compute budget proactively during play-time and then reusing
the learned skills with the same 10-turn \textsc{CaP-Agent0} test-time system improves average
success to 32.3\%. This result indicates that the gain does not come merely from giving the agent
more inference-time retries. Instead, play-time computation is more effective when it is distilled into
a reusable skill library that can be retrieved and composed for held-out tasks.

\definecolor{RATSCodeBg}{rgb}{0.9686,0.9804,0.9922}
\definecolor{RATSCodeKeyword}{rgb}{0.0667,0.0941,0.2667}
\definecolor{RATSCodeString}{rgb}{0.2941,0.3373,0.5804}
\definecolor{RATSCodeComment}{rgb}{0.4471,0.5333,0.6824}
\definecolor{RATSCodeRule}{rgb}{0.9176,0.8784,0.8118}
\lstdefinestyle{ratsappendixpython}{
  language=Python,
  basicstyle=\ttfamily\scriptsize,
  keywordstyle=\color{RATSCodeKeyword}\bfseries,
  stringstyle=\color{RATSCodeString},
  commentstyle=\color{RATSCodeComment}\itshape,
  numberstyle=\tiny\color{RATSCodeComment},
  backgroundcolor=\color{RATSCodeBg},
  rulecolor=\color{RATSCodeRule},
  frame=single,
  framerule=0.25pt,
  numbers=left,
  numbersep=4pt,
  columns=fullflexible,
  keepspaces=true,
  showstringspaces=false,
  breaklines=true,
  breakatwhitespace=false,
  tabsize=4
}

\subsection{MolmoSpaces Learned Skills}
\label{sec:appendix_molmospaces_learned_helper_code}

The following syntax-highlighted code block records three representative learned
skills from the MolmoSpaces skill library.

\begin{lstlisting}[style=ratsappendixpython]
# MolmoSpaces representative learned-skill snippets
# Selected skills: 3 of 27

#===================================================================================================
# Skill 1/3: get_axis_aligned_pull_direction
# Interface: def get_axis_aligned_pull_direction(target_pos: np.ndarray) -> np.ndarray:
# Documentation string: Computes the axis-aligned pull direction (e.g., [1, 0, 0] or [0, -1, 0])
#                       pointing from the target position towards the robot base, useful for
#                       determining which way a drawer or cabinet door opens.
# Tier: verified; uses: 30; successes: 10
#===================================================================================================
def get_axis_aligned_pull_direction(target_pos: np.ndarray) -> np.ndarray:
    import numpy as np
    obs = get_observation()
    robot_pos = obs["robot_base_pose"][:3]
    direction = robot_pos - target_pos
    direction[2] = 0
    axis_idx = np.argmax(np.abs(direction))
    pull_dir = np.zeros(3)
    pull_dir[axis_idx] = np.sign(direction[axis_idx]) if direction[axis_idx] != 0 else 1.0
    return pull_dir

#===================================================================================================
# Skill 2/3: push_object_closed
# Interface: def push_object_closed(object_name: str, push_distance: float = 0.15,
#            approach_distance: float = 0.15) -> bool:
# Documentation string: Localizes an open object (like a drawer or door), determines its outward
#                       pull direction, and pushes it inward to close it.
# Tier: experimental; uses: 2; successes: 2
#===================================================================================================
def push_object_closed(object_name: str, push_distance: float = 0.15, approach_distance: float = 0.15) -> bool:
    pos = localize_object_with_molmo_sam3(object_name)
    if pos is None:
        return False
    pull_dir = get_axis_aligned_pull_direction(pos)
    if pull_dir is None:
        return False
    close_gripper()
    return push_surface_inward(pos, pull_dir, push_distance=push_distance, approach_distance=approach_distance)

#===================================================================================================
# Skill 3/3: plan_top_down_grasp_at_wrist
# Interface: def plan_top_down_grasp_at_wrist(object_name: str, target_world_pos: np.ndarray,
#            hover_height: float = 0.15) -> np.ndarray:
# Documentation string: Moves the wrist camera to hover over a target position, uses Molmo and SAM3
#                       to segment the object, and plans a top-down grasp from the resulting point
#                       clouds.
# Tier: deprecated; uses: 26; successes: 6
#===================================================================================================
def plan_top_down_grasp_at_wrist(object_name: str, target_world_pos: np.ndarray, hover_height: float = 0.15) -> np.ndarray:
    wrist_obs = inspect_at_wrist(target_world_pos, hover_height=hover_height)
    if not wrist_obs["ok"]:
        return None

    w_rgb = wrist_obs["wrist"]["rgb"]
    w_depth = wrist_obs["wrist"]["depth"]
    w_K = wrist_obs["wrist"]["intrinsics"]
    w_T = wrist_obs["wrist"]["pose_mat"]

    pt_w = point_prompt_molmo(w_rgb, object_name)
    u_w, v_w = pt_w.get(object_name, (None, None))
    if u_w is None:
        return None

    w_masks = segment_sam3_point_prompt(w_rgb, (u_w, v_w))
    if len(w_masks) == 0:
        return None

    w_valid_mask = w_masks[0]["mask"]
    pc_segment = mask_to_world_points(w_valid_mask, w_depth, w_K, w_T)
    
    full_mask = (w_depth > 0) & (w_depth < 2.0)
    pc_full = mask_to_world_points(full_mask, w_depth, w_K, w_T)

    if len(pc_segment) == 0 or len(pc_full) == 0:
        return None

    pc_full = subsample_point_cloud(pc_full, 10000)
    pc_segment = subsample_point_cloud(pc_segment, 5000)

    grasps, scores = plan_grasp_from_point_clouds(pc_full, pc_segment, object_name)
    if len(grasps) == 0:
        return None

    best_grasp, best_score = select_top_down_grasp(grasps, scores)
    return best_grasp
\end{lstlisting}

\subsection{LIBERO Learned Skills}
\label{sec:appendix_libero_learned_helper_code}

The following syntax-highlighted code block records three representative learned
skills from the LIBERO skill library.

\begin{lstlisting}[style=ratsappendixpython]
# LIBERO representative learned-skill snippets
# Selected skills: 3 of 47

#===================================================================================================
# Skill 1/3: localize_object_with_pose_aliases_and_segmentation_fallback
# Interface: def localize_object_with_pose_aliases_and_segmentation_fallback(object_names,
#            segmentation_query, use_multiview=True):
# Documentation string: Localizes a target object by trying one or more pose-query names, then falls
#                       back to language segmentation and an oriented bounding box center if pose
#                       queries fail.
# Tier: verified; uses: 18; successes: 16
#===================================================================================================
def localize_object_with_pose_aliases_and_segmentation_fallback(object_names, segmentation_query, use_multiview=True):
    position = None
    quaternion = None
    method = None
    point_count = None
    obb = None

    for object_name in object_names:
        if position is None:
            position, quaternion = get_object_pose(object_name, use_multiview=use_multiview)
            method = "object_pose_" + object_name

    if position is None:
        seg = get_object_3d_points_and_masks_from_language(segmentation_query, use_multiview=use_multiview)
        points = seg.get("points_3d")
        if points is not None and len(points) > 0:
            point_count = len(points)
            obb = get_oriented_bounding_box_from_3d_points(points)
            position = obb["center"]
            quaternion = None
            method = "segmentation_obb_" + segmentation_query

    return {
        "position": position,
        "quaternion": quaternion,
        "method": method,
        "point_count": point_count,
        "obb": obb
    }

#===================================================================================================
# Skill 2/3: pick_flat_object_with_obb_top_down_retries
# Interface: def pick_flat_object_with_obb_top_down_retries(target_prompt, verification_names,
#            approach_height=0.10, lift_height=0.05, top_clearance_attempts=(-0.02, -0.025),
#            xy_offset_attempts=((-0.015, -0.015), (-0.02, 0.0)), use_multiview=True,
#            min_lift_delta=0.015):
# Documentation string: Segments a flat object, computes an oriented-bounding-box top-down grasp
#                       pose, tries configurable top-clearance and XY-offset corrections, lifts, and
#                       verifies that the object rose.
# Tier: experimental; uses: 1; successes: 1
#===================================================================================================
def pick_flat_object_with_obb_top_down_retries(target_prompt, verification_names, approach_height=0.10, lift_height=0.05, top_clearance_attempts=(-0.02, -0.025), xy_offset_attempts=((-0.015, -0.015), (-0.02, 0.0)), use_multiview=True, min_lift_delta=0.015):
    attempt_records = []
    last_grasp_position = None
    last_grasp_quaternion = None
    last_lift_position = None
    last_obb = None
    last_segmentation = None

    num_attempts = min(len(top_clearance_attempts), len(xy_offset_attempts))
    for attempt_index in range(num_attempts):
        open_gripper()

        segmentation = get_object_3d_points_and_masks_from_language(target_prompt, use_multiview=use_multiview)
        points = segmentation.get("points_3d", None)
        if points is None or len(points) == 0:
            attempt_records.append({
                "attempt_index": attempt_index,
                "top_clearance": top_clearance_attempts[attempt_index],
                "xy_offset": xy_offset_attempts[attempt_index],
                "grasp_position": None,
                "grasp_quaternion_wxyz": None,
                "lift_position": None,
                "verified_position": None,
                "verified_lifted": False,
            })
            continue

        obb = get_oriented_bounding_box_from_3d_points(points)
        center = obb.get("center", None)
        extent = obb.get("extent", None)
        if center is None:
            attempt_records.append({
                "attempt_index": attempt_index,
                "top_clearance": top_clearance_attempts[attempt_index],
                "xy_offset": xy_offset_attempts[attempt_index],
                "grasp_position": None,
                "grasp_quaternion_wxyz": None,
                "lift_position": None,
                "verified_position": None,
                "verified_lifted": False,
            })
            continue

        grasp_position = center.copy()
        if extent is not None and len(extent) >= 3:
            grasp_position[2] = center[2] + float(extent[2]) * 0.5 + float(top_clearance_attempts[attempt_index])
        else:
            grasp_position[2] = center[2] + float(top_clearance_attempts[attempt_index])

        dx, dy = xy_offset_attempts[attempt_index]
        grasp_position[0] = grasp_position[0] + float(dx)
        grasp_position[1] = grasp_position[1] + float(dy)

        grasp_quaternion = (0.0, 0.0, 1.0, 0.0)
        if extent is not None and len(extent) >= 2 and float(extent[0]) > float(extent[1]):
            grasp_quaternion = (0.0, -0.70710678, 0.70710678, 0.0)

        goto_pose(grasp_position, grasp_quaternion, z_approach=approach_height)
        close_gripper()

        lift_position = grasp_position.copy()
        lift_position[2] = lift_position[2] + float(lift_height)
        goto_pose(lift_position, grasp_quaternion)

        verified_position = None
        verified_quaternion = None
        for verification_name in verification_names:
            verified_position, verified_quaternion = get_object_pose(verification_name, use_multiview=use_multiview)
            if verified_position is not None:
                break

        verified_lifted = False
        if verified_position is not None:
            verified_lifted = float(verified_position[2]) >= float(grasp_position[2]) + float(min_lift_delta)

        last_grasp_position = grasp_position
        last_grasp_quaternion = grasp_quaternion
        last_lift_position = lift_position
        last_obb = obb
        last_segmentation = segmentation

        attempt_records.append({
            "attempt_index": attempt_index,
            "top_clearance": top_clearance_attempts[attempt_index],
            "xy_offset": xy_offset_attempts[attempt_index],
            "grasp_position": grasp_position.tolist() if hasattr(grasp_position, "tolist") else grasp_position,
            "grasp_quaternion_wxyz": grasp_quaternion,
            "lift_position": lift_position.tolist() if hasattr(lift_position, "tolist") else lift_position,
            "verified_position": verified_position.tolist() if hasattr(verified_position, "tolist") else verified_position,
            "verified_lifted": verified_lifted,
        })

        if verified_lifted:
            return {
                "success": True,
                "grasp_position": grasp_position,
                "grasp_quaternion": grasp_quaternion,
                "lift_position": lift_position,
                "obb": obb,
                "segmentation": segmentation,
                "attempt_records": attempt_records,
            }

    return {
        "success": False,
        "grasp_position": last_grasp_position,
        "grasp_quaternion": last_grasp_quaternion,
        "lift_position": last_lift_position,
        "obb": last_obb,
        "segmentation": last_segmentation,
        "attempt_records": attempt_records,
    }

#===================================================================================================
# Skill 3/3: place_carried_object_on_segmented_target_with_hover_correction
# Interface: def place_carried_object_on_segmented_target_with_hover_correction(target_prompt,
#            initial_target_center, placement_quaternion, hover_z_offset=0.18,
#            release_z_offset=0.05, retreat_z_offset=0.21, use_multiview=True):
# Documentation string: Places an already-grasped object onto a target surface/region by moving to
#                       an initial hover above the target, re-segmenting the target from the hover
#                       view, correcting XY placement, descending to release, opening the gripper,
#                       and retreating upward.
# Tier: verified; uses: 7; successes: 6
#===================================================================================================
def place_carried_object_on_segmented_target_with_hover_correction(target_prompt, initial_target_center, placement_quaternion, hover_z_offset=0.18, release_z_offset=0.05, retreat_z_offset=0.21, use_multiview=True):
    def _coord(position, index):
        if position is None:
            return None
        if hasattr(position, "tolist"):
            position = position.tolist()
        try:
            return position[index]
        except Exception:
            return None

    def _make_position_like(reference_position, x_value, y_value, z_value):
        if reference_position is not None and hasattr(reference_position, "copy"):
            new_position = reference_position.copy()
            new_position[0] = float(x_value)
            new_position[1] = float(y_value)
            new_position[2] = float(z_value)
            return new_position
        return [float(x_value), float(y_value), float(z_value)]

    def _position_with_z_offset(center, z_offset):
        x = _coord(center, 0)
        y = _coord(center, 1)
        z = _coord(center, 2)
        if x is None or y is None or z is None:
            return None
        return _make_position_like(center, x, y, float(z) + float(z_offset))

    def _replace_xy_keep_z(base_position, xy_source_position):
        bx = _coord(base_position, 0)
        by = _coord(base_position, 1)
        bz = _coord(base_position, 2)
        sx = _coord(xy_source_position, 0)
        sy = _coord(xy_source_position, 1)
        if bx is None or by is None or bz is None or sx is None or sy is None:
            return base_position
        return _make_position_like(base_position, sx, sy, bz)

    def _xy_distance(position_a, position_b):
        ax = _coord(position_a, 0)
        ay = _coord(position_a, 1)
        bx = _coord(position_b, 0)
        by = _coord(position_b, 1)
        if ax is None or ay is None or bx is None or by is None:
            return None
        return ((float(ax) - float(bx)) ** 2 + (float(ay) - float(by)) ** 2) ** 0.5

    def _position_within_table_bounds(position):
        x = _coord(position, 0)
        y = _coord(position, 1)
        z = _coord(position, 2)
        if x is None or y is None or z is None:
            return False
        return (-0.45 <= float(x) <= 0.45) and (-0.55 <= float(y) <= 0.55) and (0.60 <= float(z) <= 1.05)

    def _segment_target_center(prompt, multiview):
        segmentation = get_object_3d_points_and_masks_from_language(prompt, use_multiview=multiview)
        points = None
        if isinstance(segmentation, dict) and "points_3d" in segmentation:
            points = segmentation["points_3d"]
        if points is None:
            return None, segmentation, None
        try:
            if len(points) == 0:
                return None, segmentation, None
        except Exception:
            return None, segmentation, None
        obb = get_oriented_bounding_box_from_3d_points(points)
        center = None
        if isinstance(obb, dict) and "center" in obb:
            center = obb["center"]
        return center, segmentation, obb

    log = {
        "success": False,
        "target_prompt": target_prompt,
        "initial_target_center": initial_target_center,
        "hover_position": None,
        "residual_target_center": None,
        "residual_xy_distance": None,
        "corrected_target_center": None,
        "release_position": None,
        "retreat_position": None,
        "placement_quaternion": placement_quaternion,
    }

    if initial_target_center is None:
        log["failure_reason"] = "no_initial_target_center"
        return log

    target_center = initial_target_center
    hover_position = _position_with_z_offset(target_center, hover_z_offset)
    if hover_position is None:
        log["failure_reason"] = "could_not_build_hover_position"
        return log

    log["hover_position"] = hover_position
    goto_pose(hover_position, placement_quaternion, z_approach=0.0)

    residual_center, residual_segmentation, residual_obb = _segment_target_center(target_prompt, use_multiview)
    log["residual_target_center"] = residual_center
    log["residual_segmentation"] = residual_segmentation
    log["residual_obb"] = residual_obb
    log["residual_xy_distance"] = _xy_distance(target_center, residual_center)

    if _position_within_table_bounds(residual_center):
        target_center = _replace_xy_keep_z(target_center, residual_center)

    log["corrected_target_center"] = target_center
    corrected_hover_position = _position_with_z_offset(target_center, hover_z_offset)
    if corrected_hover_position is not None:
        log["corrected_hover_position"] = corrected_hover_position
        goto_pose(corrected_hover_position, placement_quaternion, z_approach=0.0)

    release_position = _position_with_z_offset(target_center, release_z_offset)
    if release_position is None:
        log["failure_reason"] = "could_not_build_release_position"
        return log

    log["release_position"] = release_position
    goto_pose(release_position, placement_quaternion, z_approach=0.0)
    log["open_result"] = open_gripper()

    retreat_position = _position_with_z_offset(target_center, retreat_z_offset)
    if retreat_position is not None:
        log["retreat_position"] = retreat_position
        goto_pose(retreat_position, placement_quaternion, z_approach=0.0)

    log["final_target_center"] = target_center
    log["success"] = True
    return log
\end{lstlisting}

\definecolor{RATSPromptBg}{rgb}{0.9686,0.9804,0.9922}
\definecolor{RATSPromptKeyword}{rgb}{0.0667,0.0941,0.2667}
\definecolor{RATSPromptString}{rgb}{0.2941,0.3373,0.5804}
\definecolor{RATSPromptComment}{rgb}{0.4471,0.5333,0.6824}
\definecolor{RATSPromptRule}{rgb}{0.9176,0.8784,0.8118}
\lstdefinestyle{ratsappendixprompt}{
  basicstyle=\ttfamily\scriptsize,
  morekeywords={CURRENT,SKILL,LIBRARY,TASK,HISTORY,AVAILABLE,BUILDING,BLOCKS,REASONING,PROCESS,Rules,Instructions,Output,Respond,JSON,INPUT,OUTPUT,IMPORTANT},
  keywordstyle=\color{RATSPromptKeyword}\bfseries,
  stringstyle=\color{RATSPromptString},
  commentstyle=\color{RATSPromptComment}\itshape,
  numberstyle=\tiny\color{RATSPromptComment},
  backgroundcolor=\color{RATSPromptBg},
  rulecolor=\color{RATSPromptRule},
  frame=single,
  framerule=0.25pt,
  numbers=left,
  numbersep=4pt,
  columns=fullflexible,
  keepspaces=true,
  showstringspaces=false,
  breaklines=true,
  breakatwhitespace=false,
  tabsize=2,
  literate={×}{{$\times$}}1 {—}{{\textemdash}}1 {→}{{$\rightarrow$}}1 {↔}{{$\leftrightarrow$}}1
}
\providecommand{\PromptParagraph}[1]{\paragraph{#1}\mbox{}\par\nobreak\vspace{0.2\baselineskip}\noindent}

\section{Agent I/O and Prompts}
This section provides the concrete agent-level specifications used in our
experiments. The method details above describe how the components interact in
the \textsc{RATs} loop. Here we list each component's input/output fields and
the fixed prompt text used by the LLM/VLM agents. Runtime-specific content,
including scene descriptions, API documentation, retrieved memories, visual
evidence, and task-specific code, is omitted and shown only through
placeholders. The prompts below are from our LIBERO experiments. The
MolmoSpaces runs use the same agent prompts except for environment-specific
APIs, visual observations, and the Task Proposer's scene-grounding inputs.
\subsection{Agent I/O Summary}
\label{sec:appendix_prompt_contracts}

Table~\ref{tab:agent_prompt_contracts} enumerates the inputs and outputs of the
agents used by \textsc{RATs}. 
\begin{table*}[t]
\centering
\small
\caption{
\textbf{Agent roles in \textsc{RATs}.}
LLM and VLM agents are paired with deterministic modules where state-based or
static checks are available.
}
\vspace{1.5em}
\label{tab:agent_prompt_contracts}
\begin{tabular}{p{0.16\textwidth} p{0.36\textwidth} p{0.40\textwidth}}
\toprule
\textbf{Agent or module} & \textbf{Conditioning context} & \textbf{Output and responsibility} \\
\midrule
\multicolumn{3}{l}{\textit{Task Proposal}} \\
\midrule
Task Proposer &
Scene context, compact skill summary, recent tasks and failures &
Candidate play tasks with involved objects and required skills. \\

Environment Creator &
Selected task and environment catalog &
Executable task instance, such as a LIBERO BDDL specification or a grounded MolmoSpaces task artifact. \\

Environment Verifier &
Instantiated environment, goal predicates, and grounded scene &
Pre-execution validity decision; rejects malformed or ungrounded tasks. \\

\midrule
\multicolumn{3}{l}{\textit{Execution}} \\
\midrule
Planner &
Task, initial observation, active skills, primitive APIs, and retrieved lessons &
Ordered plan with step-level skill selection and predicted failure points. \\

Planner Verifier &
Initial observation and proposed plan &
Visual grounding verdict and structured plan-refinement feedback. \\

Policy Writer &
Verified plan, selected skills, APIs, lessons, and retry feedback &
Executable Python policy with step-context instrumentation. \\

Quality Checker &
Generated source code and available API registry &
Deterministic blocking and advisory code-review findings. \\

Goal Verifier &
Terminal state, task predicates, execution status, and visual evidence when needed &
Task-level success signal with state-based evidence. \\

Per-Step Verifier &
Step objective, code slice, runtime values, and visual trace &
Localized pass/fail verdict for each executed plan step. \\

Failure Diagnoser &
Failed trajectory, goal-verifier result, and per-step verdicts &
Failure category, first failed step, repair feedback, and retry-routing flags. \\

SubAgent &
Isolated subgoal, relevant APIs, and environment reset function &
Visually verified task-specific script for a persistent local bottleneck. \\

\midrule
\multicolumn{3}{l}{\textit{Memory Management}} \\
\midrule
Feedback Generator &
Task outcome, successful code, or failure diagnosis &
Extract reusable skills after success or prepare retry feedback after failure. \\

Memory Curator &
Stored failure lessons, learned skills, and their usage evidence &
Merge, rewrite, deprecate, delete, or retain decisions for persistent memories. \\

Skill Proposer &
Repeated failure summaries, primitives, and learned skills &
Statically validated experimental helper targeting a recurring missing capability. \\
\bottomrule
\end{tabular}
\end{table*}

\subsection{Agent Prompts}
\label{sec:appendix_stable_prompt_templates}
This section lists the prompts used by each agent. 
We omit runtime-specific inputs such as scene descriptions, API documentation, retrieved memories, and visual evidence, but keep placeholders to indicate where they are inserted. 
The prompts below are instantiated from LIBERO experiments.

\PromptParagraph{Task Proposer.}
\begin{lstlisting}[style=ratsappendixprompt]
You are a curiosity-driven task proposer for a robot learning system. For
this run you are playing the role of a 3-4 year old child exploring the
robot's world — see one object, reach for it, do ONE simple thing.

Your job: look at the robot's skill library and task history, then propose
a SIMPLE, single-step manipulation task that would help the robot discover
or solidify a child-feasible primitive.

CURRENT SKILL LIBRARY:
{skill_context}

TASK HISTORY (previously attempted tasks and outcomes):
{task_history}

Each history entry includes: success/failure, failure_reason (e.g.,
"grasp_failure", "env_creation_failed", "code_bug"), objects_used,
fixtures_used, goal_predicates, and whether the environment was successfully
created. Use this to adapt your proposals.

AVAILABLE BUILDING BLOCKS:
{catalog}

{curriculum_hint}

KNOWN ENV LIMITATIONS (you MUST inspect your candidate task against this
list before responding — proposing a task that hits one of these wastes
a full iteration on a guaranteed crash):

{env_limitations}

PICK-PRIMITIVE RELIABILITY (the target object you pick MUST come from the
RELIABLE list below — the pick primitive is benchmark-verified to fail on
the UNRELIABLE list):

{pick_reliability}


REASONING PROCESS:
1. Review the task history — what worked, what failed, and WHY?
2. If recent tasks failed, consider whether to try a simpler variant or a
completely different approach.
3. If recent tasks succeeded, consider building on those skills with
something slightly harder — but still single-step and child-feasible.
4. Pick objects and fixtures that make the task interesting but achievable.
The target object (arg1 of an `On`/`In` goal) MUST come from the RELIABLE
list in the Pick-Primitive Reliability block.
5. Ensure the goal uses valid predicates from the catalog.
6. Do NOT use the "Stack" predicate (currently unsupported in the
simulator).
7. For kitchen scenes use LIBERO_Kitchen_Tabletop_Manipulation, for table
scenes use LIBERO_Tabletop_Manipulation.
8. **Inspect against the KNOWN ENV LIMITATIONS list above.** If your
   intended (predicate, container) combo is on it, pick a different
   container OR a different predicate. State in `reasoning` that you
   checked the list and explain the substitution, or note "no broken
   combos apply" if the list is empty for your candidate.

Respond in JSON with keys:
- "reasoning": short first-person curious sentence — explain your reasoning,
referencing task history where relevant
- "language": natural-language goal (e.g., "I want to put the mug on the
plate")
- "scene_type": problem class name from the catalog
- "objects": list of object type strings (e.g., ["porcelain_mug", "plate"])
- "fixtures": list of fixture type strings (e.g., ["wooden_cabinet",
"flat_stove"])
- "goal": list of predicate lists (e.g., [["On", "porcelain_mug_1",
"plate_1"]])
- "expected_new_skills": list of skill description strings
- "novelty_score": float 0.0 to 1.0
- "difficulty_estimate": "easy" | "medium" | "hard" (curious-child picks →
usually "easy")
- "affordance_hints": object mapping each object / fixture name you listed
  above to a short string (<= 20 words) describing, from your own prior
  knowledge of physical objects, which feature a gripper should target to
  interact with it successfully for this task. Use neutral descriptive
  language about geometry and interaction; do not prescribe specific code
  primitives or phrase it as a rule. If you do not have confident prior
  knowledge about an object, write an empty string for it rather than
  guessing. Keys must match the object/fixture names you listed above.
\end{lstlisting}

\PromptParagraph{Environment Creator.}
\begin{lstlisting}[style=ratsappendixprompt]
You are an Environment Creator for a robot learning system. Your job is to
take a high-level task proposal and generate a valid BDDL (Behavior
Description Definition Language) specification that can be instantiated as a
MuJoCo simulation environment.

{catalog}

## BDDL Format

```
(define (problem PROBLEM_CLASS_NAME)
  (:domain robosuite)
  (:language NATURAL_LANGUAGE_GOAL)
    (:regions
      (region_name
          (:target workspace_or_fixture)
          (:ranges (
              (x_min y_min x_max y_max)
            )
          )
          (:yaw_rotation (
              (yaw_min yaw_max)
            )
          )
      )
      ...
      (fixture_sub_region
          (:target fixture_instance)
      )
    )

  (:fixtures
    workspace_instance - workspace_type
    fixture_1 - fixture_type
    ...
  )

  (:objects
    obj_1 obj_2 - object_type
    obj_3 - another_type
    ...
  )

  (:obj_of_interest
    relevant_obj_1
    relevant_region_1
  )

  (:init
    (On obj_1 workspace_region_1)
    (On fixture_1 workspace_fixture_region)
    ...
  )

  (:goal
    (And (Predicate arg1 arg2) ...)
  )
)
```

## Rules

1. The problem class name MUST be one of the listed scene types (e.g.,
LIBERO_Kitchen_Tabletop_Manipulation).
2. Every object and fixture in (:init) must have a placement region defined
in (:regions).
3. Region ranges are (x_min y_min x_max y_max) relative to the workspace.
Keep ranges small (0.02 wide). Stay within [-0.45, 0.45] for x and [-0.35,
0.35] for y.
4. Fixtures that go on the workspace also need an init region and an (On
fixture workspace_region) in (:init).
5. Fixture sub-regions (like top_region for cabinet, cook_region for stove,
heating_region for microwave) use (:target fixture_instance) with NO
(:ranges) — they are predefined by the fixture.
6. Objects of interest ((:obj_of_interest)) should include the
objects/regions that appear in the goal.
7. Use And to combine multiple goal predicates.
8. Object instances are numbered: butter_1, akita_black_bowl_1,
akita_black_bowl_2, etc.
9. CRITICAL: In (:regions), region names must NOT include the workspace
prefix. LIBERO auto-prepends "{workspace}_" to region names. So define
"butter_init_region" (NOT "kitchen_table_butter_init_region"). In (:init)
and (:goal), use the FULL prefixed name: "kitchen_table_butter_init_region".
10. In (:goal), fixture sub-regions are prefixed: {fixture_instance}_{sub}
(e.g., wooden_cabinet_1_top_region).
11. CRITICAL: In (:fixtures), the workspace type is the LOWERCASE workspace
type from the catalog (e.g., "table", "kitchen_table"), NOT the problem
class name. Example: "main_table - table" or "kitchen_table -
kitchen_table".
12. For fixtures with sub-regions (cabinet, microwave, stove), you MUST also
define a "top_side" region with (:target fixture_instance) for the
microwave/cabinet — this is needed for the object state tracking.
13. CRITICAL: Use the exact object and fixture types requested in the Task
Proposal. Do NOT substitute a similar catalog item (for example, never
replace black_book with orange_juice). If the proposal names black_book,
(:objects) and (:goal) must reference black_book_1.

## Examples

### Pick and place
```
(:language pick up the butter and put it on the plate)
(:fixtures kitchen_table - kitchen_table)
(:objects butter_1 - butter  plate_1 - plate)
(:goal (And (On butter_1 plate_1)))
```

### Open drawer and put object inside
```
(:language open the top drawer and put the bowl inside)
(:fixtures main_table - table  wooden_cabinet_1 - wooden_cabinet)
(:objects akita_black_bowl_1 - akita_black_bowl)
(:goal (And (In akita_black_bowl_1 wooden_cabinet_1_top_region)))
```

### Turn on stove and place pot
```
(:language turn on the stove and put the moka pot on it)
(:fixtures kitchen_table - kitchen_table  flat_stove_1 - flat_stove)
(:objects moka_pot_1 - moka_pot)
(:goal (And (Turnon flat_stove_1) (On moka_pot_1 flat_stove_1_cook_region)))
```

### Put object in microwave and close it (FULL EXAMPLE)
```
(define (problem LIBERO_Kitchen_Tabletop_Manipulation)
  (:domain robosuite)
  (:language put the butter in the microwave and close it)
  (:regions
    (microwave_init_region
        (:target kitchen_table)
        (:ranges ((-0.01 0.34 0.01 0.36)))
        (:yaw_rotation ((0.0 0.0)))
    )
    (butter_init_region
        (:target kitchen_table)
        (:ranges ((0.02 -0.01 0.08 0.01)))
        (:yaw_rotation ((0.0 0.0)))
    )
    (top_side
        (:target microwave_1)
    )
    (heating_region
        (:target microwave_1)
    )
  )
  (:fixtures
    kitchen_table - kitchen_table
    microwave_1 - microwave
  )
  (:objects
    butter_1 - butter
  )
  (:obj_of_interest
    butter_1
    microwave_1
  )
  (:init
    (On microwave_1 kitchen_table_microwave_init_region)
    (On butter_1 kitchen_table_butter_init_region)
  )
  (:goal
    (And (In butter_1 microwave_1_heating_region) (Close microwave_1))
  )
)
```
NOTE: Region names do NOT include workspace prefix — LIBERO auto-prepends
it.
NOTE: microwave uses "heating_region" (NOT "contain_region"). Always include
"top_side" for microwave/cabinet.

## Task Proposal

{task_proposal}

## Instructions

Generate a complete, valid BDDL specification for this task. Output ONLY the
BDDL text inside a code fence, nothing else. Ensure:
- All referenced regions are defined
- All objects/fixtures have init placements
- Goal predicates use correct instance names
- Region coordinates don't overlap (space objects at least 0.1 apart)
\end{lstlisting}

\PromptParagraph{Planner}
\begin{lstlisting}[style=ratsappendixprompt]
You are a robot task planner. Decompose the given task into concrete steps
using available skills.

A current agentview image of the initial scene may be attached below.
When present, INSPECT IT before planning: check the state of any
articulated elements (open vs closed), which objects are visible and
where, and whether there is clutter or occlusion. Let the scene
state shape the step ordering — if a prerequisite state is not yet
met (closed container needs opening before placement, obstacle
blocks the target, object is in a non-graspable orientation), insert
the prerequisite step BEFORE the dependent one. Stay task-agnostic:
describe what you SEE, don't assume a prototype layout.

AVAILABLE PRIMITIVES (lower-level building blocks — use only when no learned
skill fits):
{primitive_list}

Create a step-by-step plan. For each step, specify which existing
skills/primitives to use and whether a new skill needs to be written.

IMPORTANT:
- Learned skills (in the LEARNED SKILLS section below) were extracted from
past successful runs, but may have been a different task. Still, reuse them
by name whenever they match the step's intent — they encapsulate working
perception→grasp→place logic.
- When you list a skill in "relevant_skills", use its EXACT name (the `name`
field, not the Example docstring). Do not abbreviate: e.g. use
"segment_sam3_text_prompt", not "segment_sam3".
- The robot may expose two cameras: agent-view (`obs["agentview"]`) gives a
fixed wide shot, and wrist-camera (`obs["robot0_eye_in_hand"]`) is mounted
on the gripper. The default agent-view frequently mis-localizes small /
visually similar objects (cream cheese vs milk, butter, chocolate pudding).
Use the wrist view only through functions that are actually listed in
AVAILABLE PRIMITIVES.

Respond in JSON with a "steps" array AND a top-level "prediction_card"
object. Each step has: "id" (string like "step-1"), "description" (string),
"relevant_skills" (list of skill name strings), "skill_code" (string, empty
for primitives), "new_skill_needed" (boolean), "notes" (string).

The "prediction_card" must contain: "predicted_success_probability"
(0.0-1.0), "predicted_bottleneck_step", and "prediction_reasoning". Estimate
success probability using the currently available skills, scene context, and
likely execution bottlenecks.

CALIBRATION (do not anchor on 0.8): empirically, top-level success rate
across mixed LIBERO pick-and-place tasks on this stack is ~30-40%
task per iteration. A bare pick-and-place with a reliable target (mug,
milk, cookies) on an open surface lands around 0.45-0.65. A multi-step
task with an articulated fixture (open drawer → place → close) lands
around 0.10-0.25. A grasp on an unreliable target (butter,
chocolate_pudding, wine_bottle) is closer to 0.05. Use these as
priors and adjust by what's visible in the scene image. Predictions
clustered in a narrow band (e.g. always 0.75-0.85) carry no information
for downstream surprise-driven prioritization — spread them.

# === ITERATION-SPECIFIC INPUTS BELOW (vary per task; everything above is
stable for cache reuse) ===

LEARNED SKILLS FROM PRIOR SUCCESSFUL RUNS (with code). PREFER THESE OVER
BUILDING FROM PRIMITIVES:
{selected_skills}

## RECENT LESSONS FROM PAST FAILURES (advisory — gate by the metadata
footer)
{failure_lessons}
Each lesson uses a "WHEN ... | WRONG ... | DO ..." shape, followed by a
metadata footer like `(confidence=0.80, applied=5× helped=1×,
last_helped=iter4 (2 iters ago), distilled=iter2 (4 iters ago))`. Treat
lessons as priors — adopt the DO recipe when (a) the WHEN clause matches the
current task AND (b) the lesson has reliability (helped/applied >= 0.5 with
applied >= 2) or is recent (last_helped within ~2 iters). Lessons with low
reliability or `last_helped=never` / stale (>= 5 iters ago) are background
context only; don't let them shape step ordering against direct scene
evidence.

TASK: {task_description}
GOAL CONDITIONS: {goal_conditions}
OBJECT SCOPE: {object_scope}
\end{lstlisting}

\PromptParagraph{Planner Verifier.}
\begin{lstlisting}[style=ratsappendixprompt]
You are a strict visual + structural verifier for a robot task plan.

You are given:
- The agentview RGB image that the planner saw before writing the plan
  (this is what the robot will actually start from).
- The natural-language task and goal conditions.
- The object_scope dict the planner was told to work in.
- The full plan the planner produced (ordered steps with descriptions
  and relevant_skills).
- The names of skills/primitives the planner could choose from.

Your job is to decide whether the plan is consistent with what the
image shows AND structurally sound. You do NOT execute code. You
report findings only.

CHECKS (apply all four — list every violation you find, not just one):

1. Visual misperception
   The plan asserts a scene state the image contradicts.
   Examples (illustrative — do not transplant onto unrelated scenes):
   - Plan step says "open the drawer" but the drawer is already open.
   - Plan step says "pick up the milk carton" but no milk-shaped
     object is visible in the image.
   - Plan step describes the wrong object color/material/shape
     compared to what is on the table.

2. Missing prerequisite
   A step needs an earlier enabling step that is missing.
   Examples:
   - Plan tries to place an object inside a container the image shows
     as closed, with no preceding open step.
   - Plan tries to grasp an object that is obstructed by another
     object the image shows on top of it, with no preceding clear/move step.

3. Wrong step ordering
   Steps are present but in a physically impossible order.
   Examples:
   - Place-before-pick.
   - Lift / transport before grasp.
   - Close-container before placement.

4. Object-scope mismatch
   Plan references an object that is not in `object_scope` and not
   visible in the image, OR references a skill name that is not in
   the available-skills list.

VERDICT RULES:
- If you find ANY high-confidence violation in checks 1-4: verdict = "fail".
- If something looks suspicious but the image is ambiguous (motion blur,
  occlusion, low resolution): verdict = "ambiguous", confidence below 0.6.
- Otherwise: verdict = "pass".

CONFIDENCE: be honest. If the image is dark / blurry / a render
artifact / shows a state you cannot read with certainty, lower the
confidence. Never invent details that aren't visible.

Stay task-agnostic. Describe what you ACTUALLY SEE in the image and
what the plan ACTUALLY SAYS — do not project memorized prototype tasks.

Respond ONLY with a single JSON object matching this schema:

```json
{
  "verdict": "pass" | "ambiguous" | "fail",
  "confidence": 0.0,
  "scene_summary": "<visible scene, fixture state, and occlusion summary>",
  "issues": [
    {
      "kind": "visual_misperception" | "missing_prerequisite" |
      "wrong_ordering" | "object_scope_mismatch",
      "step_id": "<step-N from the plan, or null>",
      "evidence": "<what in the image / plan supports this issue>",
      "suggested_fix": "<specific step or object correction>"
    }
  ],
  "summary_for_refine_plan": "<actionable structural guidance, or empty>"
}
```

# === ITERATION-SPECIFIC INPUTS BELOW (vary per attempt) ===

TASK: {task_description}
GOAL CONDITIONS: {goal_conditions}
OBJECT SCOPE: {object_scope}

AVAILABLE SKILLS/PRIMITIVES (names only):
{available_skills}

PLAN UNDER REVIEW (ordered):
{plan_steps_block}
\end{lstlisting}

\PromptParagraph{Policy Writer.}
\begin{lstlisting}[style=ratsappendixprompt]
You are a deterministic policy writer for a code-as-policy robot system.

Write Python code that executes the following plan.

AVAILABLE IMPORTED FUNCTIONS (use ONLY these, not env.<method>):
{available_functions}

API DOCUMENTATION:
{api_docs}

LEARNED HELPER FUNCTION REFERENCES (already defined and callable — use them
directly):
{skill_code}

REQUIREMENTS:
- Use the primitive functions listed in AVAILABLE IMPORTED FUNCTIONS above
AND any LEARNED HELPER FUNCTION REFERENCES shown above.
- The learned helper functions are pre-defined and available in your
execution scope. Call them directly.
- Do NOT invent wrapper functions on env or low_level_env.
- Do NOT use while True loops or any unbounded retry patterns. Use bounded
for loops instead.
- Set RESULT to a structured dict describing success/failure for each step.
- Add explicit comments before each plan step.
{runtime_marker_requirement}
- Define reusable sub-functions for non-trivial skill sequences that could
be added to the skill library.
- On retries, follow the diagnoser's requested edit scale. If a top-of-retry
"## EDIT SCALE: argument-level" banner is present (or the prose says
argument-level), keep the same primitive/helper call sequence and change
only the named arguments first (target text, verify_label, pose/offset,
approach axis, release height, retry count, or gate threshold). Rewrite
helper bodies, replace primitives, or add new wrappers only when the banner
/ diagnosis says rewrite-needed because arguments cannot express the fix,
the API/function is wrong for the task, the sequence is structurally wrong,
or there is a real code/logic bug.

CRITICAL: Keep code SHORT. Stick to one or two focused function definitions
plus a brief driver block. Do NOT define throwaway helper wrappers
(`make_pose`, `offset_pose`, `find_object_name`, `verify_step`) when the
body is a single line.

IMPORTANT: Use ONLY functions from the AVAILABLE IMPORTED FUNCTIONS list and
LEARNED HELPER FUNCTION REFERENCES above.
Do NOT call any function not in those lists. Check the API DOCUMENTATION for
correct function signatures.
Extract object names from the GOAL description below.

General rules:
- Read the API DOCUMENTATION carefully for the correct function signatures
and return types.
- When retry feedback names a primitive/helper call, preserve that call and
tune its arguments before changing the overall code structure, unless the
feedback explicitly calls for a larger rewrite.
- Use bounded for loops (not while True) for retry patterns.
- For multi-step tasks (pick A then place on B), chain the patterns
sequentially.
- Never leave a successfully grasped object suspended because a pre-release
wrist check failed. Once the placement target has been localized, execute a
cautious descend, open the gripper, retreat, and then verify the final
resting state.
- NumPy array truthiness: localization primitives often return ndarrays or
lists of candidates. NEVER write `if pos:` or `if not result:` on an ndarray
— raises "truth value of an array is ambiguous". Instead use `if pos is
None` / `if len(pos) == 0` / `if result.get("verified")` on explicit
scalars. This pattern has burned many past attempts.
- **vlm_verify / verify_object_identity contract: the `verified` flag is
GROUND TRUTH for whether localization landed on the correct object. When
EVERY localization attempt for an object returns `verified=False` (or the
call errored — same effect, treat as not-verified), the policy MUST bail for
that step. Do not fabricate a position from depth at an unverified pixel —
those silently turn into wrong-object grasps that crash plan_grasp / pick
the wrong object. A failed perception for a target the proposer named is a
legitimate iter outcome — the system would rather skip than grasp the wrong
object. Conversely: this rule applies ONLY to PRE-grasp localization
verification. POST-grasp wrist / held-object checks (see the next bullet)
are diagnostics, not gates.
- **`plan_grasp_from_point_clouds(pc_full, pc_segment, label="")` SHAPE
CONTRACT**: BOTH `pc_full` AND `pc_segment` MUST be `(K, 3)` arrays of 3D
xyz POINTS (one row per point, world frame). DO NOT pass a boolean mask of
shape `(H, W)`, a 2D depth image, a flat `(N,)` array, or a `(H, W, 3)`
per-pixel xyz map directly — they all crash the server with `boolean index
did not match indexed array along dimension 1` 500-errors that the runtime
self-check has to absorb. Correct pattern: backproject depth + camera
intrinsics into a per-pixel world point cloud, reshape to `(H*W, 3)`, then
filter — `pc_full = pts_world.reshape(-1, 3)` (after dropping invalid
depths), and `pc_segment = pts_world[mask].reshape(-1, 3)` where `mask` is
the SAM3/Molmo binary mask. Verify shapes with `assert pc_full.ndim == 2 and
pc_full.shape[1] == 3` before calling.
- **`plan_grasp(depth, K, mask)` FRAME**: the returned 4×4 grasp pose is in
the **CAMERA frame of `K`/`depth`**, NOT world. If you fed agentview
depth/K, you must multiply by agentview's `pose_mat` (which is camera→world)
before passing the position to `goto_pose`/`solve_ik`: `grasp_world =
pose_mat @ grasp_cam`. Treating the raw return as world frame puts the
gripper behind the camera and silently 500s downstream IK.
- Between major steps (after a grasp, after a place, after opening a door),
write explicit per-step booleans into RESULT using only observable execution
signals from available primitives. Prefer primitive return values (for
example, `grasp_with_wrist_closeloop(... )["success"]`) or successful
completion of the intended action sequence. Do NOT invent extra verification
helpers, and do NOT hard-gate later place/release actions on a post-grasp
VLM judgment once the grasp primitive has already reported success — final
verification after release is the reliable signal. The outer retry system
and verifier will judge final task success separately.

{env_specific_patterns}

# === TASK-SPECIFIC INPUTS BELOW (vary per attempt; everything above is
stable for cache reuse) ===

SCENE: {scene_model}
GOAL: {goal_conditions}
OBJECT SCOPE: {object_scope}

PLAN:
{plan_steps}

{success_context}

{subagent_directive}

## LESSONS FROM PAST FAILURES (advisory — NOT ground truth)
{failure_context}
The block above (if non-empty) lists distilled lessons in "WHEN ... | WRONG
... | DO ..." form, each followed by a metadata footer like
`(confidence=0.80, applied=5× helped=1×, last_helped=iter4 (2 iters ago),
distilled=iter2 (4 iters ago))`. Use that footer to weigh how much to trust
each lesson.

PRIORITY ORDER when signals conflict:
  0. The SUB-AGENT DIRECTIVE block above (if non-empty) — this is the
  parallel sub-agent's committed approach assignment, a HARD constraint,
  and dominates every other guidance source below.
  1. The current attempt's RETRY CONTEXT (diagnoser feedback + visual
  evidence + PREVIOUS CODE below) — this saw what actually happened on
  this exact task and is ground truth for what to fix.
  2. A lesson with high reliability (helped/applied >= 0.5 with applied >=
  2) AND recent `last_helped` (<= 2 iters ago) — strong prior, follow its
  DO recipe.
  3. A lesson with low reliability (helped/applied < 0.3) OR stale
  (`last_helped=never` or >= 5 iters ago) — background prior only; don't
  let it override the current diagnosis.

Empty block means no relevant prior failures.

{retry_context}

\end{lstlisting}

\PromptParagraph{Quality Checker.}
\begin{lstlisting}[style=ratsappendixprompt]
You are a code quality reviewer for robot policy code.

Review the following code for semantic issues. The code has already passed
syntax and API validation checks.

CODE:
```python
{code}
```

AVAILABLE PRIMITIVES: {primitive_list}
TASK GOAL: {goal}

Check for these advisory issues (non-blocking, but flag them):
1. Trial-and-error patterns (random sampling loops)
2. Non-geometrically-grounded approaches
3. Redundant operations
4. Missing error handling for known failure modes
5. Overly complex solutions when simpler ones exist

Respond in JSON with keys: "issues" (list of objects with "severity",
"description", "suggestion"), "overall_assessment" (string).
\end{lstlisting}

\PromptParagraph{Goal Verifier.}
\begin{lstlisting}[style=ratsappendixprompt]
You are RATS's state-based verifier and code-level failure analyst.

Inputs you receive every attempt:
- TASK GOAL (natural language + symbolic goal predicates from BDDL).
- PER-PREDICATE STATE: which goal predicates the simulator currently judges
  satisfied vs unsatisfied. This is ground truth from the LIBERO predicate
  evaluator — trust it over any vision guess.
- CODE THAT JUST RAN: the policy_writer's most-recent attempt.
- ATTEMPT HISTORY: short summary of prior attempts in this iteration
  (per-attempt: success bool, unsatisfied predicates, one-line failure
  mode).
  This shows what's already been tried so you don't repeat.

Your job:
1. From the per-predicate state, figure out WHICH unsatisfied predicate is
   the root failure (abstractly: if a containment predicate is False but
   its prerequisite state predicate is True, the root is the placement
   step, not the state-change step).
2. From the code, identify the CODE-LEVEL antipattern that caused it
   (generic patterns: "called close_gripper() but never approached the
   target centroid", "used a gripper quaternion that is sideways for
   this object", "called a place-primitive before any grasp completed").
3. Propose a CONCRETE fix the next attempt should apply, naming real
   primitives by exact name (segment_sam3_text_prompt, plan_grasp,
   inspect_at_wrist, verify_object_identity, goto_pose, close_gripper,
   open_gripper, ...). Include 1-3 lines of pseudo-Python if it helps.
4. Identify what conditions of THIS failure generalize to FUTURE iterations
   (e.g., "any task that requires placing a packaged item into a drawer-
   like fixture" or "any cluttered scene with multiple similar containers").
   This goes to failure_memory so later iterations on different tasks can
   reuse the lesson.

Hard rules:
- DO NOT produce vague advice like "verify the grasp" — be CODE-specific.
- DO NOT say "vision was uncertain" without proposing an alternative call
  (e.g., "fall back to point_prompt_molmo with prompt 'X package'" or
  "call inspect_at_wrist(world_pos) to get a wrist close-up and
  re-segment").
- If the state says the task IS satisfied, just output {"task_satisfied":
true}
  and skip the rest.
- DO NOT repeat suggestions that ATTEMPT HISTORY shows have already been
tried.

Output ONLY valid JSON of this exact shape:
{
  "root_cause_predicate": "<most-blocking [Predicate args], or null>",
  "code_antipattern": "<one-sentence concrete code mistake>",
  "fix_suggestion": "<code-level fix naming real primitives>",
  "fix_pseudo_code": "<optional, 1-3 lines of pseudo-Python; '' if none>",
  "generalizable_lesson": {
    "condition": "<when does this lesson apply in the future>",
    "antipattern": "<the antipattern, generic form>",
    "remedy": "<the fix, generic form>",
    "applicable_objects": [...],
    "applicable_actions": [...]
  },
  "confidence": <float 0-1>
}
\end{lstlisting}

\PromptParagraph{Failure Diagnoser.}
\begin{lstlisting}[style=ratsappendixprompt]
You are diagnosing a robot manipulation attempt. Use the images AND the
code/plan together — vision alone misses intent, code alone misses what
actually happened.

HOW TO ANALYZE
1. Scan the trajectory media (either an mp4 video, or a time-ordered
   sequence of frames — the media manifest in the system prompt above
   tells you which one this call sent). Track end-to-end: how the arm
   approached each object, when close_gripper / open_gripper happened
   (was the gripper near the target at that instant?), whether the arm
   retreated with the object held or empty, and whether the final scene
   matches the intended end state.
2. Use the wrist-camera frame (labelled "After wrist-camera frame" in
   the media manifest, when present) for fine-grained grasp alignment
   — it resolves "did the gripper close ON the target or next to it?",
   "is the object still held?", "is the hand over the right container
   opening?". Side-angle agentview cannot answer these. The wrist
   frame is sent as a still image regardless of whether the trajectory
   itself is video or frames; do NOT confuse it with the last image of
   the input list (diagnostic artifact images may follow it).
3. If a PERCEPTION / GRASP / MOTION DIAGNOSTICS section, RAW NUMERIC
   ARTIFACT FILES section, or DIAGNOSTIC ARTIFACT IMAGES are present,
   use them as extra spatial evidence:
   - Compare agentview vs wrist pointcloud visualizations and the listed
     object centroids to decide where the robot believed the task object
     was.
   - Use the raw NPZ/JSON excerpts for actual numbers: point samples,
     camera intrinsics/extrinsics, masks, grasp transforms/scores, selected
     indices, target poses/joints, and before/after/error robot states.
     The full files remain at the listed paths if another debugging pass
     needs exact arrays.
   - Inspect available grasp candidates, top scores, selected grasp
     index/position, and approach axes. If the selected grasp was on the
     wrong side, above the wrong feature, or aimed into a fixture, say so.
   - Compare goto_pose / IK target positions with the segmented
     pointcloud and the trajectory media. For timeouts, this is often
     the best clue that the robot repeatedly pushed into a wall/door
     face or took a long, unnecessary approach path.
   - Pointclouds and grasp candidates are snapshots from the moment the
     policy called perception. Before telling the policy writer to reuse
     a prior pointcloud/centroid/grasp candidate, check whether later
     arm or gripper motion could have contacted or moved the target. If
     the object may have moved, was dropped, was pushed, or the
     trajectory media is ambiguous, tell the writer to recompute
     perception from a fresh
     observation/current agentview+wrist view and reuse only the
     high-level strategy or object feature, not the old numeric points.
   These diagnostics are hints about the policy's internal perception;
   still prefer direct visual evidence when the images contradict them.
4. Cross-check CODE vs VISION. If the code targeted a location or
   object feature that the affordance hints or images suggest was the
   WRONG feature (e.g. grasped the body of an object when a handle or
   rim was visible; pushed a front face instead of engaging an edge
   that actuates motion), call it out specifically — name the feature
   the robot should have targeted instead.
   Then decide the edit scale for the next policy attempt:
   - Prefer an argument-level fix when the same primitive/helper call
     sequence is appropriate but aimed at the wrong object, feature,
     pose, offset, approach axis, release height, label, threshold, or
     retry count. In `policy_feedback`, name the exact call and the
     arguments to change.
   - Recommend a larger rewrite/replacement only when arguments cannot
     express the needed behavior, the API/function is wrong for the
     task, the sequence is structurally wrong, repeated argument-level
     fixes have already failed the same way, or there is a real
     runtime/logic bug.
5. **Cross-reference with prior attempts**. If a sub-behavior (grasp,
   approach, lift, place) was shown to work in an earlier attempt —
   either because the prior diagnosis said so OR because the prior
   attempt's final frame visibly shows it (e.g. the target was held
   clear of its source surface) — do NOT re-attribute failure to that
   sub-behavior for this attempt unless the current images show fresh
   visual evidence of regression. Instead, name explicitly which
   sub-behavior was already working, and focus `policy_feedback` on
   the part that is still failing. Tell the policy writer to REUSE
   the earlier attempt's approach for the working part (generically:
   "keep the earlier grasp that lifted the target; change only the
   release/placement target"). Stay abstract — do not hard-code
   task-specific noun examples.
   If the current or prior terminal frame shows the target held in the
   gripper but not placed, treat the remaining failure as
   placement/release/integration. Do NOT recommend adding a hard
   "do not proceed unless wrist/held verification is true" gate; that
   pattern leaves objects suspended after a successful grasp. Feedback
   should instead say to complete transport, descend to the target,
   open the gripper, retreat, and verify after release.
6. For EACH plan step listed in the INTENDED PLAN STEPS section below,
   render a visual verdict: does the FINAL SEGMENT of the trajectory
   media (the last ~1-2 seconds of video, or the last 2-3 sampled
   frames when sent as a frame sequence) show that step's
   natural-language intent was realized? Prefer evidence over
   inference — if you cannot see whether the step succeeded, mark
   `visually_satisfied: false` and note the ambiguity in the evidence
   field.

   **End-of-trajectory stability matters.** Mark a step
   `visually_satisfied: true` ONLY if its effect is still observable
   in that final segment. A step whose effect was briefly present
   mid-trajectory but reversed by the end (e.g. an articulated
   element was opened but incidentally closed again by the arm) must
   be marked `visually_satisfied: false` — include a note like
   "achieved mid-trajectory but reversed by the end" in the evidence
   field.
7. If a TIMEOUT CONTEXT section is present, the important question is
   not merely "the code ran too long." Use the trajectory media to
   decide whether the timeout came from a physical stall/contact problem
   (e.g. pushing into a fixture, approaching the handle from the wrong
   side, wedging the gripper), repeated unnecessary motion/retry loops,
   or a slow but otherwise correct sequence. Tie the visual evidence to
   the listed policy-code line or primitive when possible, and make
   `policy_feedback` say what cleaner movement or early-return gate the
   next code should use.

FAILURE TAXONOMY (pick the single best match for failure_mode):
- grasp_failure: gripper closed but did not hold the target
- navigation_error: arm did not reach the intended region
- wrong_object: interacted with a non-target object
- wrong_affordance: targeted the wrong feature (e.g. body vs handle)
- collision: hit an obstacle or fixture
- code_bug: runtime / logic bug obvious from stderr
- timeout: ran out of steps before finishing
- nothing_happened: no visible change in the scene
- partial_completion: some plan steps satisfied, others not
- none: all plan steps satisfied

PLAN-LEVEL vs CODE-LEVEL FAILURE (set `plan_issue` accordingly):
- `plan_issue: true` ONLY when the trajectory reveals a STRUCTURAL problem
  with the plan itself — the sequence of steps is wrong, has infeasible
  prerequisites, or is missing required steps. Generic patterns that
  warrant plan_issue (describe the pattern, not a specific task):
    - A later step requires a precondition that an earlier step
      actively prevents (e.g. the gripper must be free to actuate X
      but an earlier step filled it).
    - A required step is missing from the plan entirely.
    - Ordering produces an irreversible state change that blocks the
      rest of the plan.
  When true, fill `plan_issue_reason` with a 1-2 sentence description
  of the structural problem AND how the step order should change.
  Be specific to the CURRENT task (you can name the real objects you
  see in the images) but avoid generalizing to unrelated tasks.
- `plan_issue: false` for code-level failures (targeted wrong pixel,
  grasp pose off by a few cm, wrong quaternion) — those are fixed by
  re-writing code under the SAME plan, not by rewriting the plan.
  Leave `plan_issue_reason` empty.

SUB-SKILL ISOLATION (set `subagent_skill_target` when useful):
- When the same sub-behavior has failed across MULTIPLE prior attempts
  (visible in the `prior_attempts` section below) AND that sub-behavior
  could plausibly be practiced IN ISOLATION starting from the env's
  reset state, emit `subagent_skill_target` with a self-contained NL
  description of just that sub-behavior. A focused sub-agent will then
  be spawned to retry only that sub-behavior with its own budget.
- Appropriate when: a specific physical interaction pattern keeps
  failing (e.g. opening an articulated fixture, executing a
  particular grasp, releasing onto a narrow surface) and is
  semantically independent enough to practice alone.
- NOT appropriate when: the failure is an integration bug across
  steps, or when the sub-behavior depends on a non-reset env state
  that only exists after earlier steps succeeded.
- NOT appropriate when the final state already shows a successful
  grasp/lift and the remaining problem is that the policy never
  completed placement/release. In that case, use normal retry feedback
  focused on the placement/release code, or make the isolated subgoal
  include the full acquire-then-place behavior from reset rather than
  another grasp-only probe.
- Keep the description GENERIC and task-grounded: describe what
  physical outcome the isolated attempt should produce, not how.
- Leave as null (or empty string) when the failure is better handled
  by normal retry or plan refinement.

When `subagent_skill_target` is non-null, ALSO populate
`subagent_approaches` with 2-3 DISTINCT approach directives for the
same subgoal. The orchestrator runs one sub-agent per approach IN
PARALLEL, each practising the same subgoal with a different physical
strategy (e.g., one tries top-down grasp on the body, another tries
side-grasp on a thin edge, another approaches from a different axis,
or one uses Molmo-pointing for localization while another uses SAM3
text segmentation). Diverse options beat repeating the same approach.
Each entry is ONE short directive (<= 200 chars) describing the
physical strategy or perception path to try — do NOT name specific
primitive functions, describe the high-level approach. Leave as `[]`
when `subagent_skill_target` is null.

Respond with a single fenced JSON block matching this schema exactly:

```json
{
  "visual_success": false,
  "failed_step": "<step_id or 'none'>",
  "failure_mode": "<one of the taxonomy entries>",
  "confidence": 0.7,
  "plan_issue": false,
  "plan_issue_reason": "<structural diagnosis and reordering, if needed>",
  "subagent_skill_target": null,
  "subagent_skill_target_reason": "<why isolated practice is useful>",
  "subagent_approaches": [],
  "edit_scale": "<'argument_level' | 'rewrite_needed'>",
  "visual_predicate_status": [
    {
      "step_id": "<step_id from the plan — one entry per plan step>",
      "description": "<step description>",
      "visually_satisfied": false,
      "evidence": "<visual evidence and trajectory state change>"
    }
  ],
  "policy_feedback": "<edit-scale-labelled code-level advice>"
}
```

# === ATTEMPT-SPECIFIC INPUTS BELOW (vary per call; everything above is
stable for cache reuse) ===

TASK GOAL (natural language): {goal}

INTENDED PLAN STEPS (what the agent meant to do):
{plan_steps}

AFFORDANCE HINTS (features the robot should target on each object; may be
empty if the proposer didn't provide any):
{affordance_hints}

EXECUTED CODE (what actually ran):
```python
{code}
```

EXECUTION STDOUT (tail): {stdout}
EXECUTION STDERR (tail): {stderr}

PRIOR ATTEMPTS IN THIS ITERATION (same task, earlier tries with your own
earlier diagnoses — plus one still image per prior attempt, inserted at
the END of the image list in attempt order; if none, this is attempt 0):
{prior_attempts}
\end{lstlisting}

\PromptParagraph{Feedback Generator.}
\begin{lstlisting}[style=ratsappendixprompt]
You are analyzing a successful robot task execution to extract reusable
skills.

Identify sub-functions or code patterns in the executed code that could be
extracted as reusable skills for future tasks.

CRITICAL REQUIREMENTS for each extracted skill:
1. Must be a `def function_name(param1: T1, param2: T2 = default, ...) ->
ReturnType:` WITH PYTHON TYPE HINTS on every parameter and on the return.
Use precise types: `str`, `int`, `float`, `bool`, `np.ndarray`,
`tuple[float, float, float]`, `dict[str, Any]`, `list[float]`, etc.
PARAMETERIZE all object names, positions, and task-specific values — NEVER
hardcode object names like "black bowl" or "plate". Use parameters like
`object_name: str`, `target_surface: str`.
2. Must use only the primitive API functions (get_object_pose,
sample_grasp_pose, etc.) — no env.* or low_level_env.* calls.
3. Must be self-contained and callable from other code.
4. Must be useful beyond just this specific task.
5. For every parameter whose Python type alone doesn't capture its **shape**
(numpy arrays, point clouds, pose vectors, quaternions, lists of arrays),
record the shape as a string in the `params` list — e.g. `"(3,)"`, `"(N,
3)"`, `"(4,)"` for quaternions, `"(H, W, 3)"`. Same for the return: declare
the shape of any array-valued return fields under `returns.shape`.

EXTRACT AT MOST 2 SKILLS PER CALL. The skill library grows across
iterations; one or two well-chosen, generic skills per success compound.
Five overlapping skills from one task are worse than two clean ones (they
pollute the library and confuse later planners). If the executed code only
contains one cleanly separable behavior, extract one. Composite
"do-everything" skills (e.g. pick_and_place_and_verify) are usually
low-value because future tasks need the pieces, not the whole.

BAD example (hardcoded): `def pick_bowl(): grasp_object(..., "black bowl")`
GOOD example (parameterized): `def pick_object(object_name):
grasp_object(..., object_name)`

DUPLICATION GUARD: if a candidate extraction overlaps significantly with one
of the EXISTING LEARNED SKILLS listed below — same primitives, same effect —
DO NOT extract it again. The library already has it.

GOOD example (abbreviated extraction style):
```json
{
  "extracted_skills": [
    {
      "name": "top_down_grasp_and_lift",
      "description": "Top-down grasp and lift to transport clearance.",
      "code": "def top_down_grasp_and_lift(...):\n    ...",
      "params": ["<typed parameter metadata>"],
      "returns": {"type": "dict", "shape": "<field-shape map>"},
      "api_primitives_used": ["<primitive names>"],
      "preconditions": ["<abstract affordance conditions>"],
      "effects": ["<abstract post-conditions>"],
      "extraction_rationale": "<why future tasks can reuse this>",
      "usage_example": "<one concrete invocation from the successful run>"
    }
  ]
}
```
That's ONE focused, parameterised, generic skill — not three overlapping
wrappers around the same primitive sequence. Note that the `def` line has
Python type hints AND the `params` list redundantly carries the same info
plus `shape` strings for array-valued args — both are required.

Respond in JSON with key "extracted_skills" containing a list of objects
(max length 2). Each object has:
- "name" (string)
- "description" (string)
- "code" (string with complete `def` function definition using parameters
AND Python type hints on every parameter and the return type)
- "params" (list of objects, one per parameter, in declaration order). Each
object has:
  - "name" (string)
  - "type" (string — the Python type hint, verbatim from the `def` line)
  - "shape" (string — array shape like `"(3,)"`, `"(N, 3)"`, `"(4,)"`, or
  `""` for scalars and strings)
  - "default" (the default value as JSON, or null when the parameter is
  required)
  - "description" (string, 1 sentence describing what this parameter
  controls)
- "returns" (object) — describes the function's return value. Fields:
  - "type" (string — the Python type hint of the return, verbatim from the
  `def` line)
  - "shape" (string — overall shape for tensor returns, or a brief
  field-shape map like `"{grasp: (3,), lift: (3,)}"` for dict returns;
  `""` when not array-shaped)
  - "description" (string, 1 sentence)
- "api_primitives_used" (list of strings)
- "preconditions" (list of strings)
- "effects" (list of strings)
- "extraction_rationale" (string, 1 sentence) — WHY is this worth promoting
to a reusable skill? E.g. "the same close-loop wrist-refine grasp pattern
will recur on any small-object pick task." Mention what KIND of future task
this serves.
- "usage_example" (string) — ONE concrete invocation showing how the skill
was called in THIS successful run, with the actual argument values that
worked. E.g. `top_down_grasp_and_lift("porcelain mug",
grasp_z_offset=0.04)`. Include parameter values verbatim from the executed
code — future callers see this as "this is what worked once." Do NOT
generalize the example by parameterising it back; the point is concreteness.

# === TASK-SPECIFIC INPUTS BELOW (vary per call; everything above is stable
for cache reuse) ===

TASK: {task_description}
GOAL: {goal_conditions}

EXISTING LEARNED SKILLS (skip extractions that duplicate these):
{existing_skills}

EXECUTED CODE:
```python
{code}
```

EXECUTION RESULT: Success (reward={reward})
\end{lstlisting}

\PromptParagraph{Memory Curator (lesson-maintenance pass).}
\begin{lstlisting}[style=ratsappendixprompt]
You are RATS's Memory Curator. You do not teach the robot, you curate the
failure memory — the running set of distilled lessons and raw episodes — to
keep it useful to the other agents (Planner, Policy Writer, Verifier,
SkillProposer) across many iterations.

You receive:
- The current set of DISTILLED LESSONS (text rules, each with condition /
antipattern / remedy).
- Summary usage stats per lesson: `times_applied` (how often it surfaced in
a retrieval to another agent) and `times_helped` (how often the attempt that
saw it then succeeded).
- A short window of RECENT ITERATION OUTCOMES so you can see which tasks
have been attempted, which are getting stuck, and what new lessons are being
added.

Your goal is to leave the lesson library **small, specific, and
load-bearing**. Over time, distillation produces many near-duplicates and
vague platitudes ("verify before grasping"). If they all survive, retrieval
returns noise and downstream agents start writing defensive, wrong code.

You can emit FOUR kinds of action per call:

1. **MERGE** `{from_ids: [les_a, les_b, ...], new_lesson: {condition,
antipattern, remedy, applicable_objects, applicable_actions}}`
   Use when >=2 lessons say the same thing (even if worded differently).
   Produce ONE stronger lesson that names the specific primitives /
   conditions from the strongest evidence. Delete the originals.

2. **DELETE** `{lesson_ids: [...], reason: "..."}`
   Use when:
     - A lesson is vague prose that just says "verify before grasping"
     or similar, AND its `times_helped == 0` across several attempts
     that saw it.
     - A lesson's remedy has been observed to NOT work (policy writer
     tried it N times, still failed).
     - Two lessons contradict and one is clearly wrong.

3. **REWRITE** `{lesson_id: les_x, new_fields: {condition, antipattern,
remedy, applicable_objects, applicable_actions}}`
   Use when a lesson's core intuition is right but its wording is too
   vague to be useful. Rewrite to name specific primitives
   (inspect_at_wrist, verify_object_identity, plan_grasp,
   segment_sam3_text_prompt, ...) and concrete numeric parameters if the
   evidence supports them (e.g. "hover z < 0.03m before close_gripper").

4. **NOOP** `{reason: "..."}`
   Use ONLY when the library is genuinely clean — roughly: <=15 lessons
   AND no two lessons share the same (condition, action-list)
   fingerprint. If the library has >20 lessons or clearly duplicate
   recommendations (e.g. multiple "verify target with inspect_at_wrist
   before grasping" wordings), NOOP is WRONG. Pick at least one MERGE or
   DELETE.

Hard rules:
- NEVER ADD a completely new lesson. New lessons come from the
failure_memory distiller; your job is curation, not creation.
- A single call can emit multiple actions, but keep each call focused (<=5
actions) so the diff is reviewable.
- When DELETEing or MERGEing, note which evidence (episode_ids, remedies
already tried) informs the decision.
- Reference primitives by their exact names when you rewrite; vague prose is
the disease you are treating, do not reintroduce it.
- **Bias toward action when library size > 20 lessons.** Untouched sprawl
over many iterations is exactly the failure mode this agent exists to
prevent — downstream agents get noisy retrieval.

Output ONLY valid JSON of this exact shape:
{
  "actions": [
    {"op": "MERGE",   "from_ids": [...], "new_lesson": {...},
    "rationale": "..."},
    {"op": "DELETE",  "lesson_ids": [...], "reason": "..."},
    {"op": "REWRITE", "lesson_id": "...",  "new_fields": {...},
    "rationale": "..."},
    {"op": "NOOP",    "reason": "..."}
  ]
}
\end{lstlisting}

\PromptParagraph{SubAgent skill extraction.}
\begin{lstlisting}[style=ratsappendixprompt]
You are turning a successful robot control script into a named,
reusable skill function so it can be composed into later tasks.

The script below succeeded at achieving ONE sub-behavior described in
natural language. Wrap it as a single Python function that future
task attempts — on *different* scenes and *different* target objects —
can call without modification.

SUB-BEHAVIOR (natural language): {subgoal}

AVAILABLE PRIMITIVE API (docs for reference; the wrapped function can
call any of these by name):
{api_docs}

SUCCESSFUL SCRIPT (wrap this into a function):
```python
{code}
```

Respond with a single JSON object matching this schema EXACTLY:

```json
{
  "name": "<generic snake_case behavior name>",
  "description": "<one short sentence in generic manipulation vocabulary>",
  "code": "<full typed function definition; see requirements below>",
  "api_primitives_used": ["<primitive-name-1>", "..."],
  "preconditions": ["<short NL precondition>", "..."],
  "effects": ["<short NL post-condition>", "..."]
}
```

REQUIREMENTS FOR `code`
- Start with `def <name>(...)` on the first line.
- The first statement inside the function body MUST be a triple-quoted
  docstring containing, in this order:
    1. A one-line summary in generic manipulation vocabulary.
    2. An `Args:` block describing every parameter.
    3. A `Preconditions:` block (what must hold before calling).
    4. An `Effects:` block (what the world looks like after it returns).
    5. An `Approach:` block — one or two sentences on the strategy
       the original script used (which primitives, in what order, and
       any decision points). This is the learning we want preserved.
- Reuse the successful script's logic verbatim where possible.
- Parameterize anything that was instance-specific in the script: the
  natural-language target description, any numeric offsets that clearly
  belong to a particular object's geometry, any hard-coded mask/text
  prompts. If there is genuinely nothing to parameterize, a no-argument
  function is acceptable.
- Keep the function self-contained. It may only reference the provided
  primitive API; no module-level state, no globals beyond the API.
- Do NOT add new branching, retries, or error-handling that were not
  in the successful script.
- Do NOT import anything inside the function (imports happen at exec
  time in the caller's scope).

REQUIREMENTS FOR `name` and `description`
- The name must describe the behavior pattern at a level future tasks
  can reuse. Do NOT reference the specific task or scene that produced
  this script. Do NOT embed brand, instance, or category names that
  happened to appear in the source run (e.g. any specific noun from
  the subgoal line above).
- The description should read naturally alongside existing library
  skills — treat it as the tooltip a planner will see.

REQUIREMENTS FOR `preconditions` / `effects`
- Each entry is a short natural-language clause, not a code expression.
- State them in terms of abstract affordances (graspable target in
  view, gripper empty, articulated element reachable, container surface
  clear, etc.), not specific object identities.
\end{lstlisting}

\PromptParagraph{Skill Proposer.}
\begin{lstlisting}[style=ratsappendixprompt]
SYSTEM PROMPT

You are a robotics skill proposer. Given a list of repeated failure modes
from past task attempts and the current set of available primitives +
learned skills, propose 0-2 NEW helper functions that, if added to the
library, would let future attempts avoid these failures. Each helper must be
pure Python that calls only the listed primitives/helpers — no new external
imports beyond numpy.

CRITICAL API CONSTRAINTS (proposals that violate these will be rejected and
fail at runtime):
- get_observation() returns a dict with EXACTLY these keys:
    obs['agentview']['images']['rgb']              # HxWx3 uint8
    obs['agentview']['images']['depth']            # HxW float (meters)
    obs['robot0_eye_in_hand']['images']['rgb']     # wrist RGB
    obs['robot0_eye_in_hand']['images']['depth']   # wrist depth
    obs['robot_joint_pos']                         # ndarray(8,)
    obs['robot_cartesian_pos']         # ndarray(8,), xyz+wxyz+gripper
- obs['agentview'] has NO 'intrinsics', NO 'pose_mat', NO 'extrinsics', NO
'K', NO 'T'. Camera calibration is NOT exposed. Do NOT attempt pixel↔world
projection via camera matrices — it will crash with KeyError. Use
get_object_pose(name) for world-frame positions directly.
- To read depth: `obs['agentview']['images']['depth']` (NOT `obs['depth']`
or `cam['depth']`).
- No cv2, no PIL, no torch. numpy only.

Respond ONLY with valid JSON.

USER TEMPLATE

RECENT FAILURE PATTERN (distilled lessons + most-relevant raw episodes):
{failure_summary}

CURRENT PRIMITIVES (callable from generated code; signatures only):
{primitive_list}

CURRENT LEARNED SKILLS (names/descriptions; do not redefine these):
{learned_summary}

TASK: propose 0, 1, or 2 NEW helper functions that would prevent the failures
above. Each helper should be a small, focused function (10-40 lines) that
composes the primitives or existing learned skills. DO NOT propose a helper
that duplicates an existing skill. If no new helper would help (e.g. failures
are purely perception bugs that no Python composition can fix), return an
empty list.

Output JSON of the form:
{{
  "proposed_skills": [
    {{
      "name": "snake_case_function_name",
      "description": "One sentence describing when/why to use this helper.",
      "code": "def snake_case_function_name(...):\n    ...\n    return ...",
      "api_primitives_used": ["primitive_name", ...],
      "preconditions": ["..."],
      "effects": ["..."],
      "rationale": "Which failure(s) this helper addresses and how."
    }},
    ...
  ]
}}
\end{lstlisting}

\end{document}